\documentclass[pdflatex,sn-nature]{sn-jnl}

\usepackage{graphicx}
\usepackage{amsmath,amssymb}
\usepackage{xurl}
\hypersetup{colorlinks=true, linkcolor=black, citecolor=black, urlcolor=black}

\newcommand{\suppmeth}[2]{#2}
\newcommand{\suppapp}[2]{#2}
\makeatletter
\newcommand{\suppfig}[1]{Supplementary Fig.\@ifundefined{r@#1}{}{~\ref{#1}}}
\newif\ifsn@firstauthor\sn@firstauthortrue\renewcommand{\sn@addauthor}[2]{%
  \ifsn@firstauthor
    \global\sn@firstauthorfalse\gappto\sn@authorlist{\textbf{#2}\ifstrempty{#1}{}{\textsuperscript{#1}}\sn@pendingequal}%
  \else
    \gappto\sn@authorlist{, \textbf{#2}\ifstrempty{#1}{}{\textsuperscript{#1}}\sn@pendingequal}%
  \fi
  \gdef\sn@pendingemail{}\gdef\sn@pendingequal{}%
}
\renewcommand{\email}[1]{\gappto\sn@authorlist{\par\small Correspondence: \texttt{#1}}}
\makeatother
\newcommand{\suppfigs}[2]{Supplementary Figs.}

\begin{document}
\unnumbered

\title[Brain Researcher]{Bringing analytic rigor to agentic AI for science: The Brain Researcher platform for neuroimaging data analysis}

\author[1]{\fnm{Zijiao} \sur{Chen}}
\author[1]{\fnm{Nicholas} \sur{Lu}}
\author[2]{\fnm{Xinhui} \sur{Li}}
\author[1]{\fnm{Jocelyn A.} \sur{Ricard}}
\author[3]{\fnm{Ce} \sur{Ju}}
\author[1]{\fnm{Huan H.} \sur{Wang}}
\author[1]{\fnm{Christian} \sur{Kindermann}}
\author[1]{\fnm{Jeanette A.} \sur{Mumford}}
\author[1]{\fnm{Steven} \sur{Dillmann}}
\author[4]{\fnm{James} \sur{Kent}}
\author[4]{\fnm{Alejandro} \spfx{de la} \sur{Vega}}
\author[1]{\fnm{Sanmi} \sur{Koyejo}}
\author[2]{\fnm{Vince D.} \sur{Calhoun}}
\author[1]{\fnm{Joshua W.} \sur{Buckholtz}}
\author[5]{\fnm{Juan Helen} \sur{Zhou}}
\author[1,6]{\fnm{Steffen} \sur{Bollmann}}
\author*[1]{\fnm{Russell A.} \sur{Poldrack}}\email{russpold@stanford.edu}

\affil*[1]{\orgname{Stanford University}, \orgaddress{\city{Stanford}, \state{CA}, \country{USA}}}
\affil[2]{\orgname{Tri-institutional Center for Translational Research in Neuroimaging and Data Science (TReNDS), Georgia State University, Georgia Institute of Technology, Emory University}, \orgaddress{\city{Atlanta}, \state{GA}, \country{USA}}}
\affil[3]{\orgname{Inria, CEA, Universit\'e Paris-Saclay}, \orgaddress{\city{Palaiseau}, \country{France}}}
\affil[4]{\orgname{The University of Texas at Austin}, \orgaddress{\city{Austin}, \state{TX}, \country{USA}}}
\affil[5]{\orgname{National University of Singapore}, \orgaddress{\country{Singapore}}}
\affil[6]{\orgname{The University of Queensland}, \orgaddress{\city{Brisbane}, \state{QLD}, \country{Australia}}}

\abstract{AI agents can execute scientific analyses, but an analytic output becomes a defensible claim only after alternatives are weighed and the claim is limited to what the evidence supports. Agents may reproduce failures including selective analysis, premature declarations of success and optimization of imperfect criteria. We present Brain Researcher, an agentic research harness operating in a neuroimaging researcher's computational environment under rules for admissible analyses, required checks and claim scope. In benchmarks, Brain Researcher increased first-choice tool-selection accuracy across seven models by 70.2 percentage points (23.3\% without it versus 93.6\% with it) and verifiable grounding from 4.6\% to 22.0\%. In collaborator-led and self-evolving studies, multiverse analyses exposed analytic-choice sensitivity, and scientific review classified claims as accepted, qualified, revised, blocked, rejected or deferred. By linking decisions to evidence and provenance, Brain Researcher embeds methodological judgment within the workflow, not after it.}


\maketitle

Science often advances by absorbing its former frontiers into infrastructure: what once marked the edge of scientific practice, such as sequencing a genome or preprocessing a brain scan, becomes a routine step in a larger workflow. AI agents may represent the next phase of this progression. They increasingly interpret goals, call external tools, observe intermediate results, and choose subsequent actions \citep{boiko_autonomous_2023,swanson_virtual_2025,gottweis_coscientist_2026,huang_biomni_2025,mitchener_kosmos_2025}. Yet scientific research is not only a sequence of procedures, and executing an analysis is not the same as establishing a claim. This distinction is especially consequential in neuroimaging, where large, heterogeneous datasets enter long analysis pipelines with multiple defensible choices. Decisions about preprocessing, parcellation, confound adjustment, and model specification can materially reshape results \citep{poldrack_scanning_2017,li_moving_2024,eklund_cluster_2016,carp_secret_2012}; when seventy teams analyzed the same neuroimaging dataset, no two used the same workflow and their conclusions differed substantially \citep{botvinik-nezer_variability_2020}.

Neuroimaging has also developed one of the most mature open-science ecosystems for computational automation. BIDS and OpenNeuro standardize data organization and sharing \citep{poldrack_past_2024,markiewicz_openneuro_2021}; fMRIPrep automates functional MRI preprocessing \citep{esteban_fmriprep_2019}; and Nipype integrates software packages into reproducible workflows \citep{gorgolewski_nipype_2011}. BIDS Statistical Models and FitLins extend this standardization to machine-readable statistical models and their execution \citep{community_bids_2024,markiewicz_fitlins_2022}, while guided multiverse analysis makes alternative workflows more navigable \citep{dafflon_guided_2022}. Together, these tools make procedures, and increasingly their alternatives, reusable and comparable. They do not, however, bind a selected route to the evidence and methodological conditions required for the claim it is used to support. The challenge is therefore not simply to automate neuroimaging analysis, but to ensure that automation does not obscure the methodological conditions under which an analysis may support a claim.

Tool-using agents create both an opportunity and a risk. They can connect fragmented stages of an analysis and adapt after observing intermediate outcomes, but successful command execution does not guarantee a scientifically valid analysis. An agent may select an inappropriate tool, overlook a data or model incompatibility, stop prematurely, or optimize a criterion that is misaligned with the intended claim. These failures parallel questionable research practices in human-executed science \citep{simmons_false-positive_2011,john_measuring_2012}, and are compounded by premature declarations of task completion \citep{hasan_what_2026,kaddour_agentic_2026} and reward hacking \citep{skalse_defining_2022,gao_scaling_2023}. Scientific agents therefore need more than access to tools: they need reliable routing, evidence linked to its methodological conditions, explicit exposure of alternative specifications, and a visible boundary between formalizable checks and judgments that remain with the researcher.

\begin{figure}[!htbp]
\centering
\includegraphics[width=\textwidth]{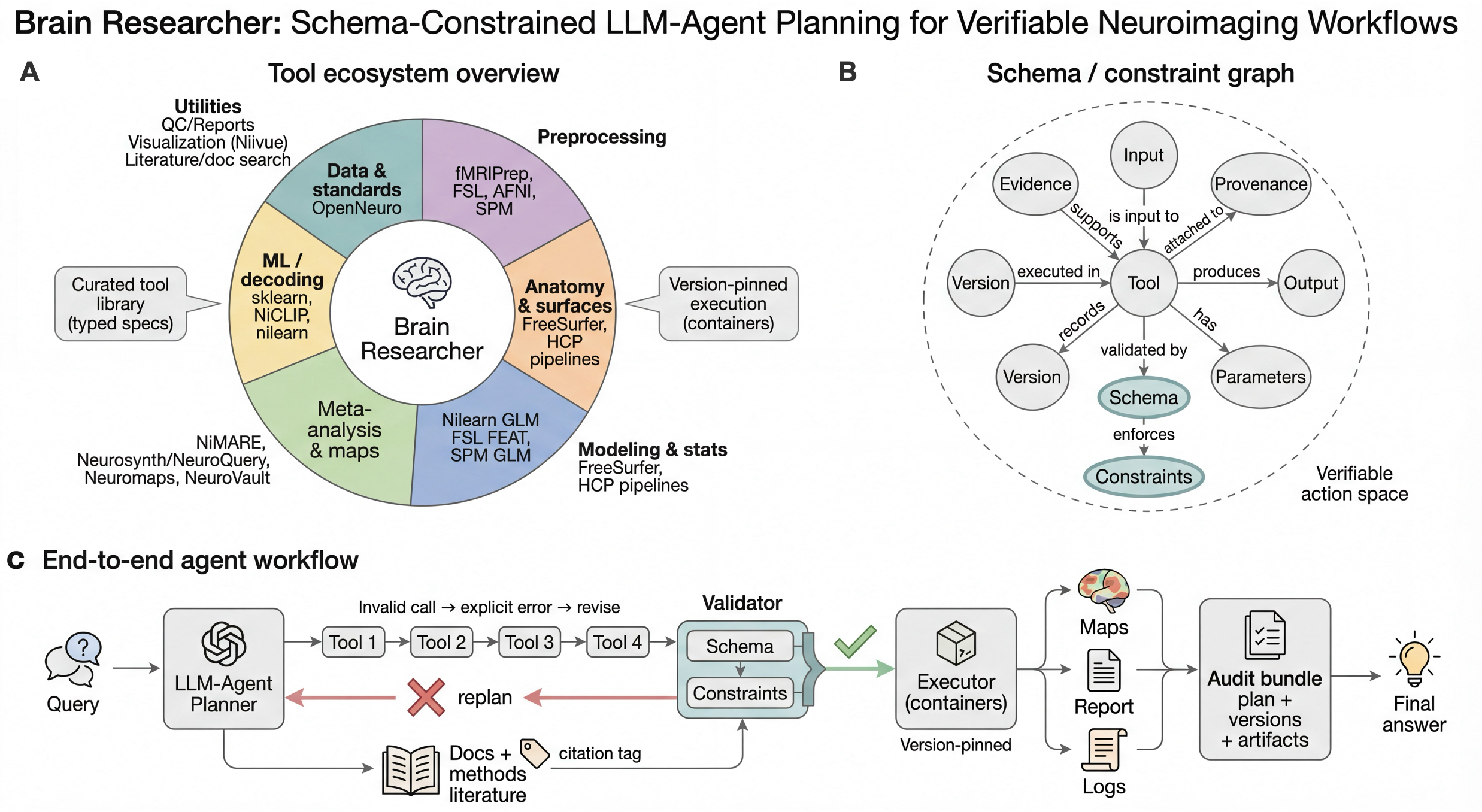}
\caption{\textbf{Brain Researcher: workspace-centric infrastructure for auditable neuroimaging research.} Brain Researcher runs inside the researcher's existing computational environment and exposes neuroimaging analyses as structured, \emph{auditable} operations: every choice, check, and input is recorded as it happens, so that a frozen record of the completed analysis can be read and audited by someone other than the person who ran it, without re-executing it. (a) The tool ecosystem: established neuroimaging software for preprocessing, modeling, meta-analysis, machine learning, quality control, and reporting, each represented by a machine-readable specification and executed in a version-pinned container. (b) Each specification declares its inputs, outputs, parameters, version, evidence anchors, and validation rules; the rule checker tests every proposed call against these clauses before it runs, and records which clauses passed or failed. (c) The episode workflow: the researcher frames a question and approves the plan at the commitment gate, valid actions are dispatched to version-pinned executors, and the resulting audit bundle, containing the committed plan, tool versions, evidence consulted, artifacts, logs, provenance, and the checks each claim passed, feeds the review layer, which writes condition-tagged claims back to memory. This audit bundle is what makes an analysis auditable: a completed run is a fully inspectable research object rather than a one-off result, exported as a compact claim card a reviewer can reopen field by field. Methodological judgment remains the researcher's; the system makes it visible at each stage.}
\label{fig:brain-researcher-main}
\end{figure}

Here we present Brain Researcher, a researcher-governed, domain-specific agentic harness that operates inside a neuroimaging researcher's existing computational environment. Brain Researcher is designed to preserve rather than replace scientific judgment: it operationalizes the parts that researchers can state in advance and records the rest for inspection. For prospectively governed analyses, the researcher specifies the question, admissible analyses, required checks, and scope of any resulting claim. A tool registry and the Brain Researcher Knowledge Graph connect analysis routes to evidence and method conditions; a Model Context Protocol server mediates model actions; and execution and review layers return an audit bundle linking the committed plan, tool calls, artifacts, evidence, and claim verdicts (Fig.~\ref{fig:brain-researcher-main}; \suppmeth{s1}{Supplementary Methods S1}). Multiverse analyses expose sensitivity to defensible specifications \citep{steegen_increasing_2016}, while commitment and claim cards preserve what was decided before and after results were observed. In self-evolving research, intermediate evidence can redirect the trajectory only within a researcher-defined action space specifying the admissible analyses, datasets, and evaluation budget; each successor analysis is frozen before execution. Judgments that resist formalization remain with the researcher.

We evaluated Brain Researcher in three settings. First, a paired seven-model tool-calling benchmark and a separate evidence-citation benchmark tested whether the harness improves upstream tool selection and evidence citation. Second, three collaborator-led studies tested whether multiverse analyses and explicit constraints make claim sensitivity and status visible. Third, two self-evolving research episodes tested whether evidence from one stage could be carried into a frozen successor analysis. Together, these evaluations ask whether AI assistance becomes more capable and more defensible when methodological commitments, evidence, and claim scope are made explicit and auditable.

\section{Results}\label{sec:results}

\begin{figure}[!htbp]
\centering
\includegraphics[width=\textwidth]{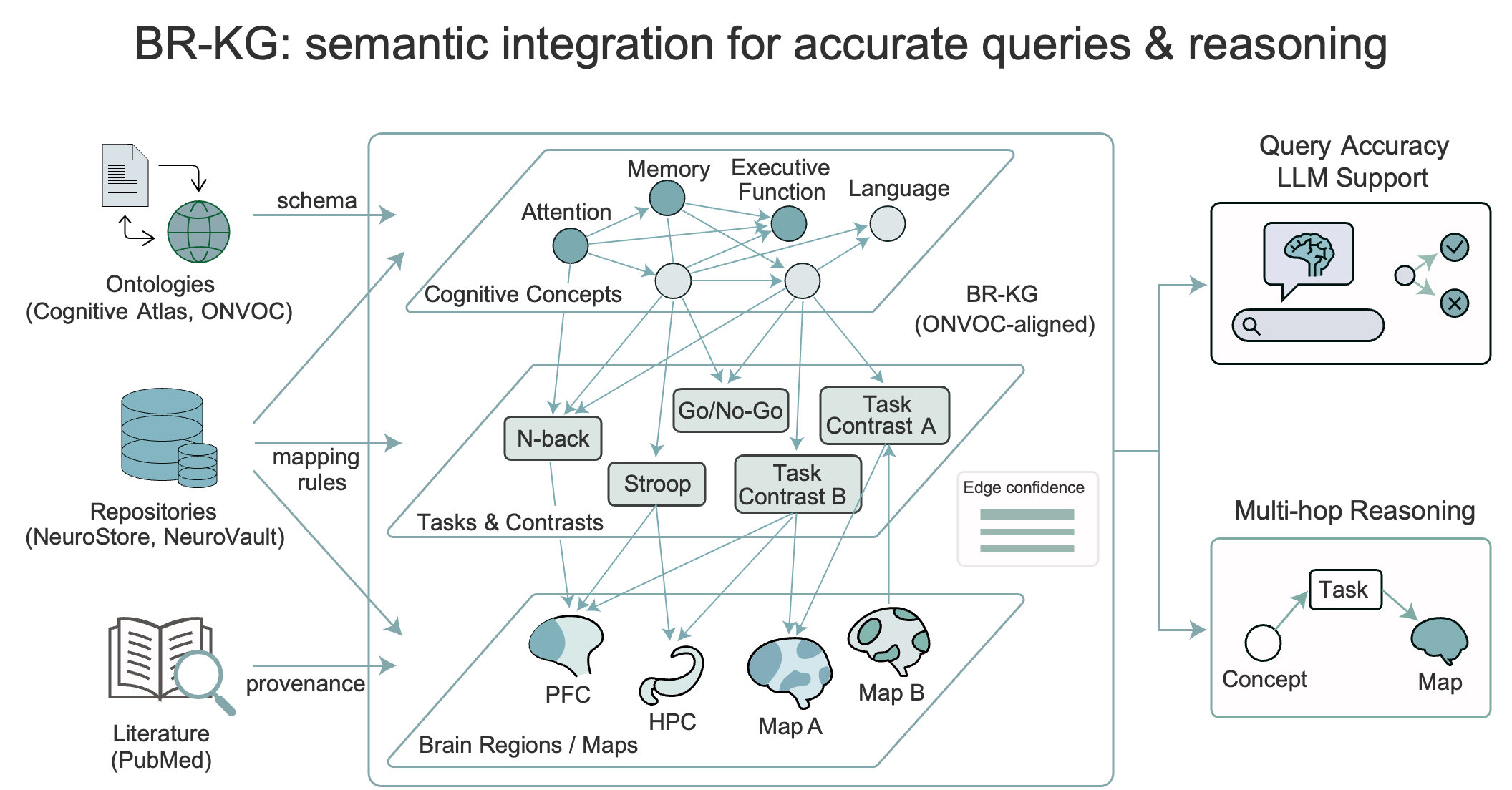}
\caption{\textbf{BR-KG: provenance-linked semantic integration for grounded, auditable neuroimaging reasoning.} BR-KG integrates existing ontologies, repositories, data resources, and literature into a single graph (745{,}949 nodes, 2{,}461{,}469 edges; 2026-07-07 release snapshot) aligned to the OpenNeuro Vocabulary (ONVOC) \citep{onvoc_bioportal}, which normalizes heterogeneous terms to shared identifiers so a query resolves consistently across sources. The central graph links neuroimaging concepts (tasks, contrasts, cognitive constructs), neural representations (brain regions, statistical maps), and research resources (datasets, tools) through typed relationships, with literature evidence attached. Crucially, source-backed facts carry explicit provenance (their source, and where available a verbatim supporting quote and grounding label), so a retrieved claim can be traced back to the study and passage that support it rather than taken on trust; coverage is partial and tracked, and this is what makes retrieval here \emph{auditable}. Downstream panels show the payoff: grounded query answering and multi-hop reasoning over concept-task-map paths, each hop inspectable down to its underlying nodes, edges, and cited evidence, which lets the review layer attach a recommendation's method-condition checks (cohort, paradigm, preprocessing, statistical model) before it is accepted.}
\label{fig:br-kg}
\end{figure}

\subsection{Brain Researcher converts research questions into auditable claim records}
At the center of Brain Researcher is a complete, auditable record of a single run: a persistent \emph{episode} linking the question to its assembled evidence, the analyses the researcher deemed admissible, the committed plan, the execution, the review, and a condition-tagged conclusion (\suppmeth{s2}{Supplementary Methods S2}). Two dated records can anchor it. A \emph{commitment card}, written before any analysis runs, fixes the question, the allowed alternatives, and the success and failure criteria, and is sealed with a content hash so that any later change to the plan is detectable. A \emph{claim card}, written afterward by the review layer, records the resulting claim, its assigned state, scope, and the checks it passed and failed; the supplement includes a specific worked example, built on public Neurosynth data, that a reader can open and inspect field by field (\suppmeth{s8.5.1}{Supplementary Methods S8.5.1}; \suppapp{g}{Appendix G}). A reviewer can then reopen and audit the record without re-executing the analysis. Across the collaborator-led and self-evolving evaluations reported below, every evaluated claim is assigned one of six states (accepted, qualified, revised, blocked, rejected, or deferred), each defined by an explicit adjudication rule.
For example, when a researcher asks whether two groups differ on a functional-connectivity measure, the commitment card records the cohort, the subject groups, and the estimator (either entered by the researcher or resolved from the dataset) and fixes the checks that must pass before anything runs (aligned subject groups, a full-rank design matrix, no feature leakage across folds). If the difference then holds in only some admissible specifications, the claim is recorded as qualified, with its conditions attached.
These methodological decisions are the researcher's; Brain Researcher records each one and enforces the checks it implies.

\subsection{Brain Researcher improves tool calling and evidence citation}
\begin{figure}[!htbp]
\centering
\includegraphics[width=\textwidth]{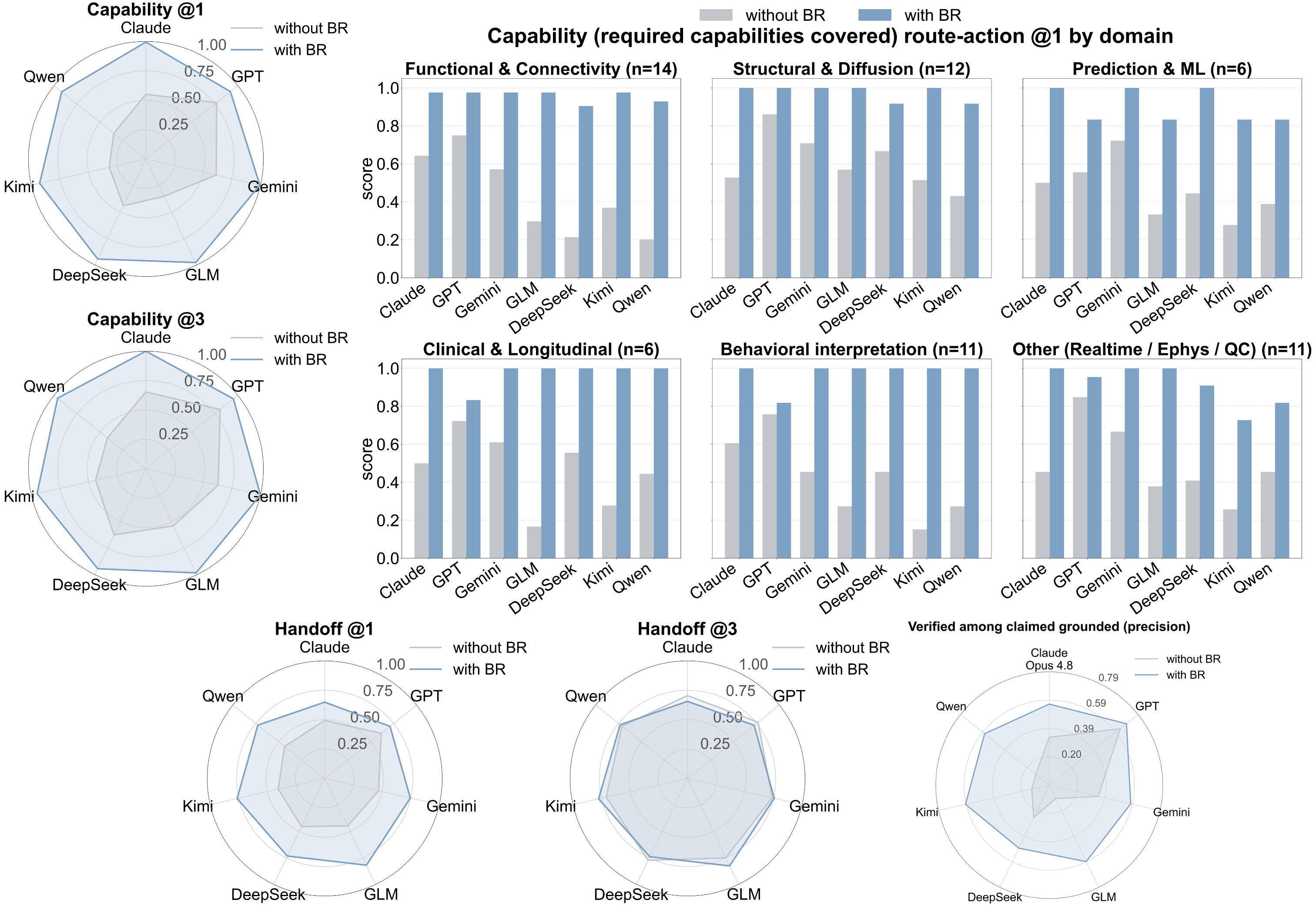}
\caption{\textbf{Summary of Brain Researcher effects across quantitative benchmark tasks.} Without-BR (gray) and with-BR (blue) benchmark performance. Capability@k is mean coverage of required task capabilities after the first k non-neutral actions. The left column reports Capability@1 and @3 across the seven model variants (Claude Opus 4.8, Codex GPT-5.5, Gemini 3.1 Pro, GLM-5.1, DeepSeek-V4-Pro, Kimi K2.5, Qwen3.6-Plus). Upper-right panels break Capability@1 down by task domain. Lower panels report Handoff score@1 and @3 (whether the first route carries enough information for another agent to continue) and a Gemini 2.5 Flash single-judge safeguard: precision among claims marked grounded (fraction whose cited evidence was both locatable and judged supportive). Correct route/tool@1, the first-action selection accuracy reported in the text (23.3\% to 93.6\%), is detailed in \suppmeth{s11.1.1}{Supplementary Methods S11.1.1}. Metrics are interpreted within panel, as denominators and scoring rules differ across benchmarks.}
\label{fig:benchmark-summary}
\end{figure}

We first isolated the infrastructure's effect on two decisions that precede scientific review: choosing the correct analysis tool and citing checkable evidence. The tool-calling benchmark pairs each request with a hidden scoring target and reports Correct route/tool@k (all-or-nothing match of the analysis family and required capabilities), Capability@k (graded capability coverage), and Handoff score@k (whether the proposed route is specified completely enough to execute, with the required inputs present, so it can be handed to a downstream executor and run without gaps) (\suppmeth{s11.1.1}{Supplementary Methods S11.1.1}); three condition-blind LLM judges credit any response that reaches the required capabilities, whether through a Brain Researcher call or an equivalent executable route, so the measured gain reflects reaching those capabilities rather than credit for naming Brain Researcher's specific tools. Both conditions used the same seven frontier models \citep{anthropic_claude_opus_2026,openai_gpt55_2026,google_gemini31_model_card_2026,zai_glm51_2026,deepseek_v4_preview_2026,moonshot_kimi_k25_2026,alibaba_qwen36plus_2026} and general-purpose tools, the without-BR condition lacking only Brain Researcher's registry (\suppmeth{s5}{Supplementary Methods S5}), knowledge graph, and constraint layer. Across 60 tool-calling tasks and seven models, scores without versus with Brain Researcher were 23.3\% versus 93.6\% for first-action correct route/tool selection (with-BR 95\% CI 88.8--97.1, task-clustered), 49.8\% versus 94.5\% for mean Capability@1, and 47.4\% versus 76.1\% for handoff sufficiency. The reference route for each task was fixed before either condition ran and curated with a co-author who does not develop the Brain Researcher system; equivalent non-BR routes were set by two model reviewers and one human (\suppmeth{s11.1.2}{Supplementary Methods S11.1.2}). All seven models improved on all three tool-calling metrics (7/7 positive paired differences; exact two-sided Wilcoxon signed-rank $p=0.016$ for each), with mean gains of 70.2 percentage points for Correct route/tool@1, 44.7 points for Capability@1, and 28.7 points for Handoff score@1. In a routing ablation across 60 tasks and seven models, Brain Researcher without direct KG calls selected an acceptable exact top-1 route in 362 of 420 episodes (86.2\%; model range, 81.7--90.0\%; details in \suppmeth{s11.1.5}{Supplementary Methods S11.1.5}). A separate 50-question benchmark counted a claim as grounded only when its cited evidence could be located and judged supportive; under a three-judge majority vote, the descriptive question-level verified-groundedness rate rose from 4.6\% to 22.0\% (95\% CI 16.8--27.2, question-clustered), a 4.8-fold increase, though most evidence rows still failed, so grounding improved substantially without being solved. Among the 444 non-verified with-BR rows present in all three judge outputs, 65\% received an exact-label majority of real but off-topic and 28\% of partial support; the remaining 7\% lacked an exact-label majority or could not be judged, and none had a fabricated or malformed majority (\suppmeth{s11.1.1}{Supplementary Methods S11.1.1}). Inter-judge reliability and its dependence on judge strictness are reported in \suppmeth{s11.1.2}{Supplementary Methods S11.1.2}. Secondary single-judge safeguards confirmed that the gain was not accompanied by more unrelated citations or lower answer correctness (\suppmeth{s11.1.1}{Supplementary Methods S11.1.1}; \suppapp{j}{Appendix J}; Fig.~\ref{fig:benchmark-summary}). Because the without-BR condition removes Brain Researcher's registry, knowledge graph, and constraint layer together, this contrast measures the harness as a whole rather than isolating any single component; and because the reference routes were curated with a co-author, target construction may share vocabulary with the registry.

\subsection{Brain Researcher runs multiverse analyses to expose claim sensitivity}
\begin{figure}[!htbp]
\centering
\includegraphics[width=0.9\textwidth]{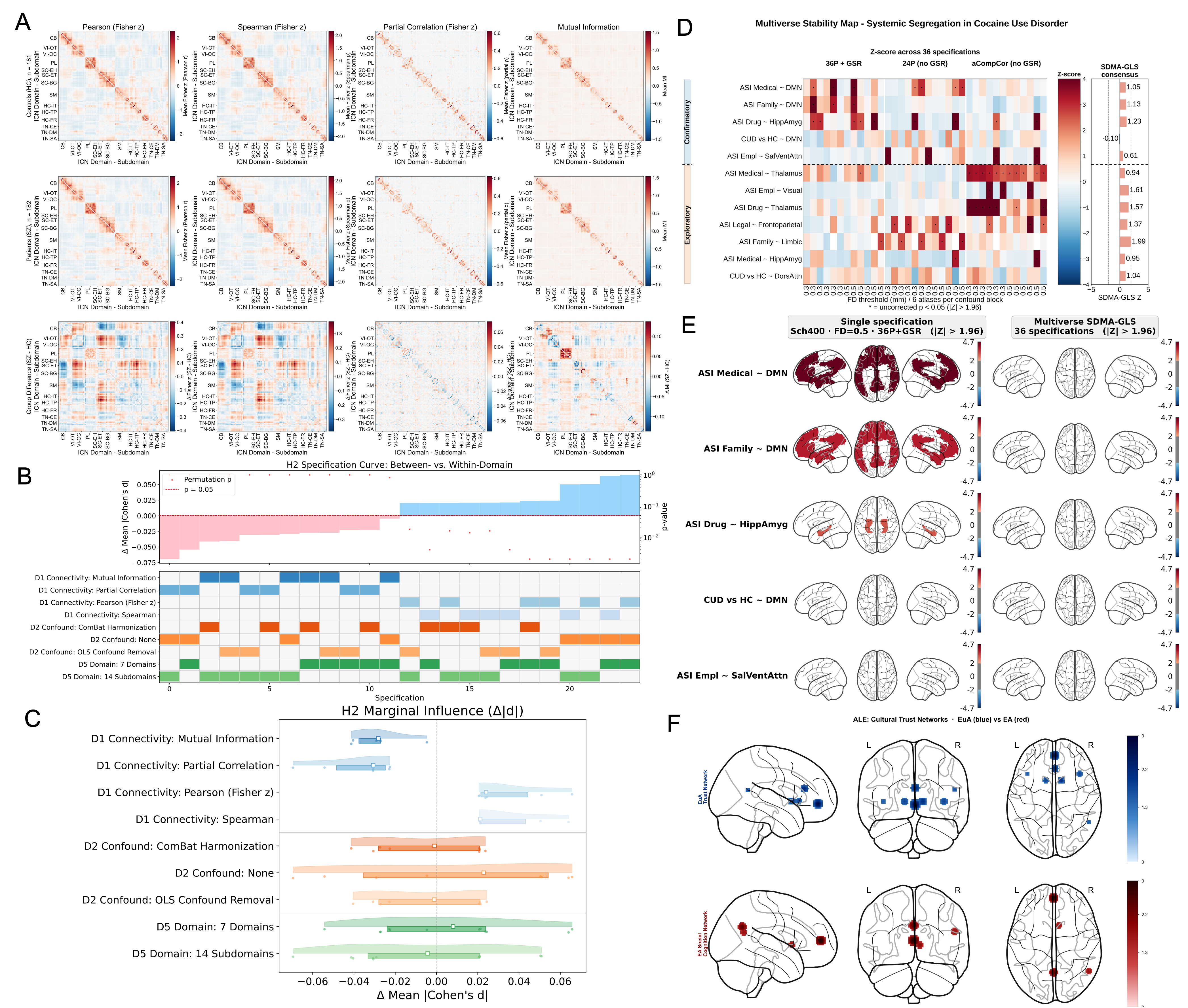}
\caption{\textbf{Multiverse sensitivity and claim-review outcomes across three collaborator episodes.} \textbf{(A--C)} Schizophrenia NeuroMark audit: (A) group-mean functional connectivity for controls (HC, $N=181$), patients (SZ, $N=182$), and their difference across four estimators; (B) NM-H2 (between- versus within-domain) specification curve over the 480-specification multiverse; after sign-aware rescoring, its estimand comprises 24 unique connectivity--confound--domain contrasts, with favorable support at 100\% for Pearson and Spearman and 0\% for partial correlation and mutual information. This complete estimator partition, rather than the pooled 12-of-24 fraction, is the informative result: NM-H2 is measure-dependent, and the mechanism underlying the partition remains unresolved. (C) Marginal influence of each analytic choice on NM-H2. \textbf{(D, E)} Cocaine-use-disorder episode: (D) multiverse stability of systemic-segregation associations across 36 specifications with SDMA-GLS consensus; (E) single-specification versus multiverse SDMA-GLS maps for five network--outcome pairs. \textbf{(F)} Cross-cultural social cognition: culture-stratified ALE maps contrasting Euro-American trust networks with East Asian social-cognition networks.}
\label{fig:leading-case}
\end{figure}

We next evaluated Brain Researcher on three active neuroimaging research questions from collaborating scientists (schizophrenia NeuroMark connectivity, cocaine-use-disorder connectivity, and cross-cultural social-cognition meta-analysis), chosen for heterogeneity in evidence structure without regard to the specific outcomes (Fig.~\ref{fig:leading-case}). Every reported analysis case was run by a coding agent on the local system, which called Brain Researcher for grounding, logging, and review (\suppmeth{s11.2}{Supplementary Methods S11.2}). The NeuroMark case starts from a single, well-established pipeline that its developers use as their standard, giving the audit one clearly defined baseline to build the multiverse around; the other two cases have no such established single pipeline, and test whether the workflow extends to that more difficult setting. The collaborators' hypotheses were pre-specified in their own protocols rather than sealed as commitment cards, so the NeuroMark record is a post-hoc audit of the completed multiverse.

A collaborator studying schizophrenia functional network connectivity using the NeuroMark framework \citep{du_neuromark_2020,iraji_identifying_2023} brought three pre-specified hypotheses for robustness audit: latent connectivity factors outperform individual edges for patient-versus-control classification (NM-H1); between-domain connections show larger group differences than within-domain ones (NM-H2); and latent factors concentrate loading mass on between-domain edges (NM-H3). We evaluated these hypotheses in the FBIRN cohort \citep{keator_fbirn_2016} ($N=363$; 181 controls, 182 patients), parcellated through NeuroMark~2.2 template-based independent component analysis into 5{,}460 edges per subject. In the collaborator's workspace, Brain Researcher expanded the analysis into a 480-specification multiverse spanning connectivity, confound, dimensionality-reduction, classifier, and domain-granularity choices, and recorded and reviewed the resulting runs.

None of the three hypotheses was supported uniformly across specifications; all were recorded as qualified, but for different patterns of conditional support. Under the corrected sign-aware criterion ($p<0.05$ and $\Delta\,\mathrm{mean}|d|>0$), 12 of 24 unique connectivity--confound--domain contrasts favored NM-H2. This pooled fraction obscured a complete estimator split: 100\% of contrasts were favorable under Pearson and Spearman and 0\% under partial correlation and mutual information. NM-H2 is therefore an estimator-regime--dependent finding rather than a generally robust effect. Because partial correlation and mutual information alter the dependence measure in non-equivalent ways, distinguishing shared covariance from estimator scale, power, or nonlinearity requires targeted follow-up. NM-H1 and NM-H3 were also weak: edges outperformed latent factors in aggregate (median $\Delta$AUC $=-0.032$; only 18.8\% of specifications favored latent features), and only 26.0\% favored between-domain loading mass. These claims were qualified rather than rejected because support persisted within identifiable analytic subfamilies (\suppmeth{s11.2.1}{Supplementary Methods S11.2.1}; Fig.~\ref{fig:leading-case}A--C); a claim with no supporting subfamily is rejected instead (\suppmeth{s8.4}{Supplementary Methods S8.4}).

NM-H2 also supplied the audit's governance lesson: automated review missed an error that a human caught. After a server-side fault triggered fallback to a general-purpose coding agent, the agent scored any specification with permutation $p<0.05$ as favorable regardless of sign, inflating apparent support for a directional hypothesis to near-universal levels. The review layer did not flag the error; a human reviewer detected it by inspecting the code, outputs, and specification curve, leading to the corrected rescoring above. Two checks were then added to the Brain Researcher skillset: a directionality test requiring the statistic and acceptance rule to match the hypothesized sign, and a warning whenever execution falls back to a general-purpose agent. The review missed this error. The record nevertheless provided value by binding each claim to its hypothesis, statistic, and conditions, thereby turning a one-off correction into an enforced check.

In the other two episodes, prespecified checks in the scientific review layer determined whether a result could receive confirmatory status.  The SUDMEX CONN (OpenNeuro ds003346; $N=138$) \citep{angeles-valdez_mexican_2022,ds003346:1.1.3} example assessed associations between brain connectivity and behavior; a 36-specification multiverse rejected all five pre-specified connectivity--behavior associations under same-dataset meta-analysis (SDMA-GLS) \citep{lefort-besnard_statistical_2025} (all $Z<1.24$, false discovery rate [FDR] $q>0.58$), and an exploratory screen over 70 combinations surfaced no FDR-surviving effect, so the system blocked it from confirmatory promotion and converted the null into a replication plan (Fig.~\ref{fig:leading-case}D,E). In another test case that applied coordinate-based neuroimaging meta-analysis to a small cross-cultural neuroscience literature, subgroup activation-likelihood estimation (ALE) on 21 studies \citep{eickhoff_coordinate-based_2009,eickhoff_behavior_2016,salo_nimare_2023,dockes_neuroquery_2020} produced a medial prefrontal cortex (mPFC)-topology interpretation, but the system blocked it as exploratory: the subgroups held only $k=6$--$8$ entries (below the recommended $k\ge17$), paradigm composition was imbalanced, and centroid shifts alone cannot establish non-overlapping distributions. The case ended in a paradigm-matched follow-up with no settled claim (Fig.~\ref{fig:leading-case}F). Full statistics are in \suppmeth{s11.2.1}{Supplementary Methods S11.2.1} and \suppapp{j}{Appendix J}.

Across the three episodes, the multiverse exposed which findings were sensitive to analytic choices, while scientific review determined what each result could support. Prespecified review checks withheld confirmatory status from the SUDMEX exploratory screen and the underpowered cross-cultural ALE; in the post-hoc NeuroMark audit, a human reviewer identified the sign-blind scoring error.

\subsection{Brain Researcher converts adaptive searches into frozen successor analyses in two self-evolving episodes}

We next asked whether Brain Researcher could transform open-ended exploration into frozen, auditable successor analyses. We examined two extended research episodes that differed in what was searched. Using the Human connectome Project (HCP) data, Brain Researcher searched over candidate analysis workflows for a fixed question about connectivity-based prediction of behavioural variation. Using the TRIBE foundation model, it searched over candidate scientific questions about a model's internal representations and then converted one question into a frozen test on newly sampled stimuli (\suppmeth{s11.3}{Supplementary Methods S11.3}).

\begin{figure}[!htbp]
\centering
\includegraphics[width=0.9\textwidth]{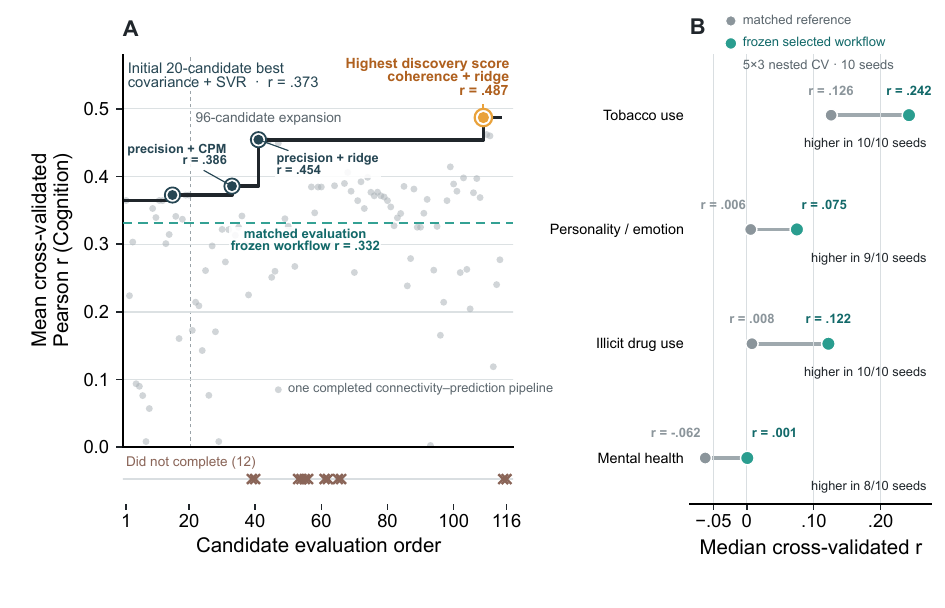}
\caption{\textbf{Brain Researcher searches 116 HCP prediction pipelines and identifies a workflow that consistently exceeds a matched reference.} \textbf{A.} Brain Researcher first evaluated 20 candidate pipelines for Cognition prediction, reaching a best discovery score of $r=.373$. Brain researcher then launched a 96-candidate expansion; 84 candidates returned scores and 12 ended in transport failure. Within the expanded episode, Brain Researcher adapted its proposals to the accumulating results: 27 candidates exceeded the initial search maximum, and the highest discovery score was $r=.487$, obtained with whole-band coherence and ridge regression. Following a selector audit, the researcher froze a related coherence-based workflow for matched evaluation. Across 10 repeated family-grouped $5\times3$ nested-cross-validation runs, this workflow achieved median $r=.332$, compared with $.235$ for the matched reference (median $\Delta r=.098$; conditional one-sided $p=.006$), and was higher in all 10 runs. \textbf{B.} The same frozen selected workflow was then refit for each of four additional behavioural outcomes without target-specific retuning. It produced a higher median correlation for every outcome and exceeded the matched reference in 37 of 40 repeat-level comparisons, giving 47 of 50 directional wins across all five outcomes.}
\label{fig:hcp-search-trajectory}
\end{figure}

In the HCP episode, we began with a published study with openly available code and shared analysis materials \citep{liu_benchmarking_2025}. Brain Researcher allocated 116 candidate prediction-pipeline evaluations for Cognition, of which 104 returned scored results in the parent runs. Following a selector audit, the researcher designated a frozen selected workflow. In 10 repeated same-cohort nested-cross-validation splits, the frozen selected workflow achieved a higher pooled out-of-fold correlation than a matched local reconstruction of the published procedure (median $\Delta r=.098$; conditional one-sided $p=.006$). When the frozen selected workflow was refit to four additional behavioural outcomes, it again produced higher correlations in 37 of 40 comparisons, giving the same direction in 47 of 50 comparisons across all five outcomes. Median out-of-sample $R^2$ was positive only for two variables (Cognition and Tobacco Use), and multiplicity-aware transfer inference remained inconclusive (Fig.~\ref{fig:hcp-search-trajectory}; \suppmeth{iterative-hcp}{Supplementary Methods S11.3.1}).

\begin{figure}[!htbp]
\centering
\includegraphics[width=0.81\textwidth,trim=0 25 0 4,clip]{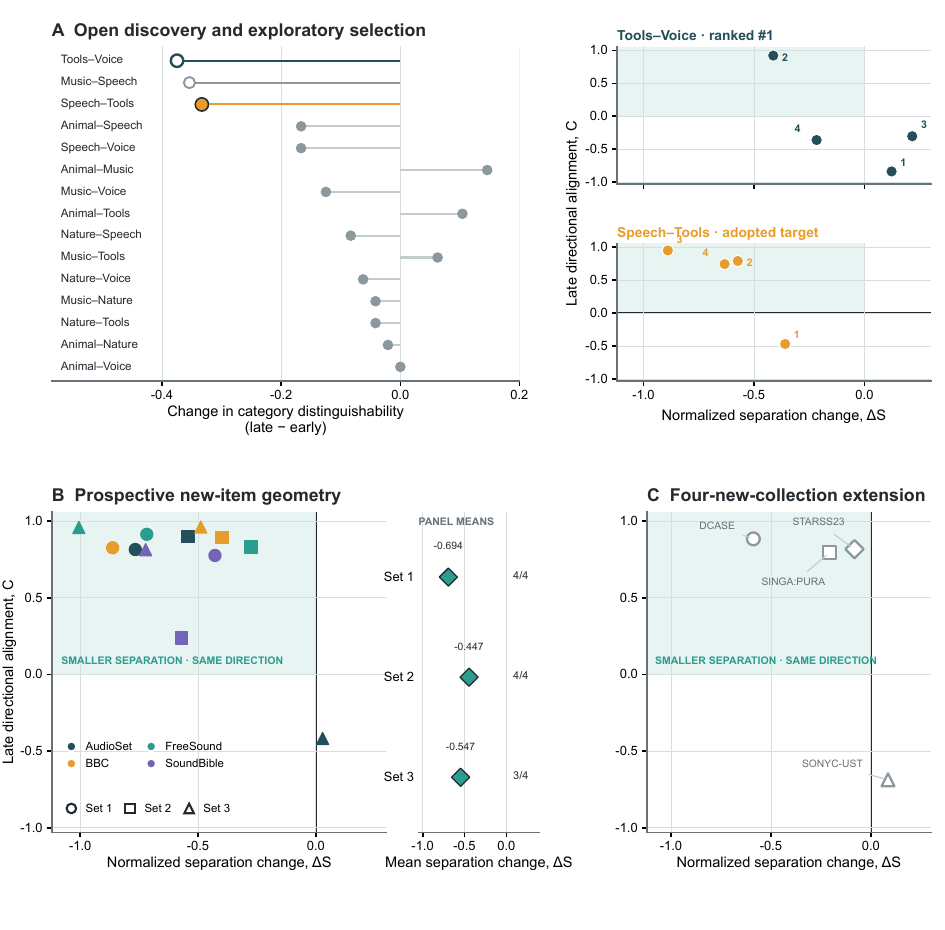}
\caption{\textbf{Brain Researcher turns an open question about TRIBE into successive tests with new sounds and collections.} \textbf{A.} Brain Researcher began by asking how TRIBE changes natural-sound representations from early to late layers. It screened category contrasts by their change in held-out distinguishability (AUC), without choosing a target in advance. Tools--voice changed most, but the pattern varied across collections. Rather than simply following the top-ranked result, Brain Researcher identified speech--tools as a clearer lead: the categories moved closer in later layers while usually keeping the same representational direction. The contrast could also be retested with new, acoustically matched sounds from several collections. The researcher approved this direction and froze the prediction and analysis. \textbf{B.} Brain Researcher then evaluated three non-overlapping 48-item panels. All three showed a smaller speech--tools separation on average in later layers. In 11 of 12 collection-by-panel comparisons, the categories became less separated while retaining the prespecified direction. \textbf{C.} After the pattern recurred across all three panels, Brain Researcher extended the test to four previously unused sound collections. Three of four showed the same geometry, and the late-layer separation was again smaller on average ($\Delta S=-0.198$). In all geometry plots, horizontal position shows the late-minus-early change in normalized separation ($\Delta S$), and vertical position shows late directional alignment ($C$); the upper-left quadrant therefore marks smaller separation with retained direction. A uses fold-specific references, whereas B and C use the frozen speech--tools reference.}
\label{fig:tribe-full-trajectory}
\end{figure}

TRIBE v2 \citep{dascoli_foundation_2026} is a tri-modal foundation model that predicts human fMRI responses from video, audio, and language inputs. We asked how natural-sound category geometry changes across its internal audio layers. Brain Researcher screened category contrasts without choosing one in advance and ranked them by changes in source-held-out discrimination (Fig.~\ref{fig:tribe-full-trajectory}A). Tools--voice showed the largest change but varied across sound collections. Brain Researcher instead proposed speech--tools for follow-up because early layers strongly separated the categories, whereas later layers brought them closer while largely preserving the same representational direction. This contrast could also be tested prospectively using new recordings sampled from multiple collections and matched on seven prespecified acoustic measurements. The researcher approved this direction and froze the hypothesis and analysis before the new stimuli were evaluated.

Brain Researcher then evaluated three successive, non-overlapping 48-item panels. The normalized speech--tools separation became smaller in later layers in 11 of 12 collection-by-panel comparisons, and all three panels met the prespecified directional criterion. In most collections, later TRIBE layers preserved the representational direction separating speech from tools while bringing the categories closer together. The result was a direction-preserving contraction of speech--tools geometry. After the pattern recurred across all three panels, Brain Researcher extended the test to four previously unused sound collections. Three showed the same geometry, although the corrected collection-level test remained inconclusive (Holm-adjusted $p=.396$; Fig.~\ref{fig:tribe-full-trajectory}B,C; \suppmeth{iterative-tribe}{Supplementary Methods S11.3.2}).

Both episodes converted adaptive searches into frozen follow-up analyses. In separately initiated sessions without Brain Researcher, the same coding agent completed substantial analyses, but neither session generated and froze a follow-up study. Because these sessions were not matched controls, this contrast is descriptive and does not establish that Brain Researcher caused the transition from an initial result to a frozen follow-up (\suppmeth{iterative-controls}{Supplementary Methods S11.3.3}).

\section{Discussion}\label{sec:discussion}
Our central contribution is to treat the unit of AI-assisted research as a governed research trajectory rather than a model output. The paired benchmarks tested whether models could reach relevant tools and evidence. The collaborator studies showed how multiverse analysis and explicit constraints change what can be claimed from a completed analysis. The HCP and TRIBE episodes went one step further: a result from one stage became an input to the next. Taken together, these evaluations support a view of scientific agents not as systems that generate a final answer, but as infrastructure that augments human judgments by keeping questions, decisions, evidence, and claim states connected as a project evolves.

This is the sense in which the research episodes were self-evolving. In HCP, Brain Researcher searched over candidate workflows for a fixed question; following a selector audit, the researcher designated the frozen selected workflow and carried it into a matched comparison and four additional behavioural outcomes. In TRIBE, Brain Researcher did not simply promote the contrast with the largest change. It set aside an unstable lead, proposed a more coherent speech--tools question and, after the researcher froze it, carried that question into newly sampled panels. In both cases, intermediate evidence changed the next analysis without rewriting the analysis already under test. The research trajectory evolved, but the evidentiary standard did not.

This trajectory-level view also changes the role of verification. Formal criteria can operate prospectively when they are specified in advance; multiverse analysis can show how a result depends on the enumerated defensible choices \citep{steegen_increasing_2016,simonsohn_specification_2020,dafflon_guided_2022,li_moving_2024,burkhardt_comet_2026}; and judgments that resist formalization remain with the researcher (\suppmeth{s6.1}{Supplementary Methods S6.1}). Relative to systems evaluated primarily for task execution or output correctness \citep{swanson_virtual_2025,huang_biomni_2025,boiko_autonomous_2023,gottweis_coscientist_2026,ghareeb_robin_2026,neuroclaw_2026,chan_mle-bench_2024,chen_scienceagentbench_2025,anthropic_claude_science_2026}, Brain Researcher makes the relationship among the estimator, comparison, evidence base, and claim part of the persistent research record. The NeuroMark example illustrates the limit of this formal layer: automated review missed sign-blind scoring, and a human reviewer found the error. Brain Researcher therefore makes formalizable conditions visible and auditable; it does not make expert inspection unnecessary.

Once decisions and claim states persist, qualified, negative, and failed results need not be terminal outputs. They can narrow the next question, retire an unproductive branch, or define a frozen successor analysis. This is the broader infrastructure implication of self-evolving research: progress can accumulate across successive episodes instead of restarting from an unstructured prompt each time. The mechanism is not unconstrained model autonomy, but the combination of an adaptive trajectory with researcher-defined scope, explicit evidence standards, and durable records of what was tried and why it changed. Researchers retain authority to define the action space, select and freeze successor questions, interpret the evidence, and decide whether the trajectory should continue.

This work has several important limitations (\suppmeth{s12}{Supplementary Methods S12}). Brain Researcher produces auditable evidence but leaves interpretation and writing to the researcher. Its foundation-model and retrieval priors reflect the literature, datasets, and instrumented tools, which may favor well-represented, operationalized questions over negative results, low-resource populations, and unusual paradigms \citep{hao_artificial_2026}. Several episodes rely primarily on same-dataset multiverse or internal validation; these improve auditability but do not replace independent replication or constitute external confirmation \citep{button_power_2013,marek_reproducible_2022}. Several of these datasets are public (HCP, OpenNeuro), so a frontier model may have encountered the associated published findings during training; we cannot rule out memorization, which is a further reason not to treat these results as novel detections. Evidence grounding was scored by condition-blind LLM judges (three frontier models that are also among the seven evaluated, a potential source of self-preference). A reproducible human audit of 20\% of scored results (272 of roughly 1,360 items) agreed with 96\% of verdicts (Cohen's $\kappa = 0.94$); discrepancies were one-step severity differences, never reversals between supported and unsupported, and the judges erred strictly (\suppmeth{s11.1.3}{Supplementary Methods S11.1.3}). Runtime and researcher effort were not measured. Review-layer error was estimated against a 60-case calibration library (16 invalid, 5 valid controls, 39 warn), which produced no false-accepts (0 of 16; rule-of-three 95\% upper bound 19\%) and no false-blocks (0 of 5, a loose bound); this library was assembled after the sign-direction check identified through the NeuroMark case and is not an independent, field-scale estimate (Appendix~G; \suppmeth{s11.3.4}{Supplementary Methods S11.3.4}). The calibration therefore measures internal consistency on canonical scenarios, not how often flawed claims escape review in deployed research workflows. Independent replication and field-scale adjudication remain separate tests of scientific validity, which will require labeled real analyses. Finally, claim records are exportable files, but their value as shared, contestable infrastructure across laboratories remains a future objective. AI assistance should make the conditions under which results become reproducible knowledge easier to see, test, and share.

\section{Online Methods}\label{sec:methods}
Detailed methods, including the runtime stack, the BR-KG substrate and sources, the operation registry, execution backends, benchmark scoring contracts, multiverse and validation-gated search protocols, and all per-case statistics, are provided in the Supplementary Information (Supplementary Methods S1--S12; Appendices A--K, with the per-case episode reports and the item-level benchmark audit sheet released as extended-data Appendices L and M; Supplementary Figures).

\backmatter

\bmhead{Supplementary information}
Supplementary Information accompanies this manuscript.

\bmhead{Funding}
JAR is supported by Stanford University Knight-Hennessy Scholars Program, National Academies of Sciences, Engineering, and Medicine’s Ford Foundation Predoctoral Fellowship, Institute of International Education Quad Fellowship, the National Science Foundation’s Graduate Research Fellowship Program, the Center for Mind, Brain, Computation and Technology, and the Wu Tsai Neurosciences Institute. V.D.C. received support from NSF 2112455 and NIH R01MH123610. A.d.l.V., R.P. and J.K. were supported by the National Institute of Mental Health under award R01MH096906. Z.C. and R.P. received cloud-computing credits through the 2025 HAI-Google Cloud Credits Grant Program to support Brain Researcher API development and computation.

\bmhead{Competing interests}
S.K. reports part-time employment with Meta, which began recently and after most of the work reported here. The other authors declare no competing interests.

\bmhead{Ethics, consent and materials availability}
Not applicable: this work analyzed only previously collected, publicly available or collaborator-provided de-identified neuroimaging data under their original ethics approvals and consents, and generated no new human- or animal-subjects data or materials.

\bmhead{Data availability}
BR-KG is archived at Zenodo (\url{https://doi.org/10.5281/zenodo.21966011}) and linked from the public project site (\url{https://brain-researcher.com/}). The release includes graph snapshots, node and edge schemas, provenance fields, registry links, benchmark manifests, scoring tables, aggregate outputs, figure source data, run-bundle schemas, a worked auditable claim-record example (an exported claim card with its evidence verdicts, on public Neurosynth data), and deployment notes. Users can access the public MCP interface and released Brain Researcher skills from the project site, which describes how users can suggest additions or corrections to BR-KG. Source neuroimaging datasets remain under their original terms: public resources are cited and linked in \suppmeth{s4}{Supplementary Methods S4} and Appendix~C, and controlled-access, collaborator-provided, or license-restricted human-subject data are not redistributed. Artifact and provenance records are described in \suppmeth{s7.4}{Supplementary Methods S7.4} and Appendix~F; benchmark records in \suppmeth{s11.1}{Supplementary Methods S11.1} and Appendix~J. To keep the Supplementary Information self-contained, the full audit ledgers it condenses are released in the same archival repository as an extended-data package: the complete BR-KG, evidence-bundle, dataset, tool-registry, constraint, execution--provenance, and memory data cards (Appendices A--F and H), the full per-rule review registry (Appendix G9.1--G9.4 and G9.6), the automatically generated per-case episode reports (Appendix L), the current HCP and TRIBE research-line reports and their supporting run bundles, and the item-level benchmark human-audit sheet (Appendix M); a crosswalk maps each condensed Supplementary section to its archived file.

\bmhead{Code availability}
The Brain Researcher system (Python package, CLI, agent runtime, MCP server with versioned tool contracts, orchestrator, web UI, and deployment recipes) is available under the MIT license at \url{https://github.com/brain-researcher/brain-researcher-public}, with the companion agent layer (skills, agent templates, MCP adapters, and AutoResearch evaluation rubrics) at \url{https://github.com/brain-researcher/brain-researcher-agent-kit}; both are linked from the project site (\url{https://brain-researcher.com/}), and the v0.3.0 release is archived at Zenodo (\url{https://doi.org/10.5281/zenodo.21966011}). Analysis code and per-specification outputs for the NeuroMark collaborator case are available at \url{https://github.com/XinhuiLi/BR-NeuroMark}.

\bmhead{Author contributions}
Z.C. and R.P. initiated and conceived the project. Z.C. designed and implemented the Brain Researcher system, ran the experiments and analyses, generated the main results, and drafted the manuscript. R.P. supervised the project and contributed to conceptual framing, study design, hands-on system testing, evaluation feedback, interpretation, and manuscript revision. J.H.Z. provided early supervision and initial computational resources for the project. N.L. gathered background information, including dataset lists and literature-review materials, helped design the benchmark questions, evaluated system outputs, and tested performance for the quantitative benchmark section. X.L. and V.D.C. designed and ran the NeuroMark schizophrenia functional-network-connectivity case. J.R. and R.P. designed and ran the cocaine-use-disorder connectivity case. H.W. designed and ran the cross-cultural social cognition case. C.K., J.K., and A.d.l.V. provided feedback on knowledge-graph design and contributed data and design requirements for the knowledge-graph and source-integration components. S.B. provided feedback on agent design, MCP infrastructure, backend integration, and execution design. J.M. provided feedback on the scientific-review layer. S.D. contributed suggestions and ideas on the agent harness and validation-gated research design, including the bounded-validation framing; S.K. provided feedback on the agent harness. C.J. contributed to system testing. J.W.B. provided feedback on the manuscript. All authors reviewed and approved the manuscript.

\bibliography{sn-bibliography}

\end{document}


\title[Supplementary Information]{Supplementary Information for ``Bringing analytic rigor to agentic AI for science: The Brain Researcher platform for neuroimaging data analysis''}

\ifdefined\anonymousbuild
\makeatletter
\gdef\sn@authorlist{}
\gdef\sn@affillist{}
\makeatother
\else
\author[1]{\fnm{Zijiao} \sur{Chen}}
\author[1]{\fnm{Nicholas} \sur{Lu}}
\author[2]{\fnm{Xinhui} \sur{Li}}
\author[1]{\fnm{Jocelyn A.} \sur{Ricard}}
\author[3]{\fnm{Ce} \sur{Ju}}
\author[1]{\fnm{Huan H.} \sur{Wang}}
\author[1]{\fnm{Christian} \sur{Kindermann}}
\author[1]{\fnm{Jeanette A.} \sur{Mumford}}
\author[1]{\fnm{Steven} \sur{Dillmann}}
\author[4]{\fnm{James} \sur{Kent}}
\author[4]{\fnm{Alejandro} \spfx{de la} \sur{Vega}}
\author[1]{\fnm{Sanmi} \sur{Koyejo}}
\author[2]{\fnm{Vince D.} \sur{Calhoun}}
\author[1]{\fnm{Joshua W.} \sur{Buckholtz}}
\author[5]{\fnm{Juan Helen} \sur{Zhou}}
\author[1,6]{\fnm{Steffen} \sur{Bollmann}}
\author*[1]{\fnm{Russell A.} \sur{Poldrack}}\email{russpold@stanford.edu}

\affil*[1]{\orgname{Stanford University}, \orgaddress{\city{Stanford}, \state{CA}, \country{USA}}}
\affil[2]{\orgname{Tri-institutional Center for Translational Research in Neuroimaging and Data Science (TReNDS), Georgia State University, Georgia Institute of Technology, Emory University}, \orgaddress{\city{Atlanta}, \state{GA}, \country{USA}}}
\affil[3]{\orgname{Inria, CEA, Universit\'e Paris-Saclay}, \orgaddress{\city{Palaiseau}, \country{France}}}
\affil[4]{\orgname{The University of Texas at Austin}, \orgaddress{\city{Austin}, \state{TX}, \country{USA}}}
\affil[5]{\orgname{National University of Singapore}, \orgaddress{\country{Singapore}}}
\affil[6]{\orgname{The University of Queensland}, \orgaddress{\city{Brisbane}, \state{QLD}, \country{Australia}}}
\fi

\maketitle

\setcounter{secnumdepth}{0}

\section*{Supplementary Methods}
\phantomsection\label{supp:methods}
\addcontentsline{toc}{section}{Supplementary Methods}

\subsection*{Organization of supplement}

The Supplementary Methods follow the life cycle of a Brain Researcher episode. The main Methods define the conceptual machinery: what a research episode is, why MCP separates the language model from scientific infrastructure, how tools and datasets are selected, how execution is observed, and how claims are reviewed before memory writeback. This supplement makes that machinery reproducible by specifying the records, checks, gates, artifacts, and evaluation rules used during implementation.

The sections are organized in the order in which an episode runs. \hyperref[supp:s1]{S1} defines the system boundary and MCP interface. \hyperref[supp:s2]{S2} defines the episode object, state machine, and a running predictive-modeling example. \hyperref[supp:s3]{S3} describes BR-KG, ONVOC normalization, and evidence assembly. \hyperref[supp:s4]{S4} describes dataset and resource resolution. \hyperref[supp:s5]{S5} defines the typed tool registry and action space. \hyperref[supp:s6]{S6} defines the constraint compiler and commitment gate. \hyperref[supp:s7]{S7} defines execution, preflight, and provenance. \hyperref[supp:s8]{S8} defines the six terminal claim states used for the reported collaborator and bounded-autonomous evaluations and routes unresolved decisions for human escalation. \hyperref[supp:s9]{S9} defines memory writeback, conflict detection, and BR-KG promotion. \hyperref[supp:s10]{S10} defines bounded self-evolving episodes and Harbor-based validation. \hyperref[supp:s11]{S11} defines the evaluation protocol. \hyperref[supp:s12]{S12} states limitations and reporting boundaries. \hyperref[app:data-cards]{Appendix/\allowbreak{}Data Cards A--J} contain concrete episode identifiers, snapshot identifiers, schema excerpts, run-bundle excerpts, review rules, claim cards, and evaluation ledgers. \hyperref[app:k]{Appendix K} presents a representative generated-code excerpt for a collaborator case; the full per-case reports for the three collaborator cases and two bounded self-evolving campaigns are released with the repository and described in \hyperref[supp:case-reports]{Appendix L}.

\subsection*{Public release and reuse}
\phantomsection\label{supp:availability}
\addcontentsline{toc}{subsection}{Public release and reuse}

The public release includes BR-KG itself. It provides graph snapshots, node and edge schemas, ontology-normalization records, public evidence and provenance fields, registry links, and example graph paths for the material described in \hyperref[supp:s3]{S3} and Appendix~\hyperref[app:b]{B}. The project site, \url{https://brain-researcher.com/}, provides access to the public MCP interface, the released Brain Researcher skills, and deployment notes: environment setup, graph loading, registry initialization, and example queries.

Source datasets are handled separately. Public resources are cited and linked, but not re-hosted unless their licenses permit redistribution. \hyperref[supp:s4]{S4} and Appendices~\hyperref[app:b]{B} and~\hyperref[app:c]{C} record the datasets, derivatives, metadata, and access checks used by Brain Researcher. Controlled-access, collaborator-provided, or license-restricted human-subject data remain with the original data custodians or contributing groups.

The release includes the records needed to check the reported results without redistributing restricted raw data. \hyperref[supp:s7]{S7} and Appendix~\hyperref[app:f]{F} define the run-bundle schema, provenance fields, artifact manifests, logs, parameters, and observed outputs. Derived summaries, aggregate tables, figure source data, task manifests, scoring sheets, and run-bundle metadata are included in the public package.

\hyperref[supp:s11]{S11} and Appendix~\hyperref[app:j]{J} define the benchmark release: task manifests, model conditions, with-BR/\allowbreak{}without-BR comparisons, scoring rules, metric denominators, exclusions, and aggregate outputs. Private paths, provider keys, human-subject identifiers, and license-restricted source material are excluded from the public release. Code-release details are tied to the registry and execution records: \hyperref[supp:s5]{S5} describes the callable tool cards, \hyperref[supp:s7]{S7} describes execution provenance, and Appendices~\hyperref[app:d]{D} and~\hyperref[app:f]{F} give the corresponding registry and artifact-manifest records.

We welcome community extensions to BR-KG. The public site explains how to propose new nodes, edges, evidence records, registry links, and deployment recipes. Contributions must include source provenance, license information, schema-valid records, and review status, so additions remain traceable as the graph grows.

\subsection*{Key operational terms}

\begin{center}
\small
\setlength{\tabcolsep}{5pt}
\renewcommand{\arraystretch}{1.15}
\begin{tabular}{|>{\RaggedRight\hspace{0pt}\arraybackslash}p{0.28\textwidth}|>{\RaggedRight\hspace{0pt}\arraybackslash}p{0.63\textwidth}|}
\hline
\rowcolor{brtablehead}\textbf{Term} & \textbf{Meaning in this supplement} \\
\hline
Research episode & Persistent runtime unit containing the question, evidence, selected plan, resources, constraints, execution artifacts, review verdicts, and memory status for one investigation. \\
\hline
MCP operation & Typed endpoint exposed by BR-MCP, for example tool discovery, dataset resolution, recipe generation, artifact inspection, or review. \\
\hline
Registry specification & Machine-readable record used for discovery, compatibility checking, parameter validation, and recipe generation. \\
\hline
Workflow specification & Planning-level description of a family of analyses and its expected inputs, outputs, constraints, and backend recipes. \\
\hline
Executable wrapper & Actual callable Python tool, command-line wrapper, Neurodesk module, container command, or scheduler script. \\
\hline
Resource ledger & Episode-specific record of datasets, derivatives, covariates, masks, target variables, fold manifests, and readiness status. \\
\hline
Constraint compiler & Component that converts method-condition records, tool-contract clauses, and validators into hard or soft checks. \\
\hline
Run bundle & Immutable provenance record containing event trace, trajectory document, observation record, analysis bundle, and run card. \\
\hline
Scientific verification layer & Acceptance gate that checks run bundles, artifacts, and candidate claims against validity rules before claim communication or memory writeback. \\
\hline
Review card & Structured verification verdict, including rule triggers, risk tags, required fixes, and claim eligibility. \\
\hline
Claim card & Reviewed memory object derived from a run bundle; it stores a claim with scope, evidence level, caveats, and provenance pointers. \\
\hline
Harbor verifier & Task-level execution checker used in bounded autonomous episodes. It can verify files, schemas, tests, and required statistics, but it is not the scientific reviewer. \\
\hline
\end{tabular}
\end{center}

\section{S1. System architecture and MCP interface}\label{supp:s1}

\subsection{S1.1. Overall architecture}\label{supp:s1.1}

Brain Researcher is organized around a research episode rather than a single prompt-response exchange. Each episode enters through an MCP-compatible client or coding-agent harness \citep{anthropic_introducing_2024,specification_specification_2024}, is mediated by the Brain Researcher MCP control plane, and is grounded by domain backends. The outer layer receives the researcher's goal, writes or runs code when needed, interacts with the user, and presents results. The middle layer, BR-MCP, exposes typed operations for each stage of the episode lifecycle: planning, discovery, execution, review, and memory writeback. The full operation surface is enumerated in \hyperref[supp:s1.3]{S1.3}. The inner layer consists of domain backends: BR-KG, dataset and BIDS resolvers \citep{gorgolewski_brain_2016,poldrack_past_2024}, the tool registry, execution backends, storage services, review-rule registries, memory stores, and governance records.

\begin{figure}[!htbp]
\centering
\includegraphics[width=\textwidth]{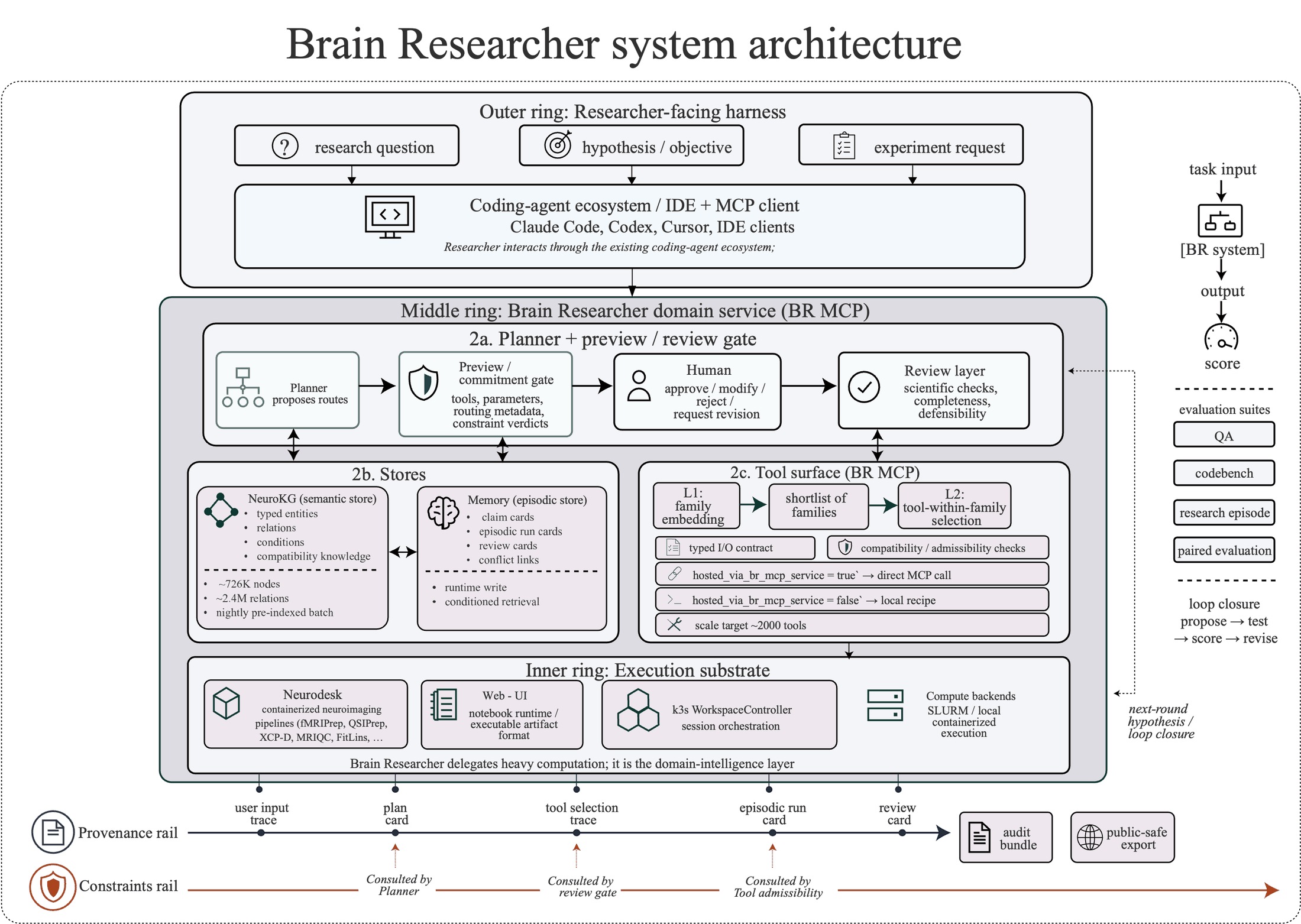}
\caption{\textbf{Detailed Brain Researcher system architecture.} The architecture separates the researcher-facing coding-agent harness, the BR-MCP control plane, and the domain execution substrate. The episode flow runs from question intake and typed intent resolution through BR-KG and registry lookup, resource checks, constraint compilation, commitment, execution, review, and memory writeback. The control-plane boxes indicate where typed MCP operations expose admissible actions, while the backend boxes indicate where Python, container, Neurodesk, Slurm/HPC, storage, and provenance services produce environment-observed artifacts. Dataset readiness, tool admissibility, execution success, and claim eligibility are returned by typed system state; the language model interprets, requests, and explains those states.}
\label{fig:system-architecture}
\end{figure}

A typical predictive-modeling request traverses the layers as follows. The user asks whether a behavioral score can be predicted from a resolved neuroimaging feature matrix. The outer client converts the request into a typed intent: prediction, fMRI/\allowbreak{}connectivity features, target variable, covariates, and required folds. BR-MCP retrieves candidate datasets, workflows, and tool families from the registry and BR-KG. The resource resolver checks whether the feature matrix, target vector, covariates, and fold manifest are reachable and authorized. If the resolver returns pass or warn, BR-MCP packages an execution recipe with expected artifacts; if it returns block, the plan returns to revision before execution. After execution, the scientific verification layer inspects the run bundle and candidate claims before memory writeback.

This separation is intentional. The language model may interpret a natural-language request, compare MCP-returned options, ask for recipes, and explain observed outputs. It cannot certify that a registry ID exists, override a resolver block, mark a run as successful without artifacts, or accept a scientific claim without a verification verdict. Tool availability, dataset readiness, policy permission, execution success, and claim eligibility are supplied by MCP, BR-KG, the registry, resolvers, execution backends, and the scientific verification layer.

\subsection{S1.2. Model routing and reproducibility}\label{supp:s1.2}

Language-model routing affects generation and interpretation, not certification. Coding-oriented tasks such as script generation, debugging, and file manipulation may be routed to code-specialized models. Scientific planning, evidence interpretation, review, and claim calibration may be routed to analytical models. The run card records the task type, selected provider, prompt-template version, policy context, registry snapshot, BR-KG snapshot, and memory namespace.

Example run-card excerpt:
\begin{suppcode}
run_card:
  episode_id: ep_hcp_predict_001
  mode: benchmark
  model_route:
    task_type: code_generation
    provider: fixed_provider_id
    prompt_template: planning_v3
  fixed_snapshots:
    registry: registry_2026_05_15
    brkg: brkg_snapshot_2026_05_10
  policy_context: benchmark_locked
  memory_namespace: benchmark_isolated
\end{suppcode}

For reported evaluations, the full configuration is fixed before launch and recorded in the run card: model identity, prompt-template and run-card schema versions, registry and BR-KG snapshots, policy flags, execution roots, backend settings, and memory namespace. Provider fallback may be used during development, but benchmark, collaborator-case, and bounded autonomous claims are reported only from fixed model versions, fixed prompt templates, fixed registry snapshots, fixed BR-KG snapshots, and fixed evaluation partitions.

Table~\ref{tab:model-ids} lists the resolved model identifiers and the coding-agent surface used for each of the seven agents in the paired benchmarks. All agents were driven through their coding-agent command-line surfaces (not direct chat or completion APIs), so decoding parameters follow each surface's defaults; we did not override temperature for the agent rollouts. The benchmark runs reported here were executed in May--June 2026 (the 60-task tool-routing condition was run on 2026-06-05). The three evidence-support judges were run deterministically at temperature~0 with a fixed random seed (7).

\begin{table}[!htbp]
\centering
\small
\setlength{\tabcolsep}{5pt}
\renewcommand{\arraystretch}{1.15}
\caption{Resolved model identifiers and coding-agent surfaces for the seven benchmarked agents.}
\label{tab:model-ids}
\begin{tabular}{|>{\RaggedRight\arraybackslash}p{0.20\textwidth}|>{\RaggedRight\arraybackslash}p{0.40\textwidth}|>{\RaggedRight\arraybackslash}p{0.30\textwidth}|}
\hline
\rowcolor{brtablehead}\textbf{Paper label} & \textbf{Resolved model identifier} & \textbf{Coding-agent surface} \\
\hline
Claude Opus 4.8 & \texttt{claude-opus-4-8} & Claude Code (\texttt{claude -p}) \\
\hline
Codex GPT-5.5 & \texttt{gpt-5.5} & Codex CLI (\texttt{codex exec}) \\
\hline
Gemini 3.1 Pro & \texttt{google/gemini-3.1-pro-preview} & OpenCode (\texttt{opencode run}) \\
\hline
GLM-5.1 & \texttt{zai-coding-plan/glm-5.1} & OpenCode (\texttt{opencode run}) \\
\hline
DeepSeek-V4-Pro & \texttt{deepseek/deepseek-v4-pro} & OpenCode (\texttt{opencode run}) \\
\hline
Kimi K2.5 & \texttt{opencode/kimi-k2.5} & OpenCode (\texttt{opencode run}) \\
\hline
Qwen3.6-Plus & \texttt{opencode/qwen3.6-plus} & OpenCode (\texttt{opencode run}) \\
\hline
\end{tabular}
\end{table}

\subsection{S1.3. MCP operation surface and registry boundary}\label{supp:s1.3}

The MCP surface is the public control plane, not the complete neuroimaging software catalog. The checked implementation contains 87 decorated MCP operations and 3 decorated MCP resources. These operations span five functional families: discovery and resolution (tools, workflows, datasets, BR-KG, literature); planning and recipe generation; execution observability and artifact inspection; grounding and scientific review; and memory access and report generation. Slurm/\allowbreak{}HPC scheduling is exposed as a separate operation family. The resources expose structured tools, datasets, and workflow lookups. Full capability-family tables, implementation evidence sources, compute-graph diagrams, safety-gate tables, and access-mode tables are reported in Appendices~\hyperref[app:a]{A}, \hyperref[app:d]{D}, \hyperref[app:f]{F}, and~\hyperref[app:i]{I} rather than repeated in the main Supplementary Methods.

The broader registry stores the objects MCP searches, validates, packages, and observes. The distinction is important because an MCP operation is not the same as a workflow specification, registry specification, or executable wrapper.

\begin{center}
\small
\setlength{\tabcolsep}{5pt}
\renewcommand{\arraystretch}{1.15}
\begin{tabular}{|>{\RaggedRight\hspace{0pt}\arraybackslash}p{0.25\textwidth}|>{\RaggedRight\hspace{0pt}\arraybackslash}p{0.28\textwidth}|>{\RaggedRight\hspace{0pt}\arraybackslash}p{0.38\textwidth}|}
\hline
\rowcolor{brtablehead}\textbf{Object} & \textbf{Example} & \textbf{Used for} \\
\hline
MCP operation & resolve\_dataset & Ask the control plane whether a dataset and required assets are usable. \\
\hline
MCP resource & workflow lookup & Retrieve structured workflow or dataset summaries. \\
\hline
Workflow specification & predictive\_modeling\_workflow & Generate a plan and expected artifact list. \\
\hline
Registry specification & fsl\_randomise\_contract & Validate parameters, compatibility, and backend requirements. \\
\hline
Executable wrapper & AFNI/\allowbreak{}FSL/\allowbreak{}Workbench command wrapper \citep{cox_afni_1996,smith_advances_2004,jenkinson_fsl_2012} & Run a backend command through an approved execution substrate. \\
\hline
\end{tabular}
\end{center}

Inventory counts are reported once in \hyperref[supp:s5.1]{S5.1} so that reader attention stays on the control boundary here.

\subsection{S1.4. LLM/MCP decision boundary}\label{supp:s1.4}

The LLM/\allowbreak{}MCP boundary separates interpretation from certification.

\begin{center}
\small
\setlength{\tabcolsep}{5pt}
\renewcommand{\arraystretch}{1.15}
\begin{tabularx}{\textwidth}{|>{\RaggedRight\hspace{0pt}\arraybackslash\bfseries}p{0.20\textwidth}|>{\RaggedRight\hspace{0pt}\arraybackslash}X|>{\RaggedRight\hspace{0pt}\arraybackslash}X|>{\RaggedRight\hspace{0pt}\arraybackslash}p{0.20\textwidth}|}
\hline
\rowcolor{brtablehead}\textbf{Step} & \textbf{LLM or outer client does} & \textbf{MCP/\allowbreak{}backends do} & \textbf{Output} \\
\hline
Intent & Interprets the user request and proposes modalities, datasets, methods, and constraints. & Receives typed query arguments. & Typed planning query. \\
\hline
Discovery & Asks for candidate datasets, tools, workflows, or evidence. & Searches BR-KG, registry, dataset catalog, and literature connectors. & Ranked candidate cards. \\
\hline
Resolution & Chooses among returned options. & Canonicalizes dataset/\allowbreak{}tool IDs and returns metadata, schemas, and readiness status. & Resolved resource/\allowbreak{}tool record. \\
\hline
Feasibility & Requests feasibility checks. & Validates roots, permissions, assets, tool contracts, and active constraints. & pass /\allowbreak{} warn /\allowbreak{} block verdicts. \\
\hline
Recipe & Requests backend-specific instructions. & Packages Python, Neurodesk, container, or Slurm recipe with expected outputs. & Execution recipe and artifact contract. \\
\hline
Execution & Delegates computation to allowed backend or coding harness. & Produces logs, artifacts, metrics, and environment observations. & Run bundle. \\
\hline
Verification & Proposes candidate claims or report language. & Checks artifacts, constraints, scorecards, QC outputs, and claim scope. & Review card and memory eligibility. \\
\hline
\end{tabularx}
\end{center}

The LLM may choose among MCP-returned options, but it cannot invent a registry ID, override a resolver block, bypass a policy gate, or promote a claim to accepted memory without a verification verdict.

\begin{figure}[!htbp]
\centering
\includegraphics[width=\textwidth]{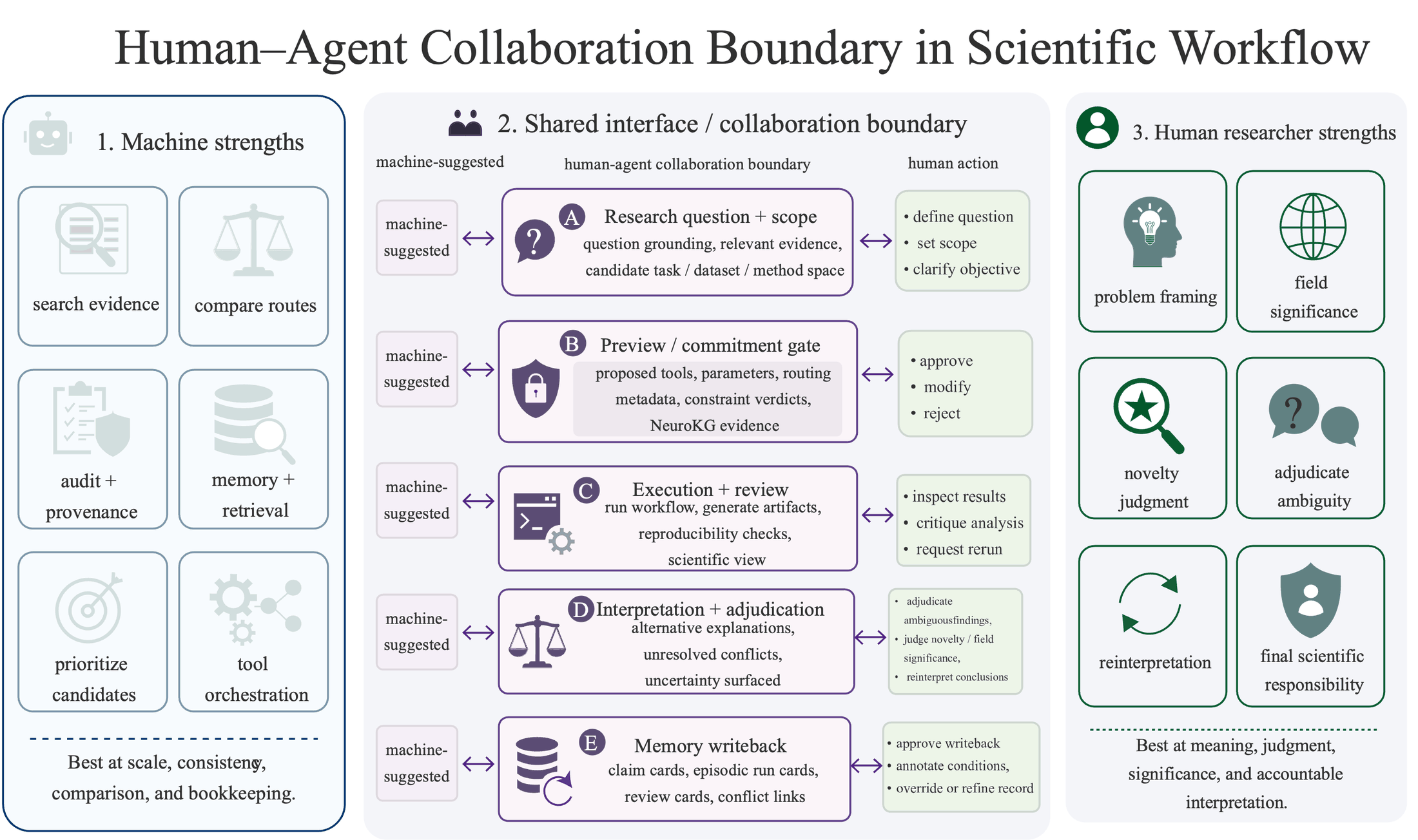}
\caption{\textbf{Human-agent authority boundary in Brain Researcher.} The diagram separates researcher authority, language-model assistance, MCP-mediated system authority, and scientific verification across the episode lifecycle. Brain Researcher can search evidence, compare options, prepare recipes, observe execution, assemble review inputs, and record memory candidates. The researcher retains authority over question framing, admissible tradeoffs, commitment-gate approval, interpretation, authorship, and final claim language. System gates prevent the model from inventing registry identifiers, overriding resource blocks, accepting missing artifacts, or promoting claims without a review verdict.}
\label{fig:human-agent-boundary}
\end{figure}

\subsection{S1.5. Orchestration, subagents, and policy gates}\label{supp:s1.5}

The orchestration service manages run creation, polling, cancellation, retry, progress tracking, streaming logs, event updates, and durable state. Generated artifacts are stored under run-specific roots with manifests and checksums. The implementation may use specialized subagents, but each has a bounded role.

\begin{center}
\small
\setlength{\tabcolsep}{5pt}
\renewcommand{\arraystretch}{1.15}
\begin{tabular}{|>{\RaggedRight\hspace{0pt}\arraybackslash}p{0.18\textwidth}|>{\RaggedRight\hspace{0pt}\arraybackslash}p{0.27\textwidth}|>{\RaggedRight\hspace{0pt}\arraybackslash}p{0.24\textwidth}|>{\RaggedRight\hspace{0pt}\arraybackslash}p{0.22\textwidth}|}
\hline
\rowcolor{brtablehead}\textbf{Component} & \textbf{Input} & \textbf{Output} & \textbf{Boundary} \\
\hline
Critic & Candidate plan, review findings, cycle record. & approve /\allowbreak{} revise /\allowbreak{} block /\allowbreak{} terminate. & Cannot certify artifacts without run-bundle evidence. \\
\hline
Recovery agent & Failure logs, missing outputs, backend errors. & Retry or substitution proposal. & Cannot erase the original failure record. \\
\hline
Provenance tracker & Decisions, tool calls, parameter bindings. & Event trace and trajectory entries. & Records actions; does not judge scientific validity. \\
\hline
Router & Task type and policy context. & Model/\allowbreak{}backend/\allowbreak{}subagent route. & Routing is recorded in the run card. \\
\hline
Supervisor & Open caveats, validation progress, budget. & Next branch proposal. & Cannot expand beyond the declared design space. \\
\hline
\end{tabular}
\end{center}

Execution is controlled by allowed filesystem roots, network flags, dangerous-tool flags, direct-execution flags, timeout budgets, authentication checks, and approval phrases for pipeline execution. Heavy neuroimaging work is normally performed outside the MCP service by local Python, Neurodesk, a generic container runtime, Slurm/\allowbreak{}HPC, Kubernetes, AWS Batch, or another configured backend. Direct MCP execution exists only as a gated administrative path and is disabled by default in the evaluated configuration. Adaptive optimization surfaces such as contextual bandits, offline reinforcement learning, and drift detection are engineering surfaces only; they are not treated as evaluated capabilities unless explicitly declared in the evaluation card.

\section{S2. Research episode runtime and run states}\label{supp:s2}

\subsection{S2.1. Episode object and persisted state}\label{supp:s2.1}

The research episode is the core runtime unit. It persists the scientific question, planning state, evidence state, candidate alternatives, selected plan, selected tools, parameter bindings, resource-resolution results, execution outputs, QC artifacts, review verdicts, recovery attempts, limitation notes, and memory-writeback state. This prevents the system from treating later turns as isolated prompts: later decisions can depend on prior evidence, failed attempts, unresolved caveats, resource blockers, and review outcomes.

Episode record fields are grouped by role:

\begin{center}
\small
\setlength{\tabcolsep}{5pt}
\renewcommand{\arraystretch}{1.15}
\begin{tabular}{|>{\RaggedRight\hspace{0pt}\arraybackslash}p{0.26\textwidth}|>{\RaggedRight\hspace{0pt}\arraybackslash}p{0.65\textwidth}|}
\hline
\rowcolor{brtablehead}\textbf{Group} & \textbf{Fields} \\
\hline
Question state & input question, normalized entities, scientific scope, intended output type. \\
\hline
Resource state & data requirements, candidate datasets, resolved resources, missing assets, access class. \\
\hline
Planning state & evidence bundle, candidate workflows, candidate tools, active constraints, selected plan. \\
\hline
Execution state & backend recipe, expected artifacts, produced artifacts, logs, QC outputs, recovery attempts. \\
\hline
Review state & review cards, rule findings, claim candidates, caveats, verdicts. \\
\hline
Memory state & memory eligibility, claim-card status, relation events, final completion status. \\
\hline
\end{tabular}
\end{center}

Example episode-record excerpt:
\begin{suppcode}
episode_id: ep_hcp_predict_001
state: planning_blocked
question_type: prediction
selected_dataset: HCP-YA
candidate_workflow: predictive_modeling
required_assets:
  - feature_matrix
  - target_variable
  - covariate_table
  - fold_manifest
resource_verdict: block
blocker: fold_manifest_missing
next_action: resolve_asset_or_choose_alternative_workflow
\end{suppcode}

\subsection{S2.2. Scientific stages}\label{supp:s2.2}

At the scientific level, an episode moves through stages that each produce a concrete object.

\begin{center}
\small
\setlength{\tabcolsep}{5pt}
\renewcommand{\arraystretch}{1.15}
\begin{tabular}{|>{\RaggedRight\hspace{0pt}\arraybackslash}p{0.22\textwidth}|>{\RaggedRight\hspace{0pt}\arraybackslash}p{0.48\textwidth}|>{\RaggedRight\hspace{0pt}\arraybackslash}p{0.21\textwidth}|}
\hline
\rowcolor{brtablehead}\textbf{Stage} & \textbf{What happens} & \textbf{Concrete output} \\
\hline
Question framing & Natural-language goal is converted into typed scientific entities and admissible outputs. & Typed question record. \\
\hline
Evidence assembly & BR-KG, literature, dataset catalogues, tool registries, and meta-analysis stores are queried. & Evidence bundle. \\
\hline
Option-set formation & Candidate datasets, tools, workflows, and parameterizations are identified. & Candidate ledger. \\
\hline
Constraint compilation & Method-condition records and tool contracts are turned into active checks. & Active constraint set. \\
\hline
Commitment & Human researcher, critic, or evaluation harness approves/\allowbreak{}revises/\allowbreak{}blocks the plan. & Commitment-gate record. \\
\hline
Execution & Approved recipe is run under the selected backend. & Run bundle and artifact manifest. \\
\hline
Verification & Artifacts, logs, QC, scorecards, and candidate claims are checked. & Review card. \\
\hline
Memory writeback & Accepted or qualified claims are stored with provenance; rejected claims remain in the run bundle. & Claim card or no-write decision. \\
\hline
\end{tabular}
\end{center}

\subsection{S2.3. Running example: one predictive-modeling episode}\label{supp:s2.3}

The running example used throughout this supplement is an HCP-YA predictive-modeling replay \citep{van_essen_wu-minn_2013,wu-minn_human_connectome_project_consortium_hcp_2017}. The question is whether a behavioral target can be predicted from Schaefer-100 x 7 features \citep{schaefer_local-global_2018} using a fixed subject intersection. The resource card specifies N = 326, five Liu components \citep{liu_benchmarking_2025}, a pyspi statistic catalogue, feature matrices, candidate target variables, subject-intersection logic, missingness rules, and backend reachability. A valid run requires a feature matrix, target vector, covariates, and a fold manifest. During resolution, the dataset can pass dataset-level readiness while failing asset-level readiness if the fold manifest or target variable is missing. The constraint compiler then creates a hard fold-manifest check and soft warnings for confound specification or controversial preprocessing choices. If the fold manifest is absent, the commitment gate returns block before execution. If all required assets pass, BR-MCP packages a recipe and expected artifacts: predictions, scorecard, model metadata, fold manifest, and artifact manifest. Review then checks leakage, grouped cross-validation, permutation or null controls, robustness probes, and claim language. Only an accepted or explicitly qualified result can become a claim card.

\subsection{S2.4. Runtime states, checkpointing, and recovery}\label{supp:s2.4}

Scientific stages are conceptual. Runtime states are the system state machine used to resume or audit the episode.

\begin{center}
\small
\setlength{\tabcolsep}{5pt}
\renewcommand{\arraystretch}{1.15}
\begin{tabular}{|>{\RaggedRight\hspace{0pt}\arraybackslash}p{0.25\textwidth}|>{\RaggedRight\hspace{0pt}\arraybackslash}p{0.66\textwidth}|}
\hline
\rowcolor{brtablehead}\textbf{Runtime state} & \textbf{Meaning} \\
\hline
initialization & Create episode record, run root, policy context, and memory namespace. \\
\hline
planning & Populate candidates, evidence, constraints, and recipe drafts. \\
\hline
execution & Dispatch or delegate computation. \\
\hline
review & Inspect artifacts, logs, QC outputs, scorecards, and claims. \\
\hline
recovery & Handle missing outputs, failed commands, invalid artifacts, or blocked constraints. \\
\hline
completion & Record accepted, qualified, or non-acceptance endpoint. \\
\hline
blocked & A known hard constraint or missing resource prevents valid progress. \\
\hline
terminated & Budget, user action, critic decision, or unrecovered infrastructure failure stops the episode. \\
\hline
\end{tabular}
\end{center}

A missing fold manifest is a blocked state: the reason is known and can be addressed by resolving the asset or changing workflow. Budget exhaustion after repeated unresolved failures is a terminated state. Checkpoints store the current state, selected candidates, active constraints, resource-resolution outputs, branch history, and run-bundle pointers. Recovery attempts are appended rather than overwritten, so a repaired run still preserves the original failure and repair path.

\section{S3. BR-KG construction, ONVOC normalization, and evidence assembly}\label{supp:s3}

\subsection{S3.1. BR-KG role and snapshot metadata}\label{supp:s3.1}

BR-KG is a Neo4j knowledge graph that links scientific records (publications, activation coordinates, statistical maps), cognitive structure (tasks, concepts, brain regions), resources (datasets, tools), embeddings, and operational records (review verdicts, governance entries, run objects). Its role is retrieval and provenance-aware traversal, not scientific proof. It provides source linking, coverage awareness, negative knowledge, method-condition records, and graph paths that can be inspected by planners and reviewers.

In the checked release snapshot (2026-07-07), BR-KG contains 745,949 nodes and 2,461,469 relationships. The manuscript-facing schema defines 27 primary node labels and 34 primary relationship types. Production snapshots may contain additional compatibility, migration, enrichment, operational, or retrieval labels (this snapshot carries 108 active node labels and 125 active relationship types), so primary-schema counts and all-active-label counts are reported separately in the \hyperref[app:data-cards]{Appendix/\allowbreak{}Data Cards}. Deployment metadata such as Neo4j 5.20.0 Community, one ready K3s BR-KG pod, 139 indexes, and 48 constraints is release-readiness metadata, not evidence for scientific correctness.

Example traversal role:

\begin{suppcode}
query: "N-back working memory activation"
normalized_task: ONVOC:N_BACK
linked_concepts:
  - working_memory
retrieved_records:
  - publications
  - task contrasts
  - activation coordinates
  - brain regions
  - method constraints
used_by:
  - evidence_bundle
  - planner
  - reviewer
\end{suppcode}

\subsection{S3.2. Source governance, ingestion, and validation}\label{supp:s3.2}

Graph ingestion separates accepted records from candidate records. Accepted records are source-backed objects such as publications, coordinates, maps, task records, dataset records, tool descriptors, anatomical regions, and curated method-condition records. Candidate records may come from automated extraction, KGGEN, GABRIEL, or real-time retrieval. They remain in candidate lanes until they pass schema validation, source marking, provenance recording, license/\allowbreak{}citation checks, and acceptance rules.

\begin{center}
\small
\setlength{\tabcolsep}{5pt}
\renewcommand{\arraystretch}{1.15}
\begin{tabular}{|>{\RaggedRight\hspace{0pt}\arraybackslash}p{0.25\textwidth}|>{\RaggedRight\hspace{0pt}\arraybackslash}p{0.39\textwidth}|>{\RaggedRight\hspace{0pt}\arraybackslash}p{0.27\textwidth}|}
\hline
\rowcolor{brtablehead}\textbf{Record type} & \textbf{Example} & \textbf{Status before use as graph fact} \\
\hline
Accepted record & Publication with DOI, source loader, snapshot date, and citation. & Usable as graph evidence. \\
\hline
Candidate record & Extracted task-region relation from KGGEN. & Candidate only until validated and accepted. \\
\hline
Held or rejected record & Unsupported generated relation or source-incomplete extraction. & Not promoted to graph fact. \\
\hline
\end{tabular}
\end{center}

Each releasable source row records snapshot date, source URL, license, required citation, loader path, loader version, command line, configuration hash, input artifact path, output manifest, artifact path, and release disposition. Source aliases are normalized, source-like missing values are backfilled or removed, and source-specific licenses and citations are pinned before manuscript claims are made.

\paragraph{Manual record-quality audit.}
We manually reviewed 400 randomly sampled BR-KG records: 200 nodes and 200 directed edges. We classified 280 as correct, 45 as incomplete or uncertain, 53 as pre-existing source-integration errors, and 22 as BR-side errors. The three non-correct categories comprised:
\begin{itemize}[leftmargin=1.5em,itemsep=2pt,topsep=3pt]
    \item \textbf{Incomplete or uncertain ($n=45$):} coordinate-space metadata was incomplete for 23 map nodes: 14 FitLins statistical maps had a known template but stored \texttt{space=unknown}; five NeuroVault maps lacked a materialized canonical space despite \texttt{GenericMNI} target-template metadata; and four additional NeuroVault maps did not canonicalize \texttt{target\_template\_image=GenericMNI} into the KG space field. Available evidence was insufficient to assess 21 records: nine \texttt{HAS\_COORDINATE} edges had an unresolved source coordinate frame; seven Neurosynth coordinate nodes had source metadata marked \texttt{UNKNOWN} while the loader assigned \texttt{MNI}; three \texttt{EvidenceSpan} nodes had insufficient frozen content and live-source snapshot drift; and two GABRIEL \texttt{MeasurementRun} nodes lacked a matching fixed upstream run record. One Neurobagel \texttt{Dataset} node had a name containing an unresolved literal backslash-n formatting artifact.
    \item \textbf{Pre-existing source-integration errors ($n=53$):} 37 Neurosynth records in which TAL coordinates were materialized as MNI, and 16 NeuroStore records in which \texttt{StudyObjective} was materialized in publication or collection title fields. These failures arose in pre-existing source-specific integration paths and do not imply that the raw Neurosynth or NeuroStore records were incorrect.
    \item \textbf{BR-side errors ($n=22$):} five NiCLIP index-offset errors; five dataset-to-contrast crosswalk over-propagations; three configuration-derived target mismatches; five GABRIEL/BR claim, evidence, or cohort-extraction errors; one unsupported manual contrast mapping; two tool/package registry default errors; and one mixed duplicate-DOI/BR merge error.
\end{itemize}

We are extending this audit into a systematic review across the full BR-KG and will iteratively repair and re-audit the ingestion and normalization issues it identifies; the counts above describe the audited snapshot rather than a completed post-remediation release.

\subsection{S3.3. Entity resolution, relationship strength, and semantic linking}\label{supp:s3.3}

Entities are reconciled through exact identifiers, multi-attribute matching, fuzzy string matching, embedding-based alignment, and curated aliases. Matching is applied across publications, datasets, regions, concepts, paradigms, institutions, and researchers. Matched records are linked with deduplication edges that retain matching method and confidence.

Example entity-resolution excerpt:
\begin{suppcode}
raw_labels:
  - "N-back"
  - "WM_BACK"
  - "verbal working memory"
resolved_task_family: N_BACK
non_equivalent_fields:
  - contrast_definition
  - modality
  - stimulus_type
  - population
resolution_note: same task family, not interchangeable experimental objects
\end{suppcode}

Relationship strength may combine literature count, coordinate evidence, spatial overlap, embedding similarity, spatial distance, and optional user feedback. These scores affect retrieval ranking and path prioritization only. They are not review acceptance criteria and cannot by themselves establish a scientific claim.

\subsection{S3.4. ONVOC ontology-linker layer}\label{supp:s3.4}

ONVOC normalizes heterogeneous source vocabulary before graph traversal. Free-text task names, contrast labels, cognitive constructs, anatomical labels, dataset-specific aliases, and source-specific terms are mapped to typed ONVOC records. OnvocClass nodes represent normalized vocabulary classes; OntologyConcept nodes represent linked external or internal concepts; IN\_ONVOC edges connect graph entities to normalized vocabulary entries.

Example ONVOC output:

\begin{suppcode}
input_terms:
  - "N-back"
  - "WM_BACK"
  - "verbal working memory"
onvoc_task_family: N_BACK
linked_concepts:
  - working_memory
linked_graph_nodes:
  - CognitiveTask:N_BACK
  - CognitiveConcept:working_memory
non_equivalent_fields:
  - contrast
  - modality
  - coordinate_space
  - population
\end{suppcode}

ONVOC may link terms to the same task family, but it does not collapse contrast definitions, modalities, populations, or coordinate systems. When direct higher-tier graph evidence is sparse, ONVOC still provides an entity-typing and literature-query backbone. The evidence bundle records whether a decision was supported by accepted graph evidence, validated batch extraction, candidate evidence, or real-time retrieval.

\subsection{S3.5. Evidence aggregation and connector-failure reporting}\label{supp:s3.5}

\begin{figure}[!htbp]
\centering
\includegraphics[width=\textwidth]{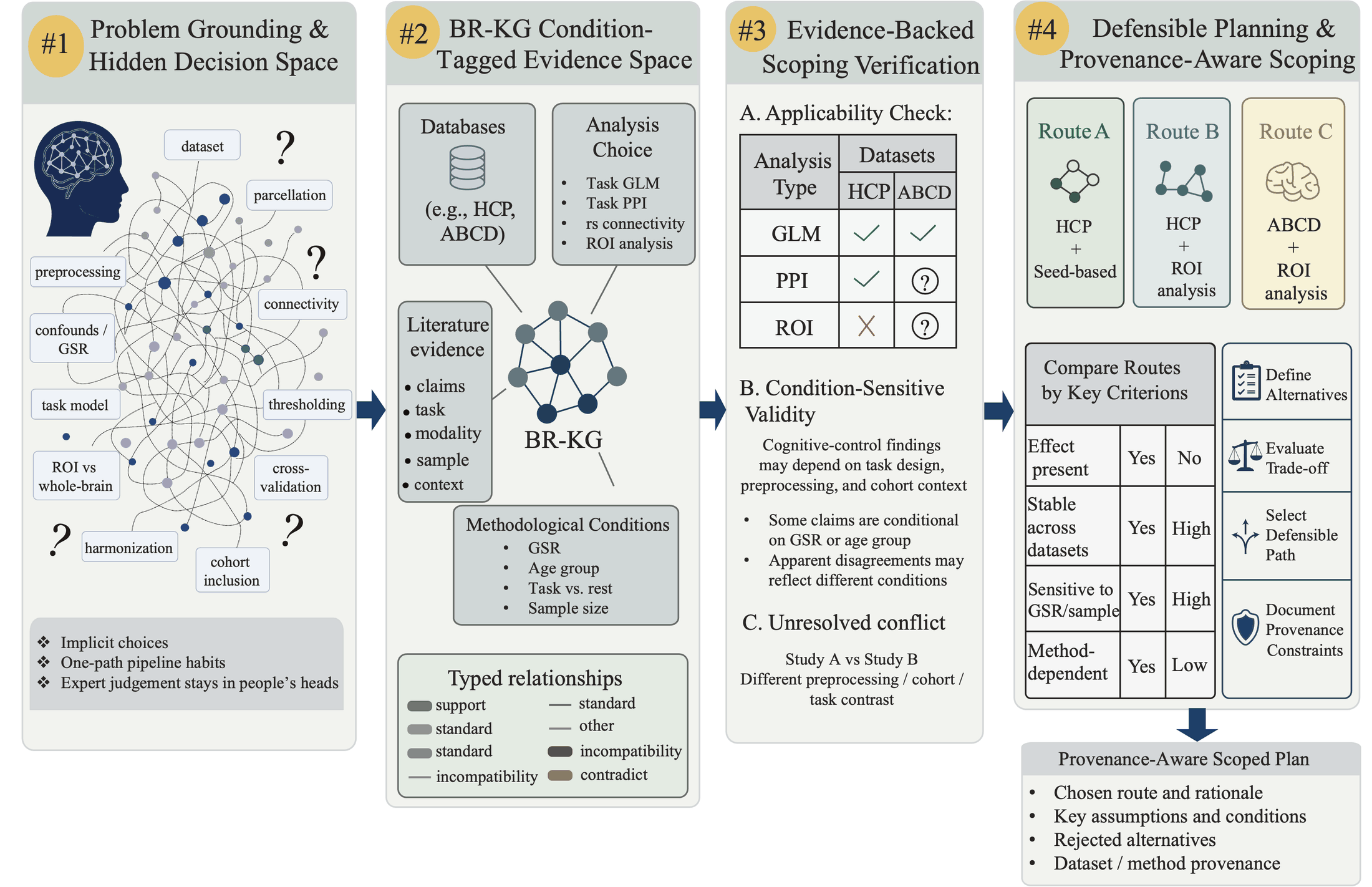}
\caption{\textbf{Condition-tagged evidence for scoped planning.} The workflow begins with an underspecified scientific request and expands it into entities, methods, datasets, assumptions, and validity conditions that can be checked before execution. BR-KG links literature evidence to those conditions with provenance, while resource and tool resolvers determine which conditions are satisfied, missing, controversial, or incompatible in the current episode. Applicability checks convert the resulting condition set into admissible plans, required sensitivity analyses, caveats, or blockers. The purpose is to expose methodological scope before a run is committed and before a positive or negative result can anchor the narrative.}
\label{fig:conditional-evidence}
\end{figure}

Evidence assembly produces an evidence bundle for each episode. A single bundle may include BR-KG paths, literature results, dataset metadata, tool-registry hits, known failure modes, and review-checklist triggers. Connector failures are stored rather than silently discarded. Figure~\ref{fig:conditional-evidence} shows how a recommendation's validity conditions become applicability checks and then admissible plans, caveats, or blockers, so that it arrives with its limits attached.

Example evidence-bundle excerpt:
\begin{suppcode}
evidence_bundle:
  input_query: "predict behavior from HCP-YA Schaefer-100 x 7 features"
  resolved_entities:
    dataset: HCP-YA
    feature_space: Schaefer100x7
    task_family: null
  graph_evidence:
    tier: accepted_graph_records
    paths: [dataset_to_feature_record, method_to_constraint]
  real_time_retrieval:
    tier: literature_gap_fill
    status: completed
  connector_failures:
    - source: PubMed
      reason: timeout
      downstream_effect: evidence_coverage_caveat
  tool_registry_hits:
    - predictive_modeling_family
  method_condition_records:
    - fold_manifest_required
    - leakage_standardization_inside_cv
\end{suppcode}

A missing connector result is treated as reduced evidence coverage. It may trigger a caveat or reviewer warning, but it is not treated as evidence that no relevant literature or resource exists.

\subsection{S3.6. Embedding lanes}\label{supp:s3.6}

Brain Researcher uses separate embedding lanes for different retrieval surfaces. Tool retrieval uses titles, descriptions, command names, and tags. Brain-text retrieval uses task, construct, region, and claim-memory text. BR-KG may also store source-specific embeddings as graph nodes or node properties for selected ingestion lanes. Lane name, model identity, vector dimension, input fields, usage surface, snapshot, and index state are recorded in Appendices~\hyperref[app:b]{B} and~\hyperref[app:d]{D}.

Embedding similarity retrieves and ranks candidates. It cannot by itself support a scientific claim. The current system operates in text-only concept-matching and retrieval modes. Cross-modal alignment between fMRI activation patterns and ontology concepts is a planned extension and is not used in reported results.

\subsection{S3.7. KG extension pipelines and views}\label{supp:s3.7}

Automated extension pipelines maintain BR-KG as a living graph. KGGEN processes full-text publications into typed candidate triples with provenance. GABRIEL converts qualitative domain corpora such as review articles, textbook descriptions, and clinical practice guidelines into schema-compliant quantitative or semi-structured records. On-the-fly retrieval fills query-time coverage gaps when graph evidence is sparse. In all channels, generated records remain candidate or comparison records until explicitly validated, source-marked, and accepted.

Focused task-family or disease-cohort views are subgraph projections over the same underlying records, not separate graphs. They may organize working memory, attention, executive function, language, perception, ADHD, MDD, schizophrenia, healthy aging, and related records for retrieval. View membership supports navigation; it does not change the acceptance status of a finding.

\subsection{S3.8. BR-KG visual atlas and release snapshot.}\label{supp:s3.8}

The BR-KG snapshot used in this work is summarized in the \hyperref[app:data-cards]{Appendix/\allowbreak{}Data Cards}. The atlas reports graph composition, dominant relationship families, schema topology, representative literature/\allowbreak{}map/\allowbreak{}task paths, source coverage, and release-readiness checks. These figures are used to document the retrieval substrate available to Brain Researcher; they are not treated as evidence that any scientific claim is true. During an episode, BR-KG supports source-backed retrieval, task/\allowbreak{}construct normalization, dataset/\allowbreak{}tool linking, method-condition lookup, and provenance inspection. Detailed counts, figure panels, source-like values, orphan-node rates, edge-confidence coverage, and release blockers are reported in \hyperref[app:b]{Appendix B} rather than repeated in the main Supplementary Methods.

\begin{figure}[!htbp]
\centering
\includegraphics[width=.72\textwidth]{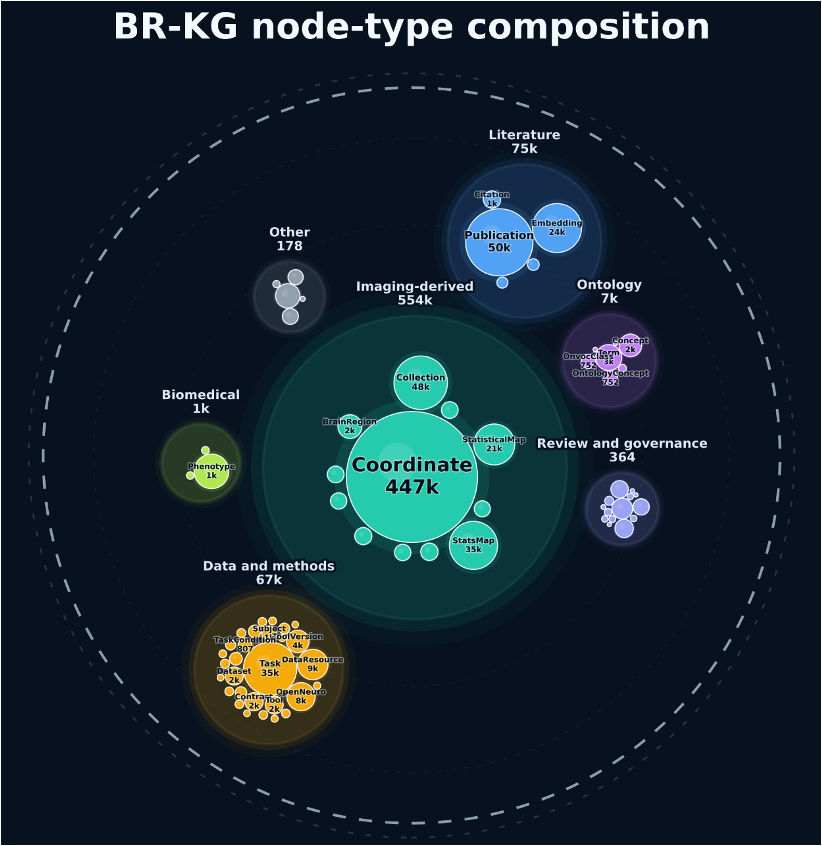}
\caption{\textbf{BR-KG node-type composition.} Node composition summarizes the object families present in the BR-KG snapshot used for retrieval, normalization, and planning support. The chart separates scientific entities such as publications, coordinates, statistical maps, tasks, contrasts, concepts, and brain regions from resource and workflow entities such as datasets, tools, methods, governance records, and prior reviewed claims. Large node families indicate coverage available for traversal and grounding, not evidentiary weight for any individual claim. Detailed counts, release metadata, and source-quality checks are reported in Appendix B.}
\label{fig:br-kg-node-types}
\end{figure}

\begin{figure}[!htbp]
\centering
\includegraphics[width=.72\textwidth]{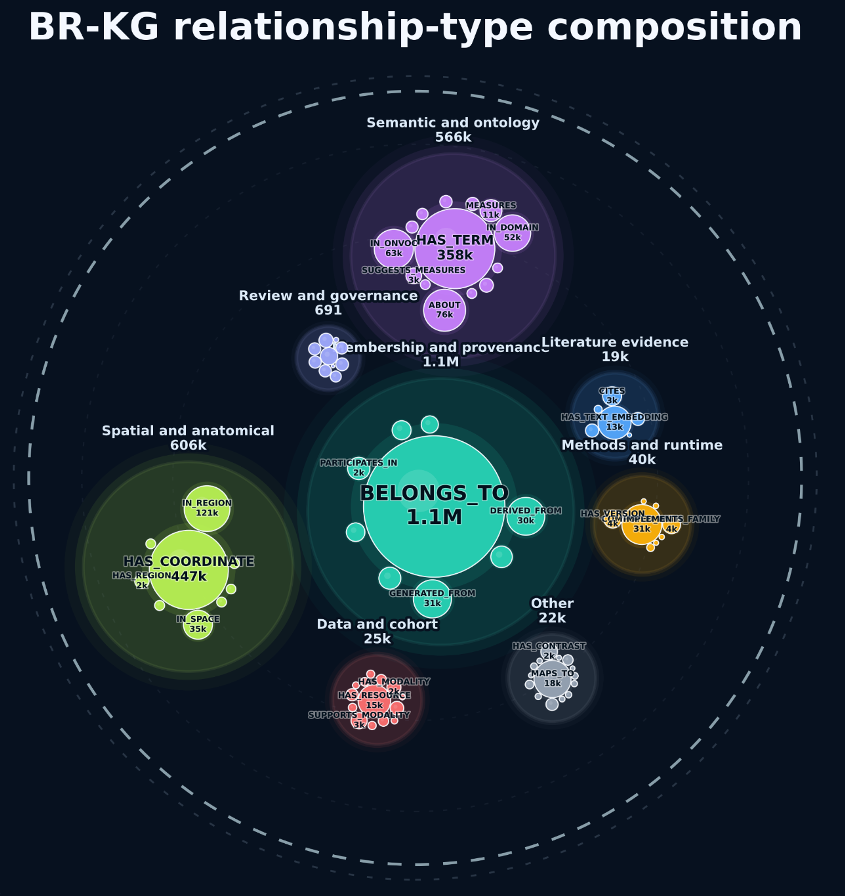}
\caption{\textbf{BR-KG relationship-type composition.} Relationship composition summarizes the edge families that connect publications, coordinates, maps, tasks, contrasts, concepts, regions, datasets, tools, constraints, and provenance records in the BR-KG snapshot. High-count relationship classes define the main traversal surfaces used for evidence assembly, task-to-construct normalization, map lookup, dataset discovery, and tool compatibility checks. Edge frequencies document graph structure and retrieval affordances; support strength for a scientific claim requires source provenance and review context. Appendix B provides the expanded relationship inventory and release-readiness checks.}
\label{fig:br-kg-edge-types}
\end{figure}

\begin{figure}[!htbp]
\centering
\includegraphics[width=\textwidth]{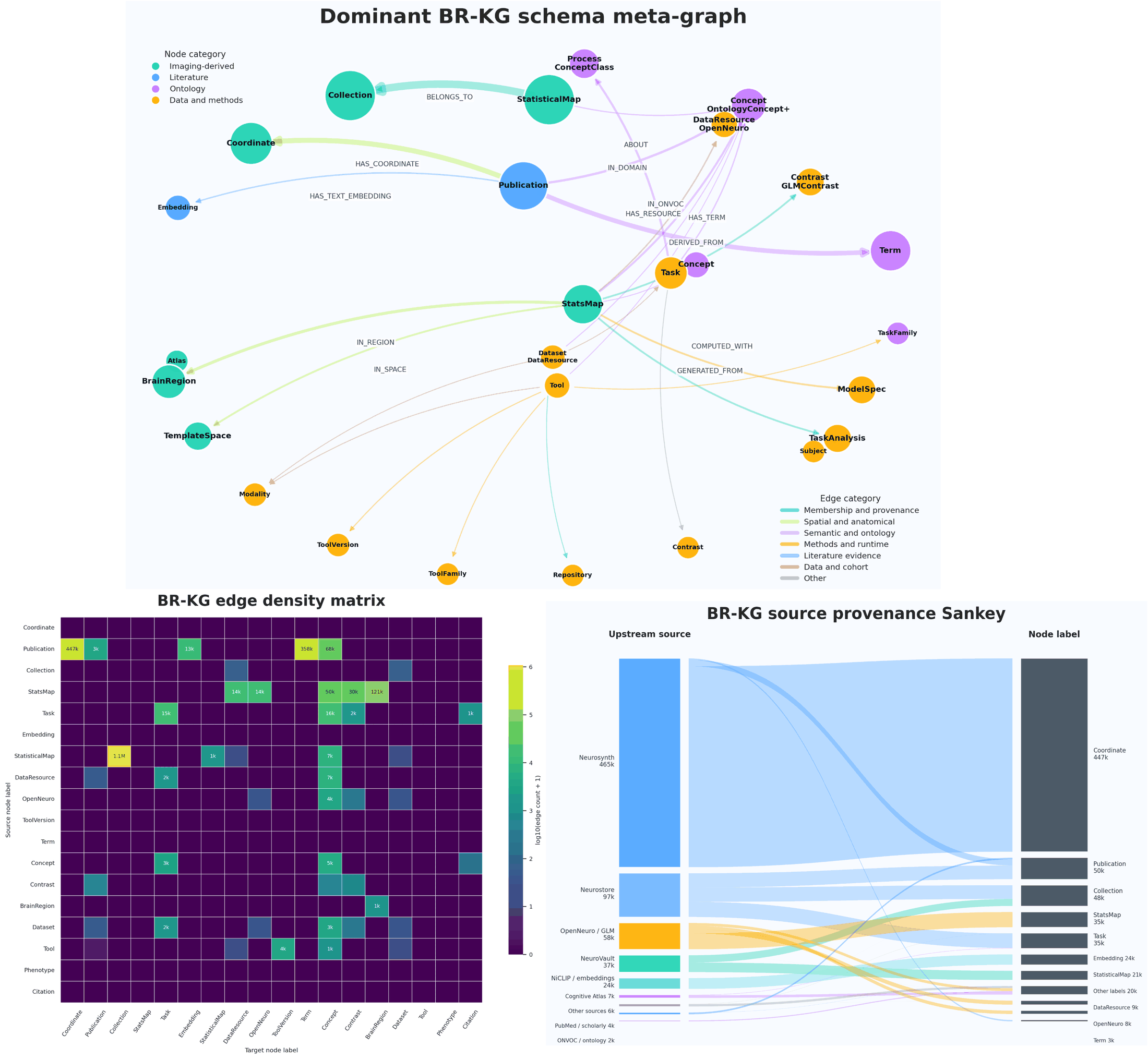}
\caption{\textbf{BR-KG schema topology and provenance flow.} The atlas view summarizes how major node classes are connected, which relationship-label pairs dominate the schema, and how source records feed derived graph objects. Dense schema paths identify the routes most often available for structured retrieval, such as publication-to-coordinate-to-region, task-to-construct-to-map, and dataset-to-tool-to-workflow traversals. Provenance flow indicates where source systems, loaders, and curation records enter the graph. The figure documents the substrate available for multi-hop planning and evidence lookup; inferential claims require episode-level review.}
\label{fig:br-kg-meta-graph}
\end{figure}

\begin{figure}[!htbp]
\centering
\includegraphics[width=\textwidth]{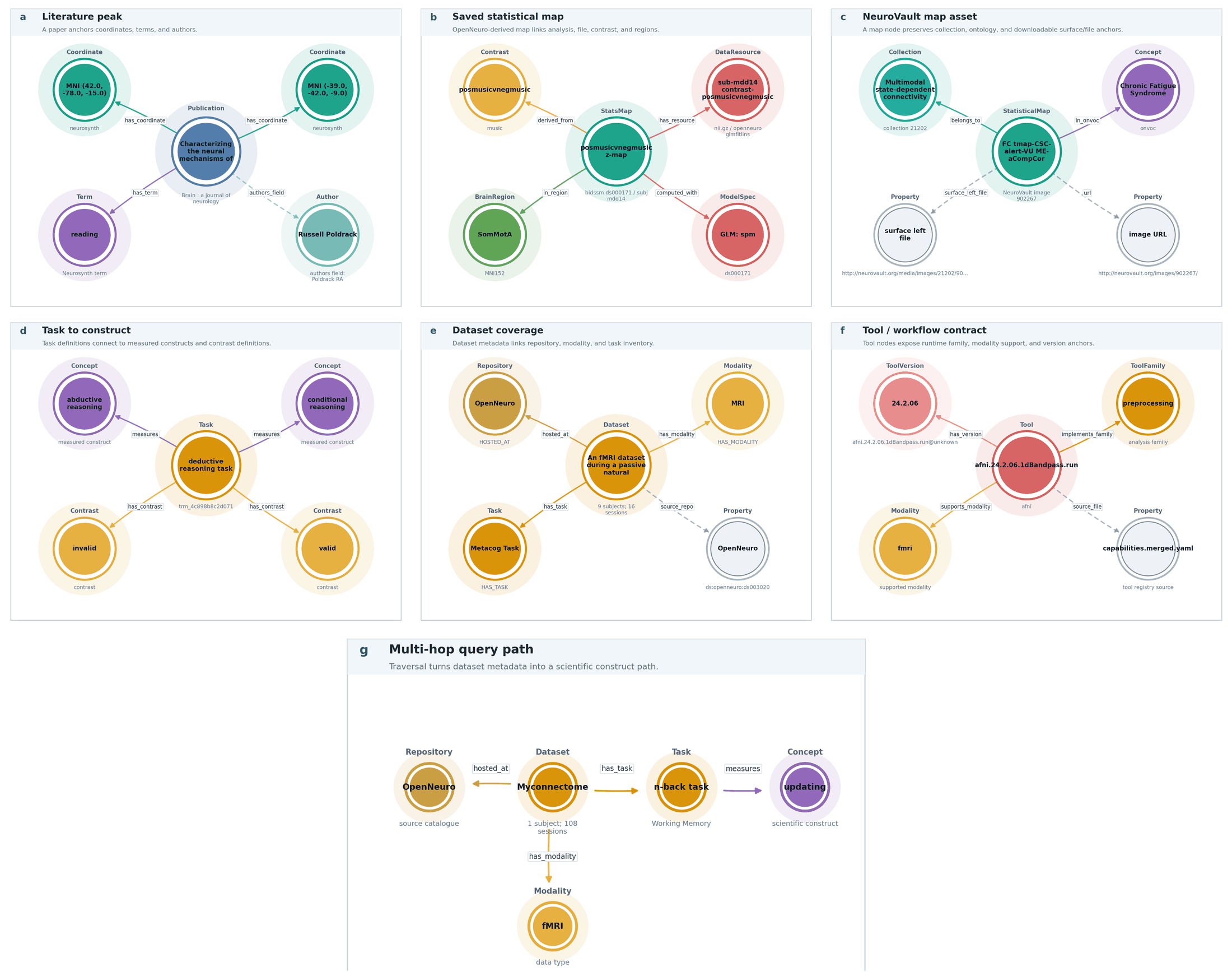}
\caption{\textbf{Representative BR-KG evidence and workflow paths.} Example paths illustrate the kinds of traversals available during an episode. Literature peaks can be linked through coordinates, brain regions, task contrasts, and cognitive constructs; saved maps and NeuroVault assets can be linked to provenance and reuse constraints; datasets can be linked to derivative readiness, phenotype availability, and compatible tool or workflow contracts. These paths show how BR-KG turns text-level mentions into typed objects that can be queried, checked, and passed to the registry or review layer. Scientific acceptance occurs later, after execution evidence and review are available.}
\label{fig:br-kg-examples}
\end{figure}

\section{S4. Dataset and resource resolution}\label{supp:s4}

\subsection{S4.1. Dataset-level and asset-level resolution}\label{supp:s4.1}

Dataset and resource resolution converts abstract dataset references into executable resources. Dataset-level resolution maps identifiers, aliases, catalogue entries, local paths, BIDS roots, derivative directories, remote URLs, access class, phenotype availability, license/\allowbreak{}governance notes, and backend reachability. Asset-level discovery locates the concrete files needed by a workflow: derivatives, covariate files, target files, masks, matrices, parcellations, connectivity matrices, score files, fold manifests, and QC outputs.

A dataset can be known, accessible, and BIDS-valid while still blocking a predictive workflow because the target variable, covariates, or fold manifest are missing. Resource resolution is therefore a precondition for plan commitment, not a background annotation.

Example resolver output:

\begin{suppcode}
resolver_output:
  dataset_id: HCP-YA
  dataset_readiness: pass
  access_class: authorized_local_or_cached
  backend_reachable: true
  workflow: predictive_modeling
  asset_readiness: block
  missing_assets:
    - fold_manifest
  warnings:
    - confound_specification_incomplete
  commitment_verdict: block
\end{suppcode}

Resolver verdicts are reported as pass, warn, block, inaccessible, or incomplete. Pass allows planning to proceed. Warn allows planning but requires caveats or sensitivity actions. Block prevents commitment. Inaccessible and incomplete identify access or missingness states that must be resolved or reported as non-executable outcomes.

\subsection{S4.2. BIDS, OpenNeuro, and derivative discovery}\label{supp:s4.2}

BIDS and OpenNeuro discovery \citep{gorgolewski_brain_2016,poldrack_past_2024,markiewicz_openneuro_2021} proceeds as a concrete resolution chain:

\begin{center}
\small
\setlength{\tabcolsep}{5pt}
\renewcommand{\arraystretch}{1.15}
\begin{tabular}{|>{\RaggedRight\hspace{0pt}\arraybackslash}p{0.12\textwidth}|>{\RaggedRight\hspace{0pt}\arraybackslash}p{0.79\textwidth}|}
\hline
\rowcolor{brtablehead}\textbf{Step} & \textbf{Resolution action} \\
\hline
1 & Resolve the dataset identifier, alias, or catalogue entry. \\
\hline
2 & Check local BIDS root, derivative root, access class, license, and governance notes. \\
\hline
3 & If direct file listing is unavailable, query BIDS entities and dataset metadata. \\
\hline
4 & Parse OpenNeuro GLM specification files when present, including task names, contrasts, condition weights, and contrast-to-concept mappings. \\
\hline
5 & Discover workflow-specific assets such as derivatives, phenotype columns, target variables, masks, parcellations, matrices, and fold manifests. \\
\hline
6 & If resources are remote-only, mark download-required or non-executable according to policy. \\
\hline
7 & Return dataset-level and asset-level readiness separately. \\
\hline
\end{tabular}
\end{center}

Manual contrast-to-concept mappings link common GLM contrasts to cognitive concepts with confidence annotations. These mappings support evidence assembly and ONVOC normalization but remain separate from review acceptance.

\section{S5. Typed tool registry and hierarchical action space}\label{supp:s5}

\subsection{S5.1. Inventory levels and registry objects}\label{supp:s5.1}

The Brain Researcher action space is larger than the MCP surface. MCP exposes controlled operations, while the registry behind MCP stores executable components, workflow specifications, parameter schemas, compatibility clauses, backend recipes, and family cards. The checked inventory distinguishes four levels.

\begin{center}
\small
\setlength{\tabcolsep}{5pt}
\renewcommand{\arraystretch}{1.15}
\begin{tabular}{|>{\RaggedRight\hspace{0pt}\arraybackslash}p{0.2\textwidth}|>{\RaggedRight\hspace{0pt}\arraybackslash}p{0.23\textwidth}|>{\RaggedRight\hspace{0pt}\arraybackslash}p{0.24\textwidth}|>{\RaggedRight\hspace{0pt}\arraybackslash}p{0.24\textwidth}|}
\hline
\rowcolor{brtablehead}\textbf{Level} & \textbf{Count} & \textbf{Object type} & \textbf{Example role} \\
\hline
MCP surface & 87 operations & Public typed endpoints and structured lookups. & Search tools, resolve datasets, inspect artifacts, request review. \\
\hline
Exposed workflow layer & 177 exposed-plus-workflow specifications & MCP-facing and workflow specs. & Plan and package a predictive-modeling recipe. \\
\hline
Broader neuroimaging registry & 2,089 registry specifications & Compatibility, parameter, backend, and resource records. & Validate that a tool accepts the resolved input and produces expected outputs. \\
\hline
Executable-wrapper inventory & approx. 196 Python-native tools + approx. 2,100 command-line wrappers & Callable tools, commands, modules, or scripts. & Run AFNI, ANTs, C3D, dcm2niix, FastSurfer, FreeSurfer, FSL, Greedy, MRtrix, MRtrix3Tissue, NiftyReg, or Workbench commands through an approved backend \citep{cox_afni_1996,avants_reproducible_2011,henschel_fastsurferfast_2020,fischl_freesurfer_2012,jenkinson_fsl_2012,tournier_mrtrix3_2019}. \\
\hline
\end{tabular}
\end{center}

Throughout this supplement, MCP operations refer to endpoints exposed by the MCP server. Executable components refer to Python tools, command-line wrappers, Neurodesk modules, container commands, or scheduler scripts. Registry specifications refer to machine-readable records used for discovery, compatibility checking, parameter validation, and recipe generation.

\subsection{S5.2. Intent extraction and resource categories}\label{supp:s5.2}

Intent extraction maps a user request into planning fields: analysis intent, modality, expected output type, required resource types, and candidate method family.

Example intent record:

\begin{suppcode}
user_request: "Can we predict cognition from resting-state connectivity?"
intent: prediction
modality: fMRI_connectivity
required_resources:
  - connectivity_matrix
  - target_vector
  - covariate_table
  - fold_manifest
expected_outputs:
  - predictions
  - scorecard
  - model_metadata
candidate_family: predictive_modeling
\end{suppcode}

The system defines approximately sixty resource categories, including BIDS roots, 3D volumes, 4D time series, masks, event files, covariate tables, matrices, connectivity matrices, coordinates, statistical maps, QC reports, and rendered figures. A candidate tool is compatible only if it accepts the resolved input resource types, produces the expected artifacts, satisfies parameter schemas, is reachable through an allowed backend, and passes applicable tool-contract and method-condition constraints.

\subsection{S5.3. Family-card retrieval and typed compatibility ranking}\label{supp:s5.3}

Tool retrieval uses a hierarchical action space. The planner first maps the request to intent, modality, required resource type, output type, and possible method family. It then retrieves precomputed family cards. Fifteen family cards are available in the checked registry, each storing a summary, typical use cases, member tools, supported inputs and outputs, common constraints, backend availability, and an embedding vector.

Example family-card excerpt:
\begin{suppcode}
family_card: predictive_modeling
inputs:
  - feature_matrix
  - target_vector
  - fold_manifest
outputs:
  - prediction_scorecard
  - trained_model_or_coefficients
  - heldout_predictions
common_constraints:
  - leakage_check
  - grouped_or_nested_cv
  - permutation_or_null_control
backends:
  - local_python
  - slurm
\end{suppcode}

Expanded candidates are ranked by typed-contract compatibility, input/\allowbreak{}output resource match, backend requirements, runtime availability, policy scope, method-condition constraints, and task fit.
\begin{center}
\small
\setlength{\tabcolsep}{5pt}
\renewcommand{\arraystretch}{1.15}
\begin{tabular}{|>{\RaggedRight\hspace{0pt}\arraybackslash}p{0.32\textwidth}|>{\RaggedRight\hspace{0pt}\arraybackslash}p{0.59\textwidth}|}
\hline
\rowcolor{brtablehead}\textbf{Ranking factor} & \textbf{Example} \\
\hline
Input compatibility & Candidate accepts a connectivity or feature matrix. \\
\hline
Output compatibility & Candidate produces predictions and scorecard artifacts. \\
\hline
Backend availability & Local Python or Slurm route is available under current policy. \\
\hline
Constraint compatibility & Candidate supports grouped CV or nested CV. \\
\hline
Policy scope & Candidate is allowed under the active allowlist mode. \\
\hline
\end{tabular}
\end{center}

\subsection{S5.4. Canonical identity, Neurodesk mapping, and NiWrap descriptors}\label{supp:s5.4}

Canonical tool IDs are consumed by the planner, execution recipes, runtime registry, provenance records, and review rules. Backend mapping translates a canonical ID into a local Python entry point, Neurodesk module name, container command, shell command, or Slurm script. This prevents the same method from being represented differently across planning, execution, provenance, and review.

Example backend mapping:

\begin{suppcode}
canonical_tool_id: fsl_randomise
registry_spec: fsl_randomise_contract
backend: Neurodesk
module: fsl/\allowbreak{}6.0
command_template: randomise -i {input_4d} -o {output_prefix} -d {design_mat} -t {design_con}
expected_outputs:
  - corrected_statistical_map
  - command_log
\end{suppcode}

NiWrap-generated machine-readable descriptors serve as metadata and compatibility records. They support discovery, parameter schemas, and compatibility checks, but execution is routed through Neurodesk recipes or other backend recipes rather than through NiWrap's own execution path.

\subsection{S5.5. Allowlists, cross-stage context, and operational cache}\label{supp:s5.5}

Tool allowlists support researcher-facing and diagnostic modes. Curated interactive-safe mode is used for routine researcher-facing sessions and excludes unsafe or irrelevant actions. Diagnostic mode broadens access for routing analysis, benchmark evaluation, registry testing, and controlled internal checks. The active allowlist determines which tools can be retrieved and is recorded in the run card.

Cross-stage context propagates structured constraints between planning rounds. Example:

\begin{suppcode}
cross_stage_context:
  controversial_choice: GSR
  required_sensitivity:
    - rerun_with_GSR
    - rerun_without_GSR
  claim_caveat_required: true
  reviewer_attention: controversial_choice
\end{suppcode}

Operational caches reduce repeated BR-KG and registry lookups during long episodes. These caches are separate from reviewed memory: cached retrieval facts can speed up planning, but they are not accepted claims and do not replace run-bundle provenance.

\section{S6. Constraint taxonomy, compiler, and commitment gate}\label{supp:s6}

\subsection{S6.1. Hard and soft constraints}\label{supp:s6.1}

Hard constraints are validity boundaries. If a hard constraint is violated, the analysis is not accepted as valid. Soft constraints are consequential methodological choices that may be defensible but require documentation, sensitivity analysis, qualification, or caveats.

\begin{center}
\small
\setlength{\tabcolsep}{5pt}
\renewcommand{\arraystretch}{1.15}
\begin{tabular}{|>{\RaggedRight\hspace{0pt}\arraybackslash}p{0.42\textwidth}|>{\RaggedRight\hspace{0pt}\arraybackslash}p{0.14\textwidth}|>{\RaggedRight\hspace{0pt}\arraybackslash}p{0.35\textwidth}|}
\hline
\rowcolor{brtablehead}\textbf{Constraint} & \textbf{Type} & \textbf{Example consequence} \\
\hline
Missing fold manifest for predictive modeling & Hard & Block commitment or execution. \\
\hline
Feature selection, standardization, or harmonization outside CV \citep{varoquaux_cross-validation_2018} & Hard & Revise pipeline before acceptance. \\
\hline
Invalid exchangeability for permutation testing \citep{winkler_permutation_2014,winkler_multi-level_2015} & Hard & Block inference claim. \\
\hline
GSR without sensitivity analysis \citep{murphy_impact_2009,ciric_benchmarking_2017,parkes_evaluation_2018} & Soft & Warn; require with/\allowbreak{}without-GSR reporting or caveat. \\
\hline
Dynamic-FC window choice without robustness check & Soft & Warn; require window-length sensitivity or null comparison. \\
\hline
Missing software version or random seed & Soft & Warn; require reproducibility completion. \\
\hline
\end{tabular}
\end{center}

BR-KG negative-knowledge records encode known failure conditions even when the software can technically run.

\begin{suppcode}
negative_knowledge_record:
  edge: KNOWN_TO_FAIL_WHEN
  method: predictive_modeling
  condition: feature_selection_outside_cv
  severity: hard
  action: block_or_revise
\end{suppcode}

Hard constraints are derived from tool contracts, method-condition records, and community validators. Soft constraints are derived from conditional-knowledge records, community-practice annotations, and tool-contract clauses.

\subsection{S6.2. Constraint edge vocabulary and compiler algorithm}\label{supp:s6.2}

Hard constraints are compiled from KNOWN\_TO\_FAIL\_WHEN and VIOLATES\_ASSUMPTION edges, tool-contract clauses, and community validators. Soft constraints are compiled from HOLDS\_UNDER, SENSITIVE\_TO, and DEPENDS\_ON edges. The same vocabulary is used in planning, preflight, review, and claim calibration.

Compiler algorithm:

\begin{center}
\small
\setlength{\tabcolsep}{5pt}
\renewcommand{\arraystretch}{1.15}
\begin{tabular}{|>{\RaggedRight\hspace{0pt}\arraybackslash}p{0.12\textwidth}|>{\RaggedRight\hspace{0pt}\arraybackslash}p{0.79\textwidth}|}
\hline
\rowcolor{brtablehead}\textbf{Step} & \textbf{Compiler action} \\
\hline
1 & Retrieve method-condition records for each candidate workflow or tool. \\
\hline
2 & Retrieve relevant tool-contract clauses and community-validator requirements. \\
\hline
3 & Bind abstract conditions to the current dataset, resource ledger, parameters, and analysis state. \\
\hline
4 & Generate executable or inspectable checks. \\
\hline
5 & Classify each check as hard or soft. \\
\hline
6 & Store the active constraint set in the run bundle. \\
\hline
7 & Feed hard checks to filters or blockers and soft checks to ranking, warnings, sensitivity requirements, or caveat language. \\
\hline
8 & Re-run the compiler whenever the candidate tool set, resource ledger, specification ledger, or analysis plan changes. \\
\hline
\end{tabular}
\end{center}

Binding means converting an abstract rule into a concrete field or path. For example, "fold manifest required" becomes a check against the fold-manifest field in the HCP-YA resource ledger.

Example compiler output:

\begin{suppcode}
active_constraints:
  - id: grouped_cv_required
    type: hard
    bound_field: resource_ledger.fold_manifest
    status: missing
    verdict: block
  - id: gsr_sensitivity_required
    type: soft
    bound_field: preprocessing.global_signal_regression
    status: unresolved
    verdict: warn
    required_action: run_with_and_without_GSR_or_add_caveat
\end{suppcode}

\subsection{S6.3. Commitment gate}\label{supp:s6.3}

The commitment gate is the boundary between planning and execution. In interactive mode, the researcher approves, modifies, or rejects the proposed plan. In bounded autonomous mode, the gate is exercised by a critic or supervisor under a predeclared policy. In benchmark pass-through mode, human approval may be disabled for evaluation-only runs, but the evidence, tool discovery, preflight, recipe, artifact-observation, and review pathway is unchanged.

\begin{center}
\small
\setlength{\tabcolsep}{5pt}
\renewcommand{\arraystretch}{1.15}
\begin{tabular}{|>{\RaggedRight\hspace{0pt}\arraybackslash}p{0.25\textwidth}|>{\RaggedRight\hspace{0pt}\arraybackslash}p{0.28\textwidth}|>{\RaggedRight\hspace{0pt}\arraybackslash}p{0.38\textwidth}|}
\hline
\rowcolor{brtablehead}\textbf{Mode} & \textbf{Gate authority} & \textbf{Possible decisions} \\
\hline
Interactive & Human researcher & approve /\allowbreak{} modify /\allowbreak{} reject. \\
\hline
Bounded autonomous & Critic or supervisor policy & approve /\allowbreak{} approve with warnings /\allowbreak{} revise /\allowbreak{} block /\allowbreak{} terminate. \\
\hline
\end{tabular}
\end{center}

Example commitment-gate record:

\begin{suppcode}
commitment_gate:
  episode_id: ep_hcp_predict_001
  authority: human_researcher
  decision: revise
  reason: missing_fold_manifest
  active_hard_blocks:
    - grouped_cv_required
  allowed_next_actions:
    - resolve_fold_manifest
    - choose_non_predictive_workflow
    - terminate_as_non_executable
\end{suppcode}

\begin{figure}[!htbp]
\centering
\includegraphics[width=\textwidth]{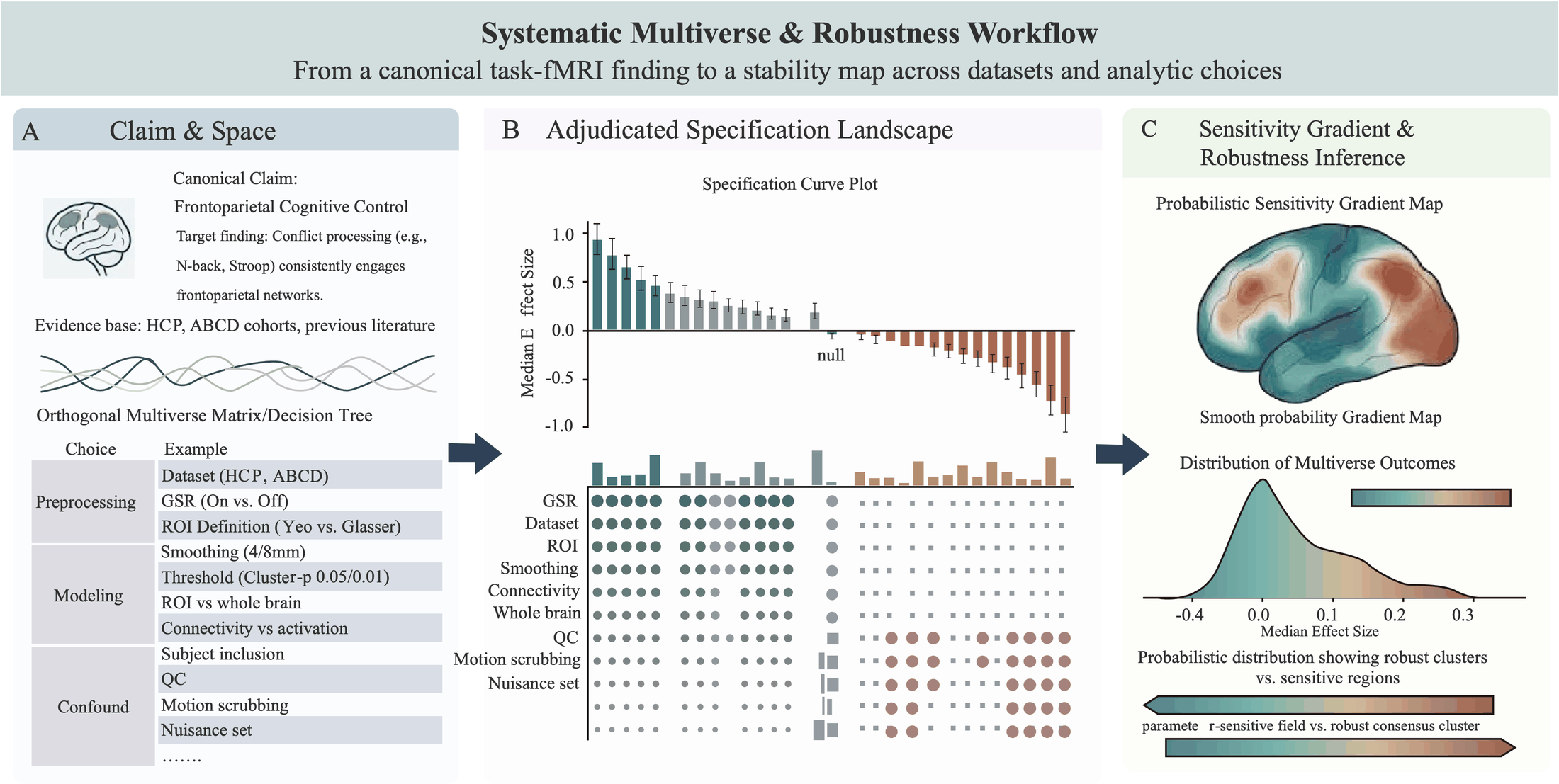}
\caption{\textbf{Systematic multiverse and robustness workflow.} A candidate claim is expanded into an admissible specification space through evidence assembly, tool-card constraints, resource checks, and researcher commitment. Each branch is either executed, blocked, or rejected under the active constraint vocabulary, preserving both successful artifacts and non-executed alternatives in the run bundle. The resulting landscape is summarized by specification curves, robustness tables, branch-level caveats, and claim-family verdicts. The workflow turns analytic multiplicity into a reviewable robustness boundary with the specification space fixed before claim calibration.}
\label{fig:multiverse-workflow}
\end{figure}

\section{S7. Execution substrate, preflight, and provenance}\label{supp:s7}

\begin{figure}[!htbp]
\centering
\includegraphics[width=\textwidth]{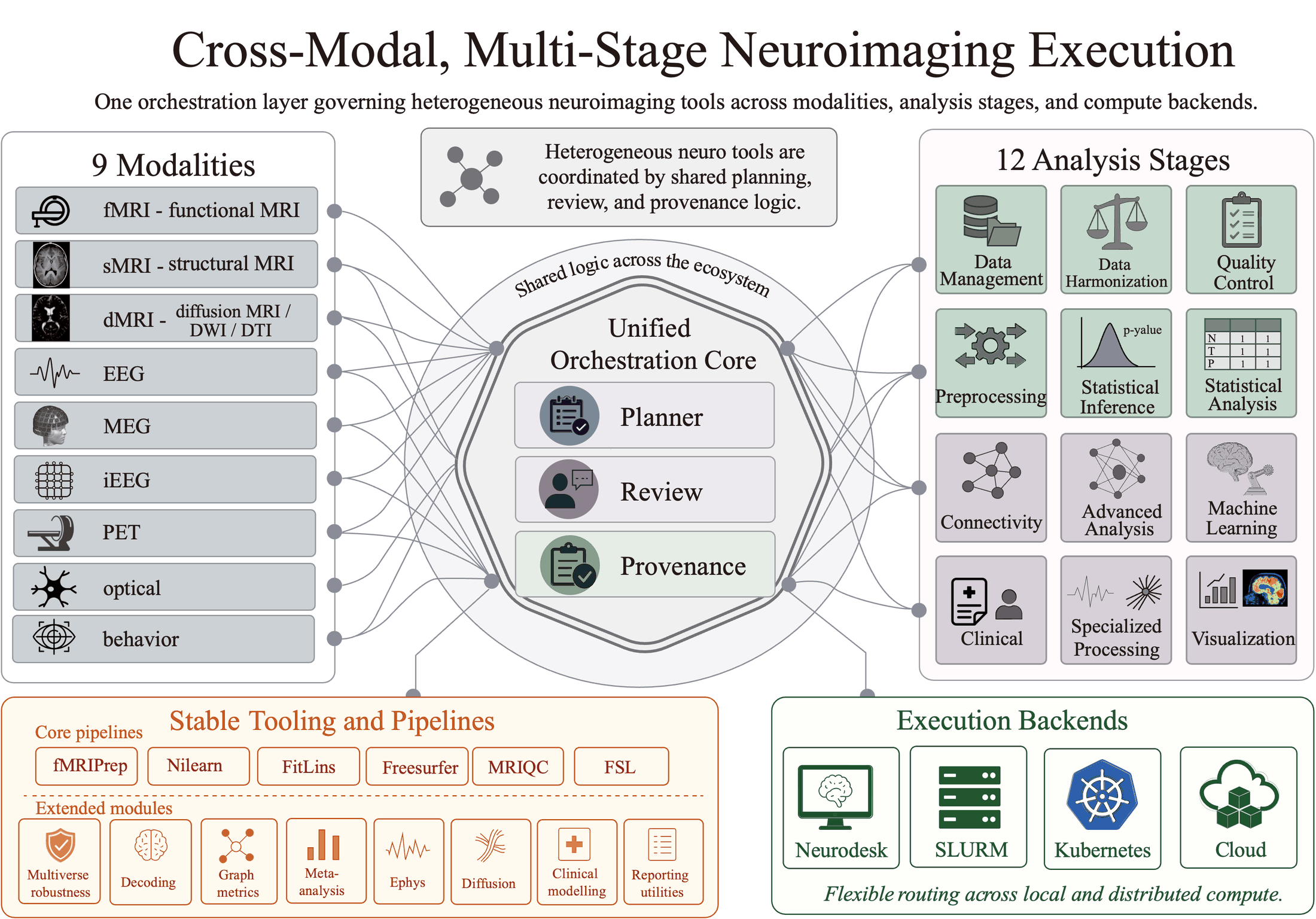}
\caption{\textbf{Cross-modal, multi-stage neuroimaging execution substrate.} Brain Researcher routes heterogeneous neuroimaging workflows through a shared execution layer that coordinates modality-specific tools, analysis stages, provenance capture, and backend selection. The figure summarizes the execution surface covered by Supplementary Methods S7: supported modalities, stable tooling and pipeline families, execution backends, and the shared planner-review-provenance logic that links a selected workflow to observed artifacts.}
\label{fig:execution-substrate}
\end{figure}

\subsection{S7.1. Execution backends and backend selection}\label{supp:s7.1}

Brain Researcher supports local Python, Neurodesk, generic containers, Slurm/\allowbreak{}HPC, Kubernetes, AWS Batch, and other configured execution substrates. Backend selection is based on resource requirements, software availability, queue status, runtime policy, cost, health, and compatibility with the selected recipe. The selected backend and routing rationale are recorded in the run card.

Example backend-selection excerpt:
\begin{suppcode}
backend_selection:
  selected_backend: slurm
  rationale:
    - expected_runtime_high
    - local_python_route_unavailable
    - container_available
    - queue_policy_allows_batch
  fallback_backend: Neurodesk_container
\end{suppcode}

The scheduler records dependency resolution, retries, state transitions, and emitted events. Detailed scheduler strategies such as eager, lazy, batch execution, checkpoint persistence, and deadlock detection are implementation details recorded in \hyperref[app:f]{Appendix F}.

\subsection{S7.2. Operational and scientific preflight}\label{supp:s7.2}

Operational preflight asks: can this run in the current environment? Scientific preflight asks: would this run be valid for the intended claim?

Operational checks include container image or CVMFS availability, module importability, Python entry-point availability, backend health, filesystem access, network policy, allowed roots, authentication state, and timeout budgets. Scientific checks include design-matrix rank, BIDS validity, coordinate-space consistency, resource-type compatibility, sample alignment, statistical prerequisites, fold-manifest integrity, and tool-contract requirements.

Example preflight result:

\begin{suppcode}
preflight:
  operational:
    filesystem_access: pass
    container_available: pass
    backend_health: pass
  scientific:
    fold_manifest: block
    confound_specification: warn
    feature_target_alignment: pass
  final_verdict: block
\end{suppcode}

Hard failures detected during preflight block execution. Soft warnings may allow execution but require caveats, sensitivity checks, or reviewer attention.

\subsection{S7.3. Runtime isolation and real-time boundaries}\label{supp:s7.3}

Runtime isolation uses container-level sandboxing, restricted environments, path checks, ephemeral writable layers, and network isolation. Tools requiring broader filesystem access are executed only under a tracked relaxed mode with additional logging. In the evaluated configuration, MCP returns recipes, policy verdicts, and observation tools; the external coding agent, shell, container runtime, Neurodesk environment, or scheduler performs mutation or computation unless gated MCP administrative execution is explicitly enabled.

Real-time paths such as neurofeedback and online two-photon pipelines operate as continuous closed-loop processes outside the standard batch scheduler. They currently receive weaker provenance and post hoc review coverage than batch workflows and are reported as limitations in \hyperref[supp:s12]{S12} rather than treated as fully evaluated execution pathways.

\subsection{S7.4. Run bundle and authoritative provenance}\label{supp:s7.4}

Each run emits a run bundle (in agent-executed mode its completeness reflects the typed tools the agent invoked, and for a fully external analysis such as NeuroMark it is a post-hoc reconstruction). The run bundle is immutable and authoritative: memory cards, claim cards, and summaries are derived projections and do not replace the full record. If a claim card conflicts with the run bundle, the run bundle is authoritative.

Example run-bundle manifest:

\begin{suppcode}
run_bundle:
  event_trace: events.jsonl
  trajectory_document: trajectory.yaml
  observation_record: observations.yaml
  analysis_bundle: analysis/\allowbreak{}
  run_card: run_card.yaml
  artifact_manifest: manifest.json
  checksums: sha256_manifest.txt
  review_cards: review/\allowbreak{}
  recovery_events: recovery.jsonl
\end{suppcode}

The event trace stores sequence numbers, timestamps, event types, session IDs, and phase markers. The trajectory document records planning decisions, tool invocations, parameter bindings, and constraint verdicts. The observation record stores intermediate results, QC outputs, diagnostics, and environment-observed state. The analysis bundle stores final artifacts, metadata, checksums, and provenance chains.

\section{S8. Scientific verification layer and claim calibration}\label{supp:s8}

The scientific verification layer is the main post-execution acceptance gate in Brain Researcher. Execution can show that a workflow ran, Harbor can show that a bounded task produced required files or statistics, and a language model can explain an output, but none of these events by itself accepts a neuroimaging claim. For the reported collaborator and bounded-autonomous evaluations, S8 defines the layer that consumes the run bundle, applies validity rules, assigns BLOCK or WARN findings, calibrates claim language, records one of six reportable claim states (accepted, qualified, revised, blocked, rejected, or deferred), routes unresolved decisions for human escalation, and determines memory-writeback eligibility. In this paper, ``verification'' means structured artifact- and claim-level challenge against declared scientific rules; it does not mean formal proof that a claim is true.

\subsection{S8.1. Verification inputs and validity layers}\label{supp:s8.1}

Scientific verification consumes the run bundle, not the model transcript alone. It checks the selected plan, active constraints, execution trace, artifact manifest, logs, QC outputs, scorecards, candidate claims, and prior evidence. This prevents fluent explanations from substituting for observed artifacts.

\begin{center}
\small
\setlength{\tabcolsep}{5pt}
\renewcommand{\arraystretch}{1.15}
\begin{tabular}{|>{\RaggedRight\hspace{0pt}\arraybackslash}p{0.28\textwidth}|>{\RaggedRight\hspace{0pt}\arraybackslash}p{0.63\textwidth}|}
\hline
\rowcolor{brtablehead}\textbf{Validity layer} & \textbf{Example checks} \\
\hline
Statistical validity & Design-model match, degrees of freedom, multiple comparisons, exchangeability, inference framework. \\
\hline
Measurement validity & Motion, distortion, registration, CIFTI/\allowbreak{}surface/\allowbreak{}volume consistency, QC, reliability. \\
\hline
Construct validity & Task-to-construct mapping, reverse-inference risk, contrast interpretation. \\
\hline
Generalization validity & Cross-validation integrity, leakage, sample size, site effects, stimulus effects, out-of-sample support. \\
\hline
Claim validity & Whether causal, mechanistic, clinical, external-validity, or biomarker language exceeds evidence. \\
\hline
\end{tabular}
\end{center}

\subsection{S8.2. Rule severity, lifecycle, and reason tags}\label{supp:s8.2}

Verification rules are organized by severity, validity layer, reason tags, detection fields, and default action. BLOCK rules indicate high-confidence validity failures requiring revision before acceptance. WARN rules identify risks that may be addressed by sensitivity analysis, supplementary reporting, alternative modeling, or claim revision.

Example review-rule card:

\begin{suppcode}
rule_id: leakage_standardization_outside_cv
severity: BLOCK
validity_layer: generalization
reason_tag: leakage
detection_field: pipeline_dag
trigger: scaler_fit_before_cv_split
default_action: revise
required_fix: fit_scaler_inside_each_training_fold
\end{suppcode}

Rule lifecycle status is defined in \hyperref[app:g]{Appendix G}, with the full rule registry, implementation-priority queue, calibration case library, review-context schema extensions, and sensitivity templates reported in the extended registry there. The main text distinguishes deterministic implemented checks from candidate or calibration-only rules when it affects the interpretation of a verification verdict: only implemented checks are treated as enforced, whereas deterministic, schema-dependent, NLP/LLM, and calibration-only candidates are reported as protocol or reviewer-facing constraints unless an episode records their enforcement. Reason tags include leakage, circularity, confound, null\_mismatch, low\_reliability, claim\_inflation, prior\_conflict, and controversial\_choice. Prior conflict is an audit trigger rather than a veto. Novelty cannot override a hard method error.

\subsection{S8.3. BLOCK and WARN rule families}\label{supp:s8.3}

Representative BLOCK families include split/\allowbreak{}CV/\allowbreak{}leakage integrity, design-model mismatch, multiple-comparison and inference-framework errors, circular analysis, and hard confound failures. Examples include feature selection outside CV, standardization outside CV, harmonization outside CV, confound regression outside CV, test-set model selection, uncorrected whole-brain inference, invalid permutation exchangeability, brain-map correlation without spatial nulls, same-data ROI definition with effect testing, and demographic confounds omitted from group models.

Representative WARN families include measurement-quality gaps, soft multiple-comparison issues, multivariate/\allowbreak{}RSA/\allowbreak{}predictive-modeling risks, reliability and sample-size concerns, claim inflation, controversial choices, prior conflict, and reporting gaps. Examples include missing MRIQC or visual QA \citep{esteban_mriqc_2017}, multiple ROI tests without correction, RSA model comparison without correction \citep{kriegeskorte_representational_2008}, above-chance classification without permutation or confidence intervals, small-sample biomarker claims, reverse inference, correlation-as-prediction, fit-as-mechanism, GSR without sensitivity analysis \citep{murphy_impact_2009}, dynamic FC without null or window sensitivity \citep{allen_tracking_2014}, graph thresholding without multi-threshold checks, missing random seed, missing software versions, and failed BIDS validation \citep{gorgolewski_brain_2016}.

BLOCK returns the episode to revision or termination. WARN can allow execution or acceptance only if the required caveat, sensitivity analysis, or reporting completion is recorded.

\subsection{S8.4. Verification verdicts and revision routing}\label{supp:s8.4}

\begin{center}
\small
\setlength{\tabcolsep}{5pt}
\renewcommand{\arraystretch}{1.15}
\begin{tabular}{|>{\RaggedRight\hspace{0pt}\arraybackslash}p{0.22\textwidth}|>{\RaggedRight\hspace{0pt}\arraybackslash}p{0.44\textwidth}|>{\RaggedRight\hspace{0pt}\arraybackslash}p{0.25\textwidth}|}
\hline
\rowcolor{brtablehead}\textbf{Verdict} & \textbf{Meaning} & \textbf{Next action} \\
\hline
accept & Artifacts and claims satisfy active constraints and review rules. & Eligible for memory writeback. \\
\hline
accept with qualifications & Claim is valid only under explicit caveats or condition tags. & Write qualified claim card. \\
\hline
revise & Result is not yet acceptable but may become acceptable with specified evidence or repair. & Return to planning, execution, or claim editing. \\
\hline
block & Hard validity failure prevents acceptance. & Stop, redesign, or terminate as non-accepted. \\
\hline
reject & Candidate claim is unsupported or contradicted. & No accepted memory writeback. \\
\hline
defer & Candidate passes an internal gate but fails a predeclared eligibility rule for the next validation tier. & Hold pending external replication or the required tier. \\
\hline
escalate & Human researcher or domain expert must resolve the issue. & Pause or route for expert judgment. \\
\hline
\end{tabular}
\end{center}

Each evaluated claim in the collaborator and bounded-autonomous cases is reported in exactly one of six states (accepted, qualified, revised, blocked, rejected, deferred; Table~\ref{tab:claim-states-supp}). These states correspond one-to-one to the \emph{accept}, \emph{accept with qualifications}, \emph{revise}, \emph{block}, \emph{reject}, and \emph{defer} verdicts above; \emph{escalate} is a routing action that hands the decision to a human rather than a terminal claim state. Prospectively governed episodes apply rules committed before the confirmatory test, whereas post-hoc audits apply explicit adjudication criteria to completed artifacts. In either tier, the state reflects the applicable evidence rule rather than the strength of the headline number alone (\hyperref[supp:s11.2]{S11.2}--\hyperref[supp:s11.3]{S11.3}). A terminal state is not intended to replace the condition structure of the evidence. When support divides along an identifiable analytic axis, the claim report pairs the state with a support-partition annotation that names the axis, its levels, and support at each level: the state governs claim promotion, while the annotation preserves the scientific pattern.

\begin{table}[!htbp]
\centering
\small
\caption{\textbf{Claim states and their explicit adjudication criteria.} Each evaluated claim in the collaborator and bounded-autonomous cases is assigned exactly one state by the review layer. For prospectively governed episodes, thresholds are committed before the confirmatory test; post-hoc audits instead apply documented criteria to completed artifacts. This is the reporting-level projection of the review verdicts in the table above.}
\label{tab:claim-states-supp}
\begin{tabular}{@{}p{0.16\textwidth}p{0.72\textwidth}@{}}
\hline
\textbf{State} & \textbf{Decision rule or adjudication criterion} \\
\hline
Accepted & Holds across the episode-defined majority of admissible analytic specifications, with no single defensible choice reversing it, and passes every check required for the stated claim language. \\
Qualified & Supported only under explicit, named conditions: support is confined to a subset of specifications or holds only with stated caveats; when an identifiable support partition exists, it is retained alongside the state. \\
Revised & The target, measurement, or validation plan is changed in response to a failed gate, and the redesigned successor passes the same rules. \\
Blocked & A resource, design, power, or validity constraint prevents promotion from exploratory to confirmatory, regardless of the point estimate. \\
Rejected & Fails the applicable null, replication, or support rule. \\
Deferred & Passes an internal gate but does not yet satisfy eligibility for the next validation tier (for example, while awaiting external replication). \\
\hline
\end{tabular}
\end{table}

Example revision-routing card:

\begin{suppcode}
review_verdict: revise
triggering_rule: gsr_without_sensitivity
hard_or_soft: soft
required_evidence:
  - rerun_with_GSR
  - rerun_without_GSR
  - report_effect_stability
allowed_claim_language: qualified_only
\end{suppcode}

The review\_context schema supplies optional fields for model specification, multiple-comparison correction, measurement/\allowbreak{}QC, pipeline DAG steps, leakage detection, ROI provenance, reproducibility, software versions, random seeds, and BIDS validation status. Missing fields cause rules to skip or warn according to rule policy; missing review\_context fields do not automatically create scientific acceptance.

\subsection{S8.5. Claim calibration and memory eligibility}\label{supp:s8.5}

Claim calibration is separate from computation. A statistically valid result cannot be reported as causal, mechanistic, externally validated, clinically predictive, or a biomarker unless the corresponding evidence exists.

\begin{center}
\small
\setlength{\tabcolsep}{5pt}
\renewcommand{\arraystretch}{1.15}
\begin{tabular}{|>{\RaggedRight\hspace{0pt}\arraybackslash}p{0.3\textwidth}|>{\RaggedRight\hspace{0pt}\arraybackslash}p{0.28\textwidth}|>{\RaggedRight\hspace{0pt}\arraybackslash}p{0.33\textwidth}|}
\hline
\rowcolor{brtablehead}\textbf{Computed result} & \textbf{Overclaim blocked} & \textbf{Allowed claim language} \\
\hline
Association in one dataset & Causal mechanism & Exploratory or dataset-qualified association. \\
\hline
Internal prediction signal & Clinical biomarker & Internally evaluated prediction result. \\
\hline
Robustness across parcellations & External validity & Condition-qualified robustness. \\
\hline
Leakage-controlled grouped CV & Mechanistic explanation & Validated internal predictive result, not mechanism. \\
\hline
\end{tabular}
\end{center}

Only accepted or explicitly qualified claims are eligible for memory writeback. Blocked, rejected, or speculative claims remain in the run bundle.

\subsubsection{S8.5.1. The claim record as a concrete object}\label{supp:s8.5.1}
The auditable record an episode produces is a set of versioned JSON objects, not an informal log. A \emph{commitment card} (\texttt{commitment-card-v1}) freezes the claim text, scope boundary, allowed alternatives, same-null constraint, and success/failure criteria, and stores a content hash (\texttt{commitment\_hash}) computed before execution; re-deriving the hash later detects any post-hoc change to the committed plan. A \emph{claim card} (\texttt{claim-card-v1}) records the final bounded status of the claim, the checks it survived and failed, the evidence it draws on, and the evidence still required, and points back to the commitment card. Both are written to disk (\texttt{commitment\_card.json}, \texttt{claim\_card.json}) alongside an append-only ledger and the run bundle, so the record is exportable and can be opened by another reader after the session without rerunning the analysis. The representative card below comes from an earlier HCP-YA aggregate-prediction campaign and is retained as an illustration of the record schema; it is not the result record for the current HCP iterative episode in \hyperref[supp:s11.3.1]{Supplementary Methods S11.3.1}. Its \texttt{commitment\_hash} was a version-control soft anchor committed alongside its validator rather than a third-party time-stamped pre-registration.

\begin{suppcode}
{
  "schema_version": "claim-card-v1",
  "claim_id": "hcp_rsfc_aggregate_predictor",
  "claim_text": "A frozen ridge predictor on resting-state FC predicts the
                  five-component behavioral target in HCP-YA (fold-mean r = 0.190).",
  "status": "accepted",
  "scope_boundary": "HCP-YA; Schaefer-100x7 FC; no-GSR; 10-fold CV;
                     accepted with search correction, not covariate-robust;
                     internal validation only",
  "commitment_card_ref": "commit:hcp_rsfc_aggregate_predictor",
  "commitment_hash": "b1e4...e7a9",
  "evidence_bundle_refs": ["eb_hcp_fc_features", "eb_liu_components"],
  "survived_checks": [
    "family_aware_permutation_null (Family_ID block, 1000 perms, plus-one p)",
    "max_over_pipelines_null (38 replayable candidates)"
  ],
  "failed_checks": [
    "covariate_retention_gate (>=70
     demographic + IQ adjustment; aggregate retained 55
     below the 0.133 floor) -> bounds scope to search-corrected, not covariate-robust"
  ],
  "falsification_budget_spent": {"null_models": 2, "permutations": 1000},
  "next_required_evidence": ["external replication (HCP-Aging)"],
  "status_not_ground_truth": true
}
\end{suppcode}

This is the object the main text refers to as the auditable claim record; the review card (\hyperref[app:g]{Appendix G}) and the memory card (\hyperref[app:h]{Appendix H}) are its review-time and memory-time counterparts. A real exported example built on public Neurosynth coordinate evidence (a working-memory claim, final status \emph{weakened} under the conservative evidence profile) is released with the code, so a reader can open an actual \texttt{claim\_card.json} and its \texttt{evidence\_verdicts.json} directly rather than relying on the listing above. This legacy artifact predates the six-state reporting vocabulary used for the collaborator and bounded-autonomous evaluations; we retain its literal \emph{weakened} label and do not retroactively relabel the exported record. For the schizophrenia NeuroMark episode, by contrast, no sealed pre-analysis commitment card was created: the collaborator's three hypotheses were prespecified in their own study protocol rather than sealed as a Brain Researcher commitment card before execution, so the audit bundle we provide for that case is a post-hoc reconstruction assembled from the completed multiverse artifacts. Its claim card accordingly records \texttt{commitment\_card\_ref: null} and \texttt{pre\_run\_commitment\_card\_found: false} rather than asserting a pre-run seal, and is labeled post-hoc by design rather than backfilled. The sealed-commitment-card mechanism is therefore demonstrated on the released Neurosynth example; the NeuroMark episode exercises the multiverse, claim-state, and directionality-audit machinery.

\section{S9. Memory system, conflict detection, and BR-KG promotion}\label{supp:s9}

\subsection{S9.1. Memory objects and writeback gate}\label{supp:s9.1}

The memory system stores reviewed knowledge derived from completed episodes. It is not a transcript archive and not a replacement for the run bundle. Its purpose is to retrieve condition-tagged claims, prior tactics, unresolved caveats, review outcomes, and relations among findings.

\begin{figure}[!htbp]
\centering
\includegraphics[width=\textwidth]{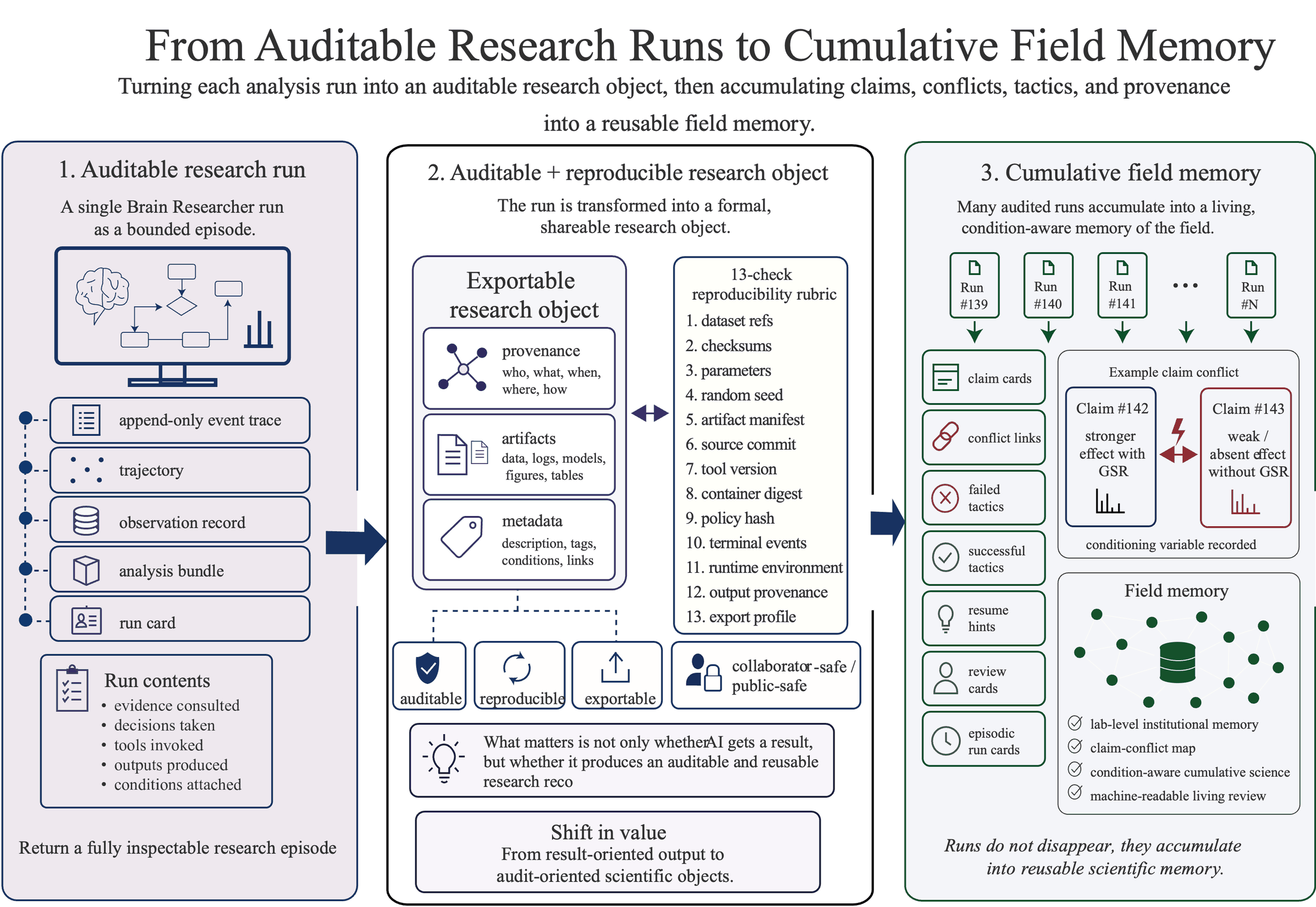}
\caption{\textbf{From auditable run bundles to cumulative field memory.} Completed episodes first produce run bundles containing the committed plan, executed actions, logs, artifacts, validation outcomes, review notes, failures, and caveats. The review layer determines which candidate claims are accepted or explicitly qualified for memory writeback; rejected, blocked, or failed candidates remain recoverable in the run bundle without becoming accepted graph facts. Eligible claims become condition-tagged claim cards with provenance pointers and relation events such as support, contradiction, refinement, conditioning, or supersession. Curated memory cards may later be promoted to BR-KG under a separate governance policy.}
\label{fig:field-memory}
\end{figure}

\begin{center}
\small
\setlength{\tabcolsep}{5pt}
\renewcommand{\arraystretch}{1.15}
\begin{tabular}{|>{\RaggedRight\hspace{0pt}\arraybackslash}p{0.28\textwidth}|>{\RaggedRight\hspace{0pt}\arraybackslash}p{0.63\textwidth}|}
\hline
\rowcolor{brtablehead}\textbf{Memory object} & \textbf{Purpose} \\
\hline
Claim card & Stores reviewed claim text, polarity, scope, conditions, evidence level, caveats, and provenance pointers. \\
\hline
Episodic run card & Stores the question, plan, tool sequence, successful tactics, failed tactics, resource blockers, and resume hints. \\
\hline
Review card & Stores review level, verdict, risk tags, issues, fixes, and final eligibility status. \\
\hline
Claim-relation event & Stores support, contradiction, refinement, conditioning, or supersession relations between claims. \\
\hline
\end{tabular}
\end{center}

Example claim card:

\begin{suppcode}
claim_card:
  claim: "Model performance remained above chance under grouped CV."
  polarity: support
  scope: HCP-YA sample only
  evidence_level: L3_internal_validation
  review_status: accepted_with_qualifications
  caveats:
    - no_external_dataset
  provenance_pointer: run_bundle/\allowbreak{}ep_hcp_predict_001
\end{suppcode}

Memory writeback is gated by review. Failed, blocked, rejected, or speculative candidate claims remain searchable through the run bundle but are not promoted as accepted memory claims.

\subsection{S9.2. Conflict detection and relation events}\label{supp:s9.2}

At writeback, the system encodes the new claim, retrieves similar prior claims above a configured threshold, and submits new/\allowbreak{}prior pairs to a critic for relation classification. The critic classifies apparent support, contradiction, refinement, conditioning, or supersession for retrieval and audit; it does not resolve scientific truth by itself. For contradictory or conditioning pairs, the system creates bidirectional claim-relation events and records the conditioning variable when available.

Example relation event:

\begin{suppcode}
relation_event:
  new_claim: claim_042
  prior_claim: claim_017
  relation: conditioning
  conditioning_variable: preprocessing_pipeline
  confidence: medium
  logged_in: run_bundle/\allowbreak{}ep_hcp_predict_001
\end{suppcode}

Memory is append-only. Claims may be superseded, conditioned, or contradicted, but they are not silently deleted.

\subsection{S9.3. Memory partitioning and BR-KG promotion}\label{supp:s9.3}

During benchmark evaluation, memory namespaces are partitioned so evaluated tasks cannot retrieve target answers from prior benchmark or collaborator episodes. Training, development, benchmark, collaborator, and autonomous-campaign namespaces are separated. Any shared memory is explicitly declared in the evaluation card.

BR-KG promotion is a separate curation path:

\begin{center}
\small
\setlength{\tabcolsep}{5pt}
\renewcommand{\arraystretch}{1.15}
\begin{tabular}{|>{\RaggedRight\hspace{0pt}\arraybackslash}p{0.12\textwidth}|>{\RaggedRight\hspace{0pt}\arraybackslash}p{0.79\textwidth}|}
\hline
\rowcolor{brtablehead}\textbf{Step} & \textbf{Curation action} \\
\hline
1 & Reviewed claim card is created. \\
\hline
2 & Candidate graph update is proposed. \\
\hline
3 & Schema validation is performed. \\
\hline
4 & Source and provenance checks are performed. \\
\hline
5 & Curation approval is recorded. \\
\hline
6 & Accepted BR-KG record is created. \\
\hline
\end{tabular}
\end{center}

This preserves the distinction between episode memory, candidate graph updates, and accepted BR-KG records.

\section{S10. Bounded research episodes and validation contracts}\label{supp:s10}

Throughout, the terms ``bounded autonomous'' and ``self-evolving'' denote the same narrow process: validation-gated agentic search in which the system proposes, tests, and revises candidate claims strictly within a researcher-declared action space, budget, and validation ladder. They do not denote autonomous scientific self-improvement, open-ended goal setting, or any relaxation of the predeclared gates.

\subsection{S10.1. Harbor task format and Brain Researcher mapping}\label{supp:s10.1}

Harbor provides a containerized task contract with an instruction file, environment definition, hidden verifier, metadata, and resource limits. In this work, Harbor bounds the rollout environment and returns task-level completion evidence. It is not the scientific reviewer.

For Brain Researcher benchmark rollouts, Harbor can realize a bounded episode as a task contract. The contract specifies the scientific objective, allowed resources, admissible action space, expected artifacts, execution budget, verifier logic, reward or partial-credit outputs, review rules, memory policy, and escalation conditions. The HCP and TRIBE research lines reported in S11.3 instead used prospective episode bundles and frozen successor contracts outside Harbor. BR-MCP remains the domain-control plane for tool discovery, dataset resolution, planning, recipe generation, policy checks, run observation, grounding, and scientific review.

\begin{figure}[!htbp]
\centering
\includegraphics[width=\textwidth]{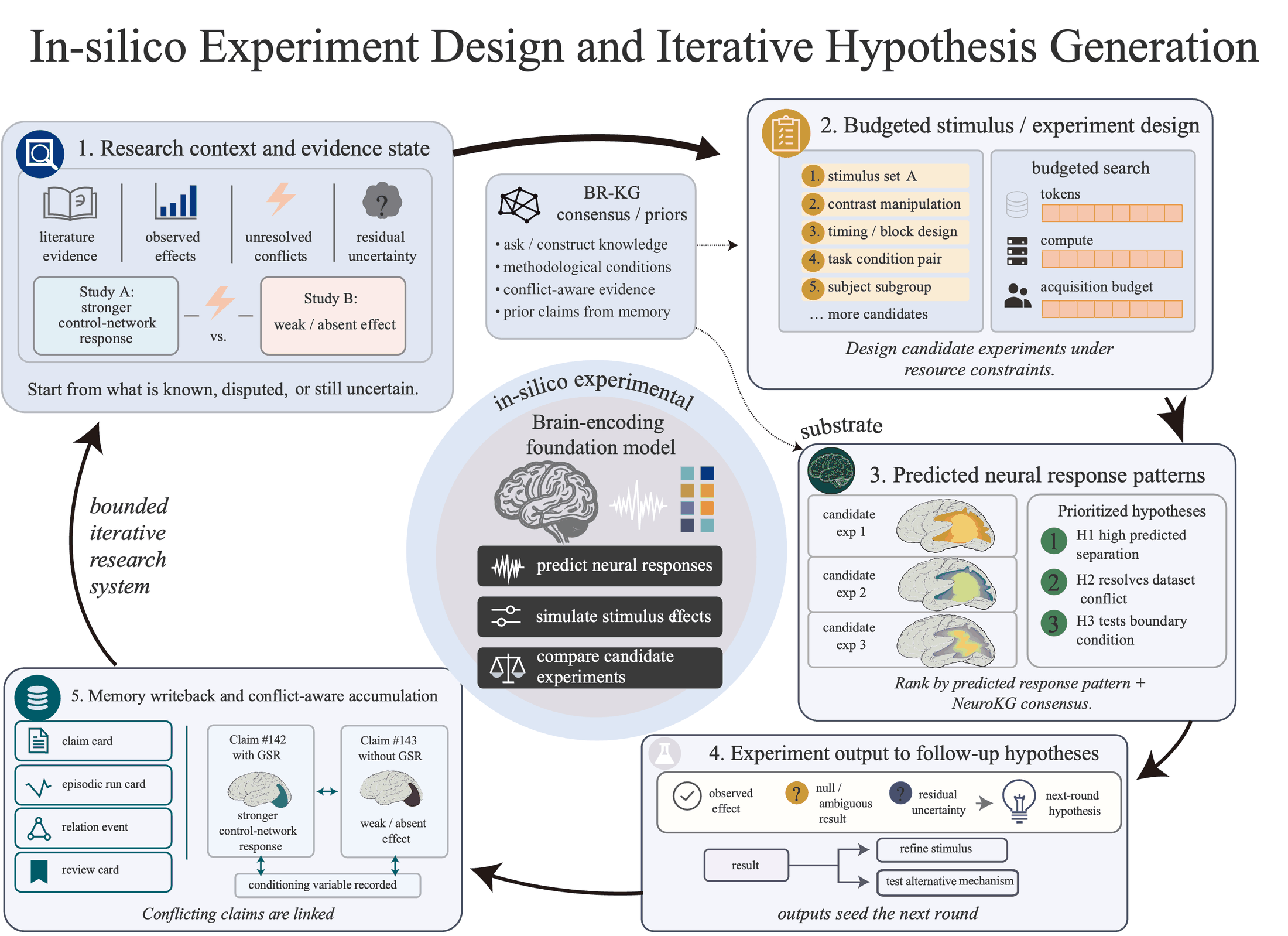}
\caption{\textbf{In-silico experiment design and bounded hypothesis iteration.} The bounded workflow starts from a declared research context, evidence state, allowed action vocabulary, budget, validation ladder, expected artifacts, and stopping rule. Candidate in-silico experiments are proposed, materialized, run through lower-resource validation tiers, and either rejected, revised, deferred, or advanced to stronger tests. Reward or search scores are kept separate from scientific acceptance, which requires the predeclared nulls, artifact checks, and review rules to pass. The loop returns reviewable follow-up hypotheses or memory candidates; acceptance remains with verification.}
\label{fig:self-evolving-workflow}
\end{figure}

Example Harbor verifier output:

\begin{suppcode}
harbor_verifier:
  artifact_manifest_present: true
  required_statistic_present: true
  schema_valid: true
  command_completed: true
  trajectory_requirements_met: partial
  reward: 0.8
\end{suppcode}

This verifier output can show that a task produced the required files and statistics while the Brain Researcher scientific verification layer still returns revise or block because of leakage, missing sensitivity analysis, invalid nulls, or claim inflation.

\begin{center}
\small
\setlength{\tabcolsep}{5pt}
\renewcommand{\arraystretch}{1.15}
\begin{tabular}{|>{\RaggedRight\hspace{0pt}\arraybackslash}p{0.24\textwidth}|>{\RaggedRight\hspace{0pt}\arraybackslash}p{0.42\textwidth}|>{\RaggedRight\hspace{0pt}\arraybackslash}p{0.25\textwidth}|}
\hline
\rowcolor{brtablehead}\textbf{Component} & \textbf{Checks} & \textbf{Does not check} \\
\hline
Harbor verifier & Files, schema, required statistic, command/\allowbreak{}test success, trajectory requirements. & Scientific validity or claim scope. \\
\hline
BR-MCP & Tool/\allowbreak{}dataset resolution, policy constraints, recipe packaging, provenance capture. & Final biological truth. \\
\hline
Execution backend & Logs, artifacts, metrics, environment-observed state. & Claim acceptability. \\
\hline
Scientific verification layer & Leakage, confounds, null controls, robustness, caveats, claim calibration. & Independent replication or complete scientific truth. \\
\hline
Memory gate & Accepted or qualified writeback. & Unreviewed or rejected claims. \\
\hline
\end{tabular}
\end{center}

\subsection{S10.2. Supervisor, critic, and bounded iteration}\label{supp:s10.2}

The bounded-loop architecture can use a supervisor--critic pair under fixed budgets and stopping criteria. The supervisor proposes the next analytic branch from the current episode state: open caveats, failed checks, memory conflicts, validation-ladder progress, and remaining compute/\allowbreak{}time budget. Candidate branches may vary a confound model, test an alternative parcellation, extend a sensitivity analysis, run a null-control analysis, investigate an anomaly, broaden the search, or abandon a target that cannot meet the contract. This subsection describes the available architecture, not an assertion that every reported episode used this exact controller.

Example branch-decision record:

\begin{suppcode}
branch_decision:
  branch_type: sensitivity_analysis
  reason: unresolved_GSR_warning
  proposed_action: rerun_with_and_without_GSR
  expected_resolution: qualify_or_clear_controversial_choice_warning
  budget_cost: one_additional_run
\end{suppcode}

After Harbor verification and review-layer inspection, the critic asks whether the previous cycle resolved a review finding, exposed a hard failure, improved validation level, revealed a method dependency, or merely repeated known work. It then returns a structured decision.

\begin{suppcode}
critic_decision:
  decision: continue
  reason: sensitivity_analysis_resolves_open_warning
  remaining_budget_cycles: 2
\end{suppcode}

Episodes terminate when the goal is reached, the validation ladder is satisfied, a hard review block cannot be resolved, progress stalls for the configured number of cycles, compute or wall-clock budgets are exhausted, the critic requests human escalation, or infrastructure recovery fails.

\begin{suppcode}
termination:
  status: terminated_budget_exhausted
  unresolved_findings:
    - external_validation_missing
  memory_writeback: qualified_claim_only
\end{suppcode}

\subsection{S10.3. Bounded autonomous episode contract}\label{supp:s10.3}

A bounded autonomous episode is evaluable only if it declares its target question, design space, budgets, stopping criteria, validation ladder, expected artifacts, review rules, memory-writeback policy, and escalation conditions. Each cycle must produce a branch-decision record, Harbor verification record, artifact manifest, review card, and claim-status update.

Example contract excerpt:
\begin{suppcode}
bounded_episode_contract:
  target_question: >
    Evaluate whether HCP-YA features predict the selected
    behavioral target.
  admissible_design_space:
    - predictive_modeling_family
    - grouped_cv
    - predeclared_covariates
    - parcellation_sensitivity
  cycle_budget: 3
  validation_ladder: L0_to_L3
  expected_artifacts:
    - prediction_scorecard
    - artifact_manifest
    - review_card
  memory_policy: accepted_or_qualified_only
  escalation_conditions:
    - unresolved_hard_block
    - out_of_design_space_request
\end{suppcode}

The contract prevents open-ended optimization from being mistaken for autonomous scientific discovery. Follow-up branches remain inside the admissible design space and must pass the same commitment, execution, Harbor verification, review, and memory-writeback rules as interactive episodes.

\subsection{S10.4. Validation ladder and claim strength}\label{supp:s10.4}

Bounded autonomous episodes use a validation ladder to prevent claim escalation from a single successful run.

\begin{center}
\small
\setlength{\tabcolsep}{5pt}
\renewcommand{\arraystretch}{1.15}
\begin{tabular}{|>{\RaggedRight\hspace{0pt}\arraybackslash}p{0.22\textwidth}|>{\RaggedRight\hspace{0pt}\arraybackslash}p{0.24\textwidth}|>{\RaggedRight\hspace{0pt}\arraybackslash}p{0.28\textwidth}|>{\RaggedRight\hspace{0pt}\arraybackslash}p{0.17\textwidth}|}
\hline
\rowcolor{brtablehead}\textbf{Ladder level} & \textbf{Purpose} & \textbf{Example evidence} & \textbf{Allowed claim strength} \\
\hline
L0 feasibility & Required data, tools, and resources exist. & Dataset/\allowbreak{}resource resolution and preflight pass. & Analysis is executable. \\
\hline
L1 minimal signal & Initial effect or prediction signal appears. & First-pass held-out test, permutation, or preliminary association. & Exploratory signal. \\
\hline
L2 robustness & Signal survives method variation. & Parcellation, confound, threshold, model, or preprocessing sensitivity. & Condition-qualified result. \\
\hline
L3 leakage/\allowbreak{}null control & Trivial validity failures are ruled out. & Grouped CV, nested CV, permutation, spatial null, or negative control. & Validated internal result. \\
\hline
L4 held-out or external support & Generalization tested beyond initial fit. & Held-out subjects, site, dataset, or independent benchmark. & Stronger generalization claim. \\
\hline
L5 claim calibration & Final language constrained by evidence. & Review card, caveats, memory relation events, claim governor. & Accepted or qualified claim. \\
\hline
\end{tabular}
\end{center}

For example, a first-pass held-out signal reaches L1. If the signal survives parcellation and confound sensitivity, it reaches L2. If grouped CV and permutation controls pass, it reaches L3. Without external dataset support, the claim cannot exceed L3; it cannot be called externally validated. Without construct or causal evidence, it cannot be called mechanistic.

\subsection{S10.5. Structured hypothesis triage and loop closure}\label{supp:s10.5}

Structured hypothesis triage is separate from bounded validation. Candidate questions, representations, estimators, or follow-up tests may be proposed during exploration, but a successor analysis must re-enter the episode machinery with a frozen target, analysis definition, evidence boundary, and stopping rule. Loop closure therefore means that one episode's result can determine the next bounded question; it does not permit the completed analysis to be rewritten after its outcome is known.

The two iterative episodes instantiated this mechanism in different ways. In HCP, a bounded development search supplied a fixed prediction configuration for a matched comparison and cross-outcome follow-up. In TRIBE, an open six-category screen supplied several candidate contrasts; a post-hoc geometry diagnostic and an explicit researcher decision selected speech versus tools, after which the estimand and evaluation rule were frozen before new stimuli were analysed. Complete episode accounting is reported in \hyperref[supp:s11.3]{Supplementary Methods S11.3}.

\begin{figure}[!htbp]
\centering
\includegraphics[width=\textwidth]{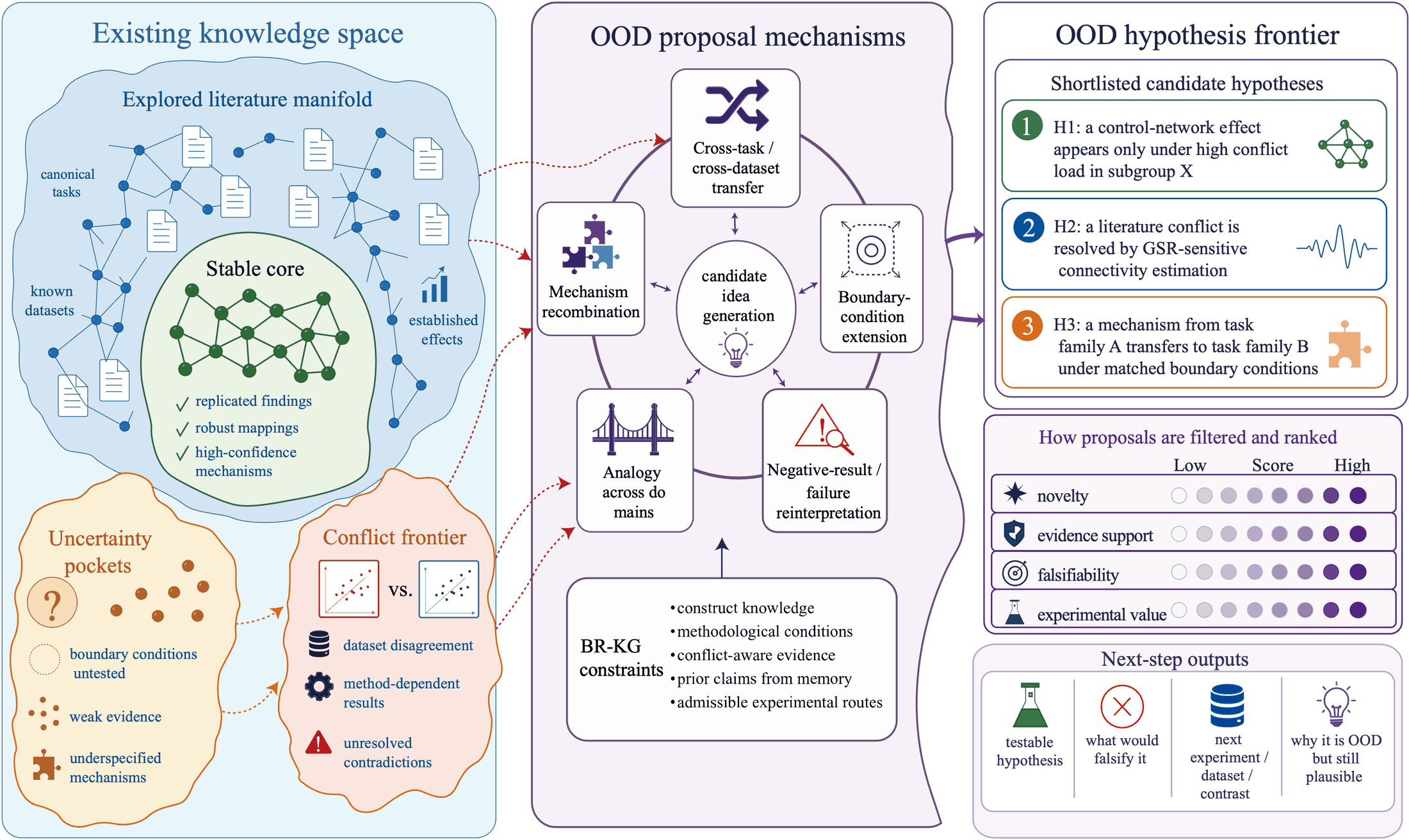}
\caption{\textbf{Out-of-distribution hypothesis proposal and triage.} Candidate hypotheses are generated near sparse, conflicting, or under-connected regions of the current knowledge space. Triage checks feasibility, novelty, evidentiary support, resource availability, and validation requirements before a candidate can enter an episode. Hypothesis generation can propose a direction, but promotion still requires evidence assembly, execution, review, and memory eligibility.}
\label{fig:hypothesis-generation}
\end{figure}

\begin{figure}[!htbp]
\centering
\includegraphics[width=\textwidth]{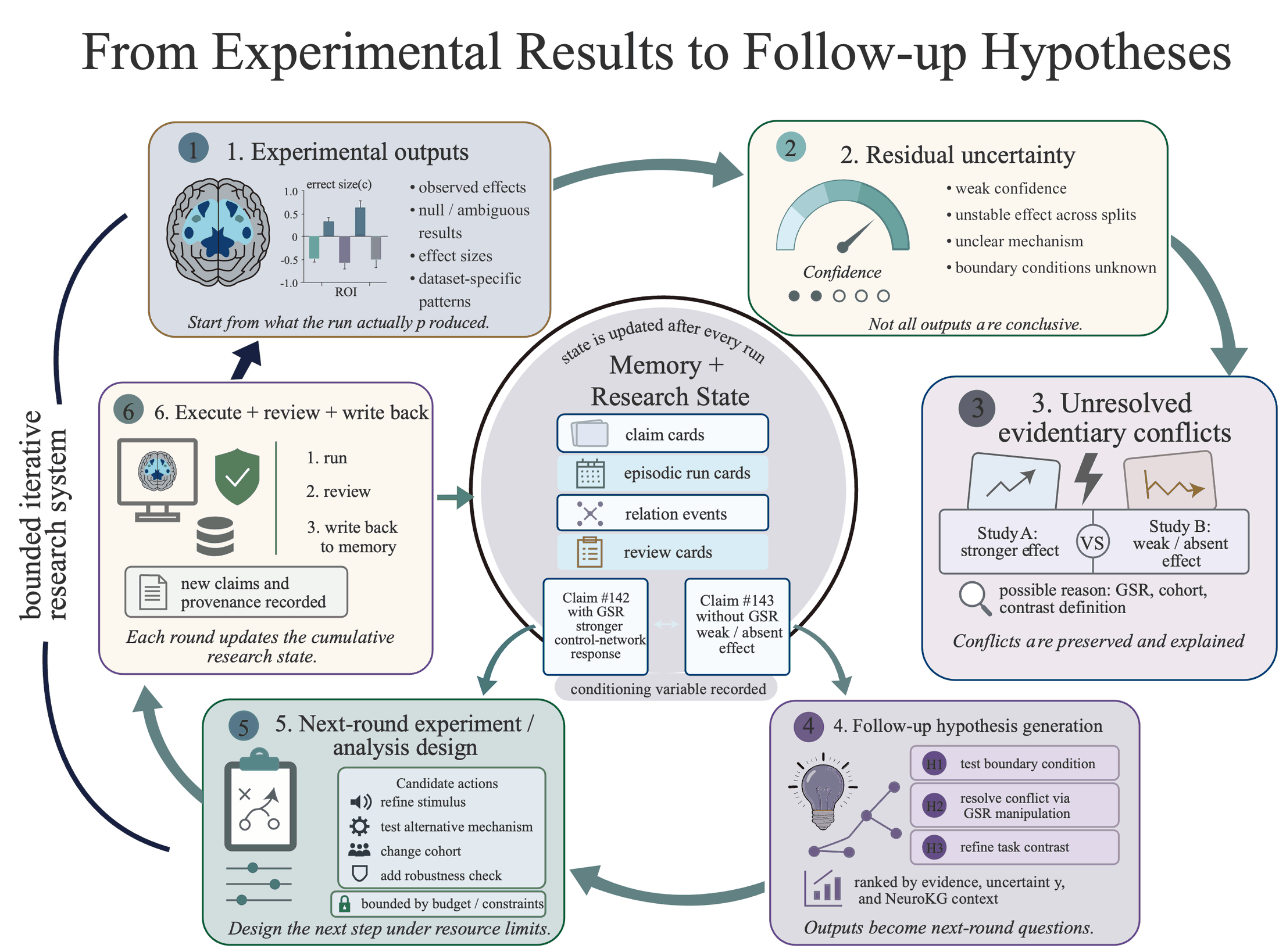}
\caption{\textbf{Result-to-hypothesis loop closure.} Completed outputs, residual uncertainty, failed branches, and condition-dependent findings can seed successor analyses. Every successor must pass the same commitment, execution, review, and memory-writeback path as a new episode.}
\label{fig:result-to-hypothesis}
\end{figure}

\subsection{S10.6. Freezing successor analyses}\label{supp:s10.6}

A successor analysis separates discovery from the next evaluation by freezing the scientific target, candidate or contrast, estimand, admissible inputs, and decision rule before the successor data are scored. Evidence may alter which successor is chosen, but it cannot alter the analysis already under evaluation. The HCP configuration was fixed and then refit separately for each behavioural target under the same evaluation procedure. The TRIBE speech--tools estimand was fixed before three non-overlapping recurring-source panels and, subsequently, before a score-blind panel drawn from four previously unused collections. Transport failures, excluded cells, and technical failures remain in the episode record rather than disappearing from denominators.

\subsection{S10.7. Same-agent sessions without Brain Researcher}\label{supp:s10.7}

We also inspected separately initiated sessions in which the same coding agent received the research goal but did not invoke Brain Researcher. These sessions provide qualitative process context, not a matched causal ablation: prompts, interaction histories, budgets, stopping decisions, and follow-up opportunities were not randomized or held identical. Their terminal states and the limits of this comparison are reported in \hyperref[supp:iterative-controls]{Supplementary Methods S11.3.3}.

\section{S11. Evaluation protocol}\label{supp:s11}

\paragraph{Three classes of verification check.} In conventional workflows verification is a check applied after an analysis is complete; in Brain Researcher it is the precondition for a methodological commitment to become infrastructure at all. Some checks admit unambiguous criteria, such as whether the required inputs exist, whether subject sets match, whether information has leaked across folds, or whether a null model is applied in the domain for which it was declared; these establish only that the stated analysis was carried out in a way that can support its claim, not which analysis is best. Other choices have no single correct answer but a set of defensible ones, such as the parcellation, confound model, or threshold, and here the system renders the navigation explicit rather than allowing a single pipeline to conceal it. A third class resists formalization altogether, including what question is worth asking, what should count as evidence, and how a result revises understanding; the system preserves commitments about this class but does not adjudicate them. That boundary is intentional, and everything beyond it remains with the researcher.

\subsection{S11.1. Evaluation components}\label{supp:s11.1}

The evaluation protocol has four components.

\begin{center}
\small
\setlength{\tabcolsep}{5pt}
\renewcommand{\arraystretch}{1.15}
\begin{tabular}{|>{\RaggedRight\hspace{0pt}\arraybackslash}p{0.34\textwidth}|>{\RaggedRight\hspace{0pt}\arraybackslash}p{0.57\textwidth}|}
\hline
\rowcolor{brtablehead}\textbf{Evaluation component} & \textbf{What is measured or documented} \\
\hline
BR-KG characterization & Graph snapshot, schema, source coverage, provenance coverage, density, orphan rates, property coverage, and example traversal paths. \\
\hline
Benchmark construction & Controlled task manifests, scoring contracts, with-BR/\allowbreak{}without-BR protocol, ablations, and evaluation partitions. \\
\hline
Collaborator cases & Real neuroimaging questions, dataset access, tool pathway, artifacts, robustness probes, review outcome, caveated conclusion, and human/\allowbreak{}system responsibility split. \\
\hline
Iterative research episode & Whether evidence from one stage is converted into a frozen successor analysis with explicit denominators, stopping states, and claim boundaries. \\
\hline
\end{tabular}
\end{center}

\subsubsection{S11.1.1. Benchmark surfaces}\label{supp:s11.1.1}
The quantitative Results use two benchmark surfaces. The first is a paired tool-calling benchmark: a correct response maps a natural-language neuroimaging request to an existing analysis tool or executable route, with required inputs available before execution. The technical metric family for this surface is routing capability. The second is an evidence-citation benchmark based on NeuroimageKnowledge: a claim is counted as grounded only when its cited evidence can be located and judged supportive for the stated claim. Support was evaluated by three LLM-based judges (Claude Opus 4.8, Codex GPT-5.5, and Gemini 3.1 Pro), each blind to benchmark condition. All reference-bearing rows were sent to all three judges (with-BR, 814 of 1{,}673 generated rows; without-BR, 469 of 1{,}525). Each label entered a three-state vote as verified, not verified, or \texttt{cannot\_judge}, and a row counted as verified when at least two judges voted verified. Every generated evidence-basis row, including rows without a reference and rows without two verified votes, remained in the denominator. Both surfaces compare with-BR and without-BR conditions over runs drawn from seven provider-defined model variants: Claude Code / Claude Opus 4.8, Codex GPT-5.5, Gemini 3.1 Pro, GLM-5.1, DeepSeek-V4-Pro, Kimi K2.5, and Qwen3.6-Plus. The tool-calling surface is a complete seven-model paired matrix; the evidence-citation headline is a descriptive question-level aggregate over all generated evidence-basis rows rather than a paired model-level estimate.

\begin{center}
\small
\setlength{\tabcolsep}{4pt}
\renewcommand{\arraystretch}{1.15}
\begin{tabular}{|>{\RaggedRight\hspace{0pt}\arraybackslash}p{0.23\textwidth}|>{\RaggedRight\hspace{0pt}\arraybackslash}p{0.33\textwidth}|>{\RaggedRight\hspace{0pt}\arraybackslash}p{0.34\textwidth}|}
\hline
\rowcolor{brtablehead}\textbf{Surface} & \textbf{Primary denominator} & \textbf{Reported metrics} \\
\hline
Tool calling & 60 task manifests \ensuremath{\times} seven models per condition (420 model-item trajectories). & Capability@k, Correct route/tool@k, and Handoff score@k. Capability@k is required-capability coverage after k non-neutral actions, not a binary route-reached indicator. \\
\hline
Evidence citation & 50 open-ended question sets; all generated evidence-basis rows per condition, with only rows receiving at least two verified votes contributing to the numerator. & \texttt{verified\_groundedness\_rate}, \texttt{verified\_among\_claimed\_grounded}, \texttt{citation\_spam\_rate}, and \texttt{answer\_correctness\_rate}. \\
\hline
\end{tabular}
\end{center}

\paragraph{What counts as verified grounding.}
For each claim a model supports with a citation, a judge blind to condition checks two things: whether the cited source can be located, and whether it supports the claim as stated. Only a full match is counted as verified. A source that supports a narrower version of the claim, or that is real but about a different point, is not counted. Each grounded item receives one label; Table~\ref{tab:grounding-label-examples} gives the three that occur here with a real example of each. Two further not-verified labels, for references that cannot be resolved or parsed, did not occur among the resolvable items. Because only full support is counted, many citations are not credited: a substantial share point to real sources that support only part of the claim (\emph{partial}) or a different point (\emph{no\_unrelated}). Judges also differ in how strictly they treat partial support, so the rate depends on the judge.

\begin{table}[!htbp]
\centering
\footnotesize
\setlength{\tabcolsep}{4pt}
\renewcommand{\arraystretch}{1.2}
\caption{\textbf{Grounding labels, with a representative with-BR item for each.} Each model claim is checked, blind to condition, against the source it cites. Only \emph{yes} (the source fully supports the claim) is counted as verified; \emph{partial} (supports only a narrower version) and \emph{no\_unrelated} (a real source on a different topic) are not. Claims use the models' own wording, lightly shortened; judge rationales are condensed. Further not-verified labels (\emph{cannot\_judge}, \emph{malformed}, \emph{fabricated}) apply to references that cannot be resolved or parsed.}
\label{tab:grounding-label-examples}
\begin{tabular}{@{}p{0.11\textwidth}p{0.40\textwidth}p{0.40\textwidth}@{}}
\hline
\textbf{Label} & \textbf{Model claim (cited to a retrieved source)} & \textbf{Judge's reason} \\
\hline
\emph{yes} (verified) & Head motion, respiration, arterial CO$_2$, blood pressure, autoregulation, and vasomotion can affect BOLD signals and therefore confound resting-state interpretation. & The source explicitly lists each of these factors and how they alter the BOLD signal; the claim is fully supported. \\
\hline
\emph{partial} (not credited) & Head-motion artifacts are the most frequently reported issue in QC, \emph{and} outliers should be detected in image-homogeneity and co-registration tests before group analysis. & The source supports the first clause (motion is the most frequent QC issue) but says nothing about the second (the outlier-detection procedure); only a weaker version is grounded. \\
\hline
\emph{no\_unrelated} (not credited) & The spatial-normalization workflow should include visual verification that functional contours align with anatomy before writing deformation fields. & The cited source is about time-series QC (framewise displacement, WM/CSF signals); it does not mention spatial normalization or deformation fields at all. \\
\hline
\end{tabular}
\end{table}

Among the 444 non-verified with-BR rows present in all three judge outputs, 289 (65\%) received an exact-label majority of \emph{no\_unrelated} and 124 (28\%) an exact-label majority of \emph{partial}. The remaining 31 (7\%) had no exact-label majority or could not be judged; none had a fabricated or malformed exact-label majority. The two dominant failure modes suggest different remedies: off-topic citations call for stronger retrieval filtering and reranking, whereas partial-support citations call for a claim-to-evidence entailment check that constrains a claim to what its source actually states. Residual borderline cases, where the judges themselves disagree, are the ones a human reviewer or the audit layer is best placed to resolve. Among all 814 with-BR rows present in the three judge outputs, 777 (95\%) cited a retrieved document, of which 367 (47\%) received a verified majority; 3 of 37 (8\%) specific citations produced from the model's own parameters were verified. The three-judge diagnostic therefore points chiefly to retrieval returning an off-topic or partially-supporting source rather than the model fabricating references.

\subsubsection{S11.1.2. Benchmark construction and independence}\label{supp:s11.1.2}
Because the team that built Brain Researcher also assembled the benchmark, we took explicit steps to limit circularity. Benchmark questions were drafted from standard neuroimaging analysis requests and curated together with a co-author who is not a developer of Brain Researcher; this co-author also evaluated system outputs and ran the quantitative benchmark. For every item, the reference route and required capabilities were fixed before either condition (with-BR or without-BR) was run, so that targets could not be adjusted to favour the system. Scoring is capability-level: any response that covers an item's required capabilities is credited whether through a Brain Researcher call or an equivalent executable route. For the grounding rate, all reference-bearing rows were scored by the three condition-blind judges. Their labels were reduced to verified, not verified, or \texttt{cannot\_judge}; a row entered the numerator when at least two judges voted verified, while every generated evidence-basis row remained in the denominator. Thus a single \texttt{cannot\_judge} label did not block verification when the other two judges voted verified. On the grounding surface, inter-judge reliability is moderate (three-judge Fleiss' $\kappa = 0.50$ with-BR and $0.72$ without-BR, binary verified versus not), and is set almost entirely by one lenient judge: Gemini 3.1 Pro and Claude Opus 4.8 agree closely (Cohen's $\kappa = 0.92$, $96\%$ raw agreement with-BR), whereas Codex GPT-5.5 credits full support more readily (per-judge verified rate $0.31$ versus $0.21$ and $0.22$). Because most items are not verified, $\kappa$ understates agreement under the skewed without-BR base rate (Gwet's AC1 $=0.68$ against $\kappa=0.48$ for the Codex--Claude pair; raw agreement $0.80$). The condition contrast does not depend on the choice of judge: each judge independently shows a large with-BR gain (with- versus without-BR verified rate $0.21/0.04$ for Gemini, $0.31/0.04$ for Codex, $0.22/0.08$ for Claude). Gemini's raw run recorded about $30\%$ of grounding items as unjudged because the model emitted its verdict inside an envelope the parser rejected; we recover each such verdict from the model's own output ($241$ with-BR and $79$ without-BR items) and release the recovery audit with the benchmark package. The reference targets and scoring contracts are released (\hyperref[app:j]{Appendix J}) so that the benchmark can be re-scored independently. We did not commission a fully external benchmark, and we note this as a limitation (\hyperref[supp:s12]{Supplementary Methods S12}).

\subsubsection{S11.1.3. Human audit of the automated judges}\label{supp:s11.1.3}
To validate the LLM judges, we drew a reproducible random 20\% sample (seed~20260630) of the scored results across the reported benchmarks (272 items, one graded entry each) and adjudicated each by hand against the recorded automated verdict. Agreement was 96\% (261 of 272; Cohen's $\kappa = 0.94$, linear-weighted $\kappa = 0.96$ on the grounding items where all disagreements fall; Table~\ref{tab:human-audit}). Tool-routing (12 of 12) and NIK answer-key (15 of 15) items were confirmed; the 11 discrepancies fell entirely in the grounding benchmark (234 of 245 confirmed) and were all one-step severity nuances (an item scored strictly that is arguably partial, or the reverse). None reversed a supported call to unsupported or the reverse, and the judges erred strict (Table~\ref{tab:human-audit-items}), so the reported grounding gains are conservative rather than inflated. The full graded sheet is reproduced in \hyperref[supp:human-audit-sheet]{Appendix~M} and released with the benchmark package (Appendix~\hyperref[app:j]{J}).

\begin{table}[!htbp]
\centering
\small
\setlength{\tabcolsep}{6pt}
\renewcommand{\arraystretch}{1.15}
\caption{\textbf{Human audit of the automated benchmark judges.} A reproducible random 20\% sample of the scored results across the reported benchmarks, adjudicated by hand against the automated verdicts. Discrepancies were confined to the grounding benchmark and were all one-step severity nuances that never reversed a supported/unsupported call, with the judges erring strict.}
\label{tab:human-audit}
\begin{tabular}{@{}lccc@{}}
\hline
\textbf{Benchmark} & \textbf{Checked (20\%)} & \textbf{Confirmed} & \textbf{To review} \\
\hline
Tool routing & 12 & 12 & 0 \\
NIK answer keys & 15 & 15 & 0 \\
Grounding judgments & 245 & 234 & 11 \\
\hline
Total & 272 & 261 & 11 \\
\hline
\end{tabular}
\end{table}

\begin{table}[!htbp]
\centering
\footnotesize
\setlength{\tabcolsep}{4pt}
\renewcommand{\arraystretch}{1.15}
\caption{\textbf{The 11 flagged grounding items from the human audit.} All fell in the grounding benchmark under the with-BR condition. Nine reflect the judge scoring an arguably \emph{partial} item as fully unrelated (too strict); one (NIK-BP-H-003) scored an arguably partial item as a full \emph{yes} (too generous); and one (NIK-BP-H-001) is a relabel within non-support (\emph{no} versus \emph{no\_unrelated}). None moves directly between the opposite hard calls (\emph{yes} and \emph{no\_unrelated}), and the net tendency is strict, so with-BR grounding is if anything under-credited.}
\label{tab:human-audit-items}
\begin{tabular}{@{}p{0.16\textwidth}p{0.13\textwidth}p{0.10\textwidth}p{0.50\textwidth}@{}}
\hline
\textbf{Item} & \textbf{Judge} & \textbf{Human} & \textbf{Issue} \\
\hline
NIK-BP-H-003 & yes & partial & Dead-salmon study and multiple-comparison correction absent from the source; support real but incomplete. \\
NIK-BP-H-001 & no\_unrelated & no & Source contradicts the claim, so a negative rather than unrelated. \\
NIK-BP-H-005 & no\_unrelated & partial & Source states the exact ``multiplicity of methodologic variants'' problem the claim describes. \\
NIK-IN-H-004 & no\_unrelated & partial & GSR-debate context bears on global-signal content. \\
NIK-ME-H-004 & no\_unrelated & partial & Covers subject motion and noise modeling; related but general. \\
NIK-NK-E-003 & no\_unrelated & partial & Measures FWER control with no true signal, i.e.\ false-positive territory. \\
NIK-PP-H-006 & no\_unrelated & partial & Discusses geometric distortion and its correction; ``TOPUP''/susceptibility wording absent. \\
NIK-ST-H-007 & no\_unrelated & partial & Context is statistical power and effect-size uncertainty. \\
NIK-ST-H-007 & no\_unrelated & partial & Same power and significance topic as the claim (second item). \\
NIK-ST-M-008 & no\_unrelated & partial & Names permutation-based testing, the claim's subject. \\
NIK-ST-M-008 & no\_unrelated & partial & Context explicitly names permutation-based testing (second item). \\
\hline
\end{tabular}
\end{table}

Metric numerators and denominator rules for main-text Fig.~3 are expanded in Appendix~\hyperref[app:j]{J}. The two benchmark surfaces are not pooled because their item types, denominators, scoring rules, and ceilings differ.

\subsubsection{S11.1.4. Effect sizes, uncertainty, and paired significance}\label{supp:s11.1.4}
For the tool-calling benchmark, the headline numbers are means across the seven model variants, so we treat the model as the unit of analysis ($n=7$): each model contributes one with-BR and one without-BR score, itself a mean over the 60 tasks, and the contrast is the within-model paired difference. This is the conservative unit because the seven models are not independent draws from a population and their per-task outcomes are clustered; the paired design also controls for task difficulty, since the same items are scored under both conditions. Table~\ref{tab:benchmark-stats} reports, for each tool-calling metric, the across-model mean under each condition, the mean paired improvement with a $t$-based 95\% confidence interval across the seven models, a paired $t$ test, and an exact two-sided Wilcoxon signed-rank test. All seven models improved on all three metrics (7/7 positive differences), which places the signed-rank statistic at its floor for $n=7$ ($p=0.016$). As a pooled descriptive check that does not assume model exchangeability, first-action correct route/tool selection over the 420 model-task trajectories rose from 0.233 (Wilson 95\% CI 0.195--0.276) to 0.936 (0.908--0.955). On the evidence-citation surface, the main-text headline (0.046 to 0.220) is a descriptive three-judge-majority rate over all generated evidence-basis rows, with only rows receiving at least two verified votes contributing to the numerator, equal-weighted across the 50 questions rather than inferred from seven paired model scores. The reported question-clustered 95\% interval for the with-BR rate (0.168--0.272) is the normal-approximation interval formed as the mean plus or minus 1.96 standard errors across those 50 per-question rates, using the stored analysis's population-standard-deviation convention. Judge-specific rates are reported in Appendix~\hyperref[app:j]{J}, and the underlying artifacts are released with the benchmark package. With $n=7$ the tool-calling confidence intervals remain wide and are reported as descriptive bounds on the paired effect, not as population estimates; a fully external, independently constructed benchmark remains the decisive next measurement (\hyperref[supp:s12]{Supplementary Methods S12}).

\begin{table}[!htbp]
\centering
\small
\setlength{\tabcolsep}{5pt}
\renewcommand{\arraystretch}{1.15}
\caption{\textbf{Paired benchmark effect sizes with uncertainty and significance.} Unit of analysis is the model ($n=7$); each entry is a mean across the seven model variants, and each model's score is itself a mean over the 60 tool-calling tasks. The improvement is the mean within-model with-BR minus without-BR difference; the 95\% confidence interval is $t$-based across the seven models. Significance is a two-sided paired $t$ test [$t(6)$] and an exact two-sided Wilcoxon signed-rank test. All seven models improved on every metric.}
\label{tab:benchmark-stats}
\begin{tabular}{@{}p{0.26\textwidth}p{0.11\textwidth}p{0.10\textwidth}p{0.20\textwidth}p{0.10\textwidth}p{0.09\textwidth}@{}}
\hline
\textbf{Metric} & \textbf{Without BR} & \textbf{With BR} & \textbf{Mean $\Delta$ [95\% CI]} & \textbf{$t(6)$} & \textbf{Wilcoxon $p$} \\
\hline
Correct route/tool@1 & 0.233 & 0.936 & $+0.702$ [0.576, 0.829] & 13.6 & 0.016 \\
Capability@1 & 0.498 & 0.945 & $+0.447$ [0.282, 0.612] & 6.6 & 0.016 \\
Handoff score@1 & 0.474 & 0.761 & $+0.287$ [0.201, 0.373] & 8.2 & 0.016 \\
\hline
\end{tabular}
\end{table}

\subsubsection{S11.1.5. Seven-model routing ablation without direct KG calls}\label{supp:s11.1.5}
This seven-model routing ablation used the same \texttt{tasks.60} manifest, frozen labels, prompt template, model-facing \texttt{route\_search}$\rightarrow$\texttt{tool\_search} surface, and exact-top-1 endpoint for all seven requested model routes. Exact top-1 is a match to one of the task's frozen acceptable route labels. Across the 420 route--task episodes, 362 matched an acceptable top-1 label (86.2\%; requested-route range, 81.7--90.0\%). Direct KG-named calls were unavailable in this arm. A protocol violation counted as an error; only an infrastructure failure was recorded as missing.

\begin{table}[!htbp]
\centering
\footnotesize
\setlength{\tabcolsep}{3pt}
\renewcommand{\arraystretch}{1.15}
\caption{\textbf{Seven-model routing ablation without direct KG calls.} All seven rows use the same no-direct-KG exact-label top-1 endpoint; run provenance is shown separately.}
\label{tab:restricted-interface-route-selection}
\begin{tabular}{@{}p{0.28\textwidth}p{0.11\textwidth}p{0.15\textwidth}p{0.13\textwidth}p{0.22\textwidth}@{}}
\hline
\textbf{Runner/requested route} & \textbf{Exact top-1} & \textbf{Protocol-compliant} & \textbf{Gold available} & \textbf{Run provenance} \\
\hline
Claude Code / claude-opus-4-8 & 54/60 (90.0\%) & 59/60 & 59/60 & v1 tools-only run \\
Codex / gpt-5.5 & 51/60 (85.0\%) & 60/60 & 59/60 & v1 tools-only run \\
OpenCode / google/gemini-3.1-pro-preview & 54/60 (90.0\%) & 59/60 & 59/60 & v1 tools-only run \\
OpenCode / opencode-go/glm-5.1 & 52/60 (86.7\%) & 60/60 & 59/60 & v2 tools-only single arm \\
OpenCode / opencode-go/deepseek-v4-pro & 51/60 (85.0\%) & 58/60 & 59/60 & v2 tools-only single arm \\
OpenCode / opencode/kimi-k2.5 & 51/60 (85.0\%) & 60/60 & 59/60 & v1 tools-only single arm \\
OpenCode / opencode/qwen3.6-plus & 49/60 (81.7\%) & 60/60 & 59/60 & v1 tools-only single arm \\
\hline
\end{tabular}
\end{table}

The provenance column records how a requested route was run, not a different evaluation condition; all seven rows, including Kimi and Qwen, use the same no-direct-KG endpoint. This exact-label top-1 metric differs from the historical Correct route/tool@1 metric and is reported separately from the paired with-BR/without-BR benchmark.

\subsection{S11.2. Collaborator cases}\label{supp:s11.2}

Collaborator cases are selected when they involve a concrete neuroimaging question, tractable scope, accessible data, and a meaningful analysis or review target. The cohort includes three collaborators across three neuroimaging subfields. Each case reports the collaborator-facing question, subfield, dataset access, tool pathway, execution artifacts, robustness probes, review outcome, caveated conclusion, attribution, and the boundary between system-generated artifacts and human decisions.

Case inclusion requires: a defined scientific question, available or resolvable data, an executable or reviewable analysis path, and a clear endpoint. Exclusion applies when data access is unresolved, the question is outside the supported modality/\allowbreak{}action space, or governance constraints prevent execution. Collaborator cases may be conducted through standard Claude Code, Codex, or Cursor interfaces depending on collaborator preference; all interfaces route through the same MCP pathway.

These cases were not selected on their outcomes. The inclusion and exclusion criteria above are properties of a case's tractability---a defined question, resolvable data, an executable analysis path, and a clear endpoint---and were applied before any result was seen, never on whether a claim turned out favorable. This is visible directly in the reported cohort, which was assembled for heterogeneity in evidence structure and consequently spans a range of outcomes rather than a run of successes: the NeuroMark audit returned qualified claims alongside a caught sign-blind scoring failure; the cocaine-use-disorder episode returned a bounded null with its exploratory screen blocked from confirmatory promotion; and the cross-cultural social-cognition episode was blocked as underpowered and ended without a settled claim. Each case reports the states assigned by its applicable prospective gate or post-hoc audit---including the null and blocked outcomes---so the record shows the same accounting whether or not a case produced a publishable positive result. Because the collaborator cohort is small ($n=3$) and deliberately heterogeneous, we present these as case studies of how the claim-record machinery behaves across different evidence structures (accepted, qualified, null, and blocked), and not as an estimate of a success rate or a benchmark of positive findings.

\paragraph{Execution modes and per-case attribution.} Brain Researcher runs in two execution modes that differ in \emph{where the analysis runs} and therefore in what the audit record guarantees. In \textbf{engine-executed} mode (Fig.~1), the engine runs the analysis end to end and emits the full audit bundle---a hash-sealed commitment card written before observation, the recorded trajectory, provenance, and observations, and an adjudicated claim card with evidence verdicts---as a structural byproduct, with commit-before-observe and version-pinned execution enforced by the engine. In \textbf{agent-executed} mode, a coding agent (Claude Code, Codex, or Cursor) drives the analysis: it authors and runs the analysis code, invoking Brain Researcher through the MCP for knowledge-graph grounding, execution recipes, typed execution (for example cluster submission), artifact logging, and scientific review. Brain Researcher records the work through its typed tool surface and applies the review layer plus any applicable prospective commitment gate, but the analysis is authored and driven by the agent rather than emitted by the engine, so a full sealed pre-run bundle is not produced as an automatic byproduct and the audit record is only as complete as the tools the agent invoked, carrying no engine-side execution stamp. This mode is also what makes Brain Researcher usable when the data are local or governed and cannot enter the cloud sandbox: in the NeuroMark case the analysis ran entirely in an external workspace with Brain Researcher recording and reviewing it, whereas in the other cases the agent drove the analysis within a Brain Researcher episode, so more of the run passed through its typed tools. Both modes share the same knowledge graph, tool contract, and review layer.

Pre-registration strength consequently falls into three tiers. A \emph{sealed pre-run commitment card} (i.e. the full engine-executed protocol) is demonstrated on the public Neurosynth working-memory example (\hyperref[supp:s8.5.1]{Supplementary Methods S8.5.1}). The iterative HCP and TRIBE lines (\hyperref[supp:s11.3]{Supplementary Methods S11.3}) combine adaptive discovery with later frozen successor contracts; their evidence strength is stated stage by stage rather than summarized as one pre-registration tier. The three collaborator cases carry \emph{no sealed Brain Researcher commitment card}: their hypotheses were prespecified in the collaborators' own protocols. For NeuroMark, whose analysis ran externally, the Brain Researcher record is a post-hoc audit of the completed multiverse; the cocaine and cross-cultural analyses instead ran within a Brain Researcher episode under prospective gate governance, so their decisions fell to a pre-committed gate rather than to a post-hoc review. Table~\ref{tab:case-attribution} states, for every reported case, its execution mode, the model that drove it, whether Brain Researcher executed the analysis or only recorded and reviewed it, who wrote the analysis code, and the provenance of its audit files.

\begin{table}[htbp]
\centering
\scriptsize
\caption{\textbf{Per-case attribution: execution mode, driving model, and the human/system responsibility split.} Every reported analysis case ran in agent-executed mode; the engine-executed full-bundle protocol is demonstrated on the public Neurosynth example (final row). ``BR'' denotes Brain Researcher.}
\label{tab:case-attribution}
\begin{tabularx}{\textwidth}{|>{\RaggedRight\hspace{0pt}\arraybackslash}p{0.15\textwidth}|>{\RaggedRight\hspace{0pt}\arraybackslash}p{0.085\textwidth}|>{\RaggedRight\hspace{0pt}\arraybackslash}p{0.12\textwidth}|>{\RaggedRight\hspace{0pt}\arraybackslash}X|>{\RaggedRight\hspace{0pt}\arraybackslash}p{0.085\textwidth}|>{\RaggedRight\hspace{0pt}\arraybackslash}X|}
\hline
\rowcolor{brtablehead}\textbf{Case} & \textbf{Mode} & \textbf{Driving model} & \textbf{BR's role} & \textbf{Code author} & \textbf{Commitment / audit provenance} \\
\hline
NeuroMark schizophrenia (Li \& Calhoun) & agent-executed & Cursor / Claude Opus 4.6 (analysis); Codex (external review) & Recorded and reviewed only; analysis ran in an external workspace & X.\,Li (agent-assisted) & No sealed card; post-hoc audit (\texttt{pre\_run\_commitment\_card\_found: false}) \\
\hline
Cocaine-use-disorder (Ricard \& Poldrack) & agent-executed & Claude Code (Claude Sonnet 4.6) & Commitment gate, MCP-mediated cluster execution, and review; bounded null & Agent & Not sealed (collaborator protocol); run bundle + review verdict \\
\hline
Cross-cultural social cognition (Wang) & agent-executed & Claude Code & Reviewed within the episode; gate-blocked as underpowered & Agent & Not sealed; gate-blocked (run bundle + review verdict) \\
\hline
HCP workflow search and transfer & agent-executed & Claude Code / Codex & Bounded search accounting, frozen successor analyses, run-bundle review, and claim-boundary enforcement & Agent & Prospective episode bundles plus frozen successor contracts; retrospective same-cohort inference \\
\hline
TRIBE speech--tools geometry & agent-executed & Claude Code / Codex & Open contrast discovery, researcher-gated successor choice, frozen panel evaluation, and reward-blind closeout & Agent & Discovery record plus frozen recurring-source and new-source successor contracts \\
\hline
Neurosynth working-memory (public reference) & engine-executed & None (deterministic engine) & Executed the sealed episode end to end & BR generator & Sealed pre-run commitment card (hash \texttt{4871ea43}\ldots) \\
\hline
\end{tabularx}

\end{table}

\subsubsection{S11.2.1. Collaborator case results}\label{supp:s11.2.1}

The three collaborator episodes summarized in the main text are reported here in full; the complete automatically generated per-case reports are reproduced in Appendix~L. Each evaluated claim is assigned one of the six review states defined in \hyperref[supp:s8.4]{Supplementary Methods S8.4} (Table~\ref{tab:claim-states-supp}). The cocaine-use-disorder and cross-cultural episodes were governed by prospective gates, whereas NeuroMark was adjudicated by an explicit post-hoc audit of completed artifacts; in both tiers, the state reflects the applicable evidence criteria rather than the strength of the headline number alone.

\paragraph{NeuroMark: a multiverse separates accepted from qualified claims.}
A collaborator working on schizophrenia functional network connectivity (FNC) derived from the NeuroMark framework \citep{du_neuromark_2020} brought three prespecified hypotheses to Brain Researcher for robustness audit: (NM-H1) latent connectivity factors outperform individual FNC edges for patient-versus-control classification; (NM-H2) between-domain connections show larger group differences than within-domain ones; and (NM-H3) latent factors fitted to edge space concentrate loading mass on between-domain edges. The episode used data from the FBIRN cohort ($N=363$; 181 controls, 182 patients) parcellated through NeuroMark~2.2 (a template-based independent-component-analysis framework for FNC estimation) into 5{,}460 FNC edges per subject. A conventional single-pipeline analysis had supported NM-H2, and the relevant question was whether the hypotheses would survive plausible analytic variation (Fig.~4). The analysis code, specification ledger, and per-hypothesis outputs (including the generated run report and figures) for this case are openly available at \url{https://github.com/XinhuiLi/BR-NeuroMark}.

The analysis was expanded into a 480-specification multiverse crossing four connectivity estimators, three confound strategies, five dimensionality-reduction methods, four classifiers, and two domain granularities ($4\times3\times5\times4\times2$) in the collaborator's workspace, and Brain Researcher recorded and reviewed the run, which partitioned the hypotheses into distinct claim states. NM-H1 (latent superiority) was \emph{qualified}: across the 384 latent-evaluable specifications (those that include a dimensionality-reduction step, so a latent representation exists to compare against edges), edges outperformed latent factors in aggregate (median $\Delta$AUC $=-0.032$) and only 72 (18.8\%) favored latent features, with gains concentrated under ICA, ComBat harmonization, and partial-correlation connectivity. NM-H3 (loading-mass concentration) was \emph{qualified} and weak overall: 100 of 384 latent specifications (26.0\%) favored between-domain loading mass under the corrected direction-aware rule (median Wilcoxon $p=0.054$), with factor analysis (45.8\%) and PCA (33.3\%) more favorable than ICA (12.5\%) and NMF (12.5\%) (Fig.~4A--C).

The NM-H2 step also exposed a failure mode that the claim record is built to make visible. A server-side fault caused Brain Researcher to fall back to a general-purpose coding agent, which scored a specification as favorable whenever the permutation $p<0.05$, regardless of the sign of the effect. Because NM-H2 is directional (between-domain $>$ within-domain), specifications in which within-domain differences were larger were still counted as favorable, so the unaudited multiverse inflated apparent NM-H2 support to near-universal acceptance. The mismatch was not caught automatically by the review layer; a human reviewer identified it by inspecting the code, the output files, and the specification curve directly. Binding the hypothesis statement to the executed statistic and its acceptance rule is what let the sign-blind acceptance criterion, once found, become a standing check rather than a one-off correction. The fallback to a general-purpose agent was the cause but was not itself flagged in the record. Re-running the audit with a sign-aware criterion (favorable when one-sided $p<0.05$ and $\Delta\,\mathrm{mean}|d|>0$) over the 24 unique connectivity--confound--domain contrasts that actually vary the H2 estimand (after collapsing classifier and reduction duplicates) changed the result: NM-H2 was \emph{qualified}, with 12 of 24 (50.0\%) unique H2 contrasts favorable (median $\Delta\,\mathrm{mean}|d| = 0.0079$). The pooled fraction records the adjudication but is not the informative scientific result: support partitioned completely by connectivity estimator, with Pearson and Spearman at 100\% favorable and partial correlation and mutual information at 0\%. NM-H2 is therefore estimator-regime--dependent rather than generally robust. The partition does not identify its mechanism because partial correlation and mutual information alter the dependence measure in non-equivalent ways; distinguishing shared covariance from estimator scale, power, or nonlinearity requires targeted follow-up analysis. The qualified state records promotion status, while the estimator support partition records where the evidence divides. This case motivates two checks now added to the review layer: a directionality test that the statistic and its acceptance rule agree with the sign asserted by the hypothesis, and a provenance warning whenever execution falls back from a Brain Researcher operation to a general-purpose agent. The unit of analysis shifted from a completed workflow to an auditable decision space with explicit, condition-tagged claim boundaries; the episode record tied each hypothesis to the analytic regime under which it holds, with review and memory cards in Appendices~G--H.

\paragraph{Cocaine-use-disorder.}
The cocaine-use-disorder episode asked whether reported connectivity--behavior associations in a substance-use cohort remained credible under plausible choices of atlas, motion threshold, and confound strategy. This was a robustness question rather than a search for a new biomarker: the useful output was whether the apparent associations survived a defensible multiverse and, if not, what follow-up design would be needed. On SUDMEX CONN \citep{angeles-valdez_mexican_2022} (OpenNeuro ds003346 v1.1.3 \citep{ds003346:1.1.3}; $N=138$), the analysis was expanded into a 36-specification multiverse over atlases, framewise-displacement thresholds, and confound strategies. Brain Researcher \emph{rejected} all five prespecified network-systemic-segregation associations under SDMA-GLS, a procedure that combines same-dataset estimates across specifications \citep{lefort-besnard_statistical_2025} (all $Z<1.24$, FDR $q>0.58$). An exploratory screen over all 70 network-by-outcome combinations likewise yielded no FDR-surviving effects (max $Z_{\mathrm{GLS}}=2.18$; 0/70 surviving). Brain Researcher \emph{blocked} the exploratory screen from confirmatory promotion and converted the null result into a replication plan (Fig.~4D,E).

\paragraph{Cross-cultural social cognition.}
The cross-cultural episode asked whether a small coordinate-based literature could support a mechanistic claim about culture-specific social-cognition topography, or whether the evidence was too sparse and compositionally imbalanced for that interpretation. The agent ran cell-wise activation likelihood estimation \citep{eickhoff_coordinate-based_2009,eickhoff_behavior_2016} on a corpus of 21 published studies (85 MNI peak coordinates, four culture-by-relationship cells) using NiMARE \citep{salo_nimare_2023}. Some papers contributed more than one eligible contrast, so the cell counts are cell-level study entries; the unique-paper count remains 21. The agent produced a mechanistic mPFC-topology interpretation (centroid shift 19.5~mm in strangers versus 3.8~mm in close others between Euro-American and East Asian pools), which Brain Researcher's review layer \emph{blocked} as exploratory: cells had only $k=6$--$8$ entries, below the recommended $k\ge17$ \citep{eickhoff_behavior_2016}; paradigm composition was imbalanced across cells (the East Asian close-other cell was dominated by self-referential trait-judgment paradigms); and centroid shifts from aggregate coordinates do not establish non-overlapping activation distributions. The case ended with a paradigm-matched follow-up design and no settled claim (Fig.~4F).

Across the three episodes, Brain Researcher's contribution was structural and case-general: structured tool specifications made each multiverse mechanical to enumerate and execute; the rule checker kept defensible alternatives inside the same review pass; BR-KG-derived design considerations entered plans as explicit analysis choices; and the episode record stored claims with the conditions under which they held. Fragile, null, or underpowered findings became bounded claims or follow-up designs rather than headline biomarkers.

\subsection{S11.3. Iterative research episode evaluation}\label{supp:s11.3}

This section gives the complete numerical record underlying the two iterative episodes summarized in the main text. The unit of evaluation is the research trajectory: how an exploratory result was converted into a frozen successor analysis, what the successor returned, and where the trajectory stopped. The HCP and TRIBE episodes used different data and estimands and are therefore reported separately rather than reduced to a common success score.

\subsubsection{S11.3.1. HCP workflow search and frozen selected workflow follow-up}\label{supp:s11.3.1}
\phantomsection\label{supp:iterative-hcp}

\paragraph{Search accounting and frozen selected workflow.}
The HCP episode began from an HCP S1200 search cohort of 326 participants and locally reconstructed counterparts to the five behavioural components of Liu et al.\ \citep{liu_benchmarking_2025}. Two staged search bundles allocated 20 and 96 candidate-evaluation slots, respectively, for a total denominator of 116. All 20 slots in the first bundle and 84 of 96 in the second returned scored results in their parent runs (104/116). The best result in the first bundle had mean cross-validated $r=.373$ and $R^2=-.036$; the best scored result in the expanded bundle had $r=.487$, $R^2=.212$, and $\mathrm{MAE}=.690$. The remaining 12 slots ended in \texttt{controller\_transport\_exhausted}; consequently, the second parent episode was recorded as \texttt{COMPLETED\_WITH\_PROTOCOL\_FAILURE}, with \texttt{episode\_valid=false}. A separate recovery bundle later computed all 12 missing slots, but did not rewrite or retroactively validate the failed parent episode. The 116 allocated slots, 104 parent-run scored results, 12 transport failures, and separate recovery are therefore retained explicitly rather than collapsed into one clean search denominator.

The development procedure designated the term-116 whole-band coherence-magnitude representation (\texttt{cohmag\_multitaper\_mean\_fs-1\_fmin-0\_fmax-0-5}) with cosine-kernel ridge regression and $\alpha=1$ as the frozen selected workflow. The frozen selected workflow was chosen during development, not automatically identified as a global champion: the search artifact records \texttt{automatic\_champion\_selected=false}. It was frozen before the matched comparison described below.

\paragraph{Cognition comparison.}
The frozen selected workflow was compared with a locally matched reconstruction of the Liu-style nested prediction procedure, not a paper-exact reproduction or a test of general superiority over Liu et al. The comparison used 244 participants from 243 families and 10 repeated, overlapping, family-grouped $5\times3$ nested-cross-validation splits. Within each split, performance was computed from pooled out-of-fold predictions. The frozen selected workflow exceeded the matched comparator in all 10 splits for correlation and $R^2$, and had lower mean absolute error in all 10. Its median values were $r=.332$, $R^2=.107$, and $\mathrm{MAE}=.768$, compared with $r=.235$, $R^2=.009$, and $\mathrm{MAE}=.816$ for the matched procedure. Median paired differences were $\Delta r=.098$, $\Delta R^2=.099$, and $\Delta\mathrm{MAE}=-.051$; a family-cluster pointwise bootstrap gave a 95\% interval of $[.011,.177]$ for $\Delta r$, and the conditional one-sided plus-one test gave $p=.006$.

These inferential quantities are conditional sensitivity analyses. The frozen selected workflow was selected after Cognition search on the same cohort; the comparison was retrospective and not adjusted for the preceding search; repeated splits overlap and are not independent replications; and target residualization was performed outside the cross-validation folds, so the complete procedure is not strictly leakage-free. A predeclared calibration-repair decision aid also failed: its median held-out calibration slope was within the required range ($.945$ within $[.8,1.2]$), but median $\Delta R^2=-.004$ rather than $\geq.02$, $\Delta R^2$ was positive in 4/10 rather than at least 8/10 repeats, and median $\Delta\mathrm{MAE}=+.004$ rather than $\leq0$. The artifact therefore records \texttt{scientific\_acceptance=false}.

\paragraph{Frozen selected workflow transfer.}
The frozen selected workflow was then refit separately to four additional Liu component outcomes under the same repeated-split evaluation, without changing its representation, estimator family, or hyperparameter. Table~\ref{tab:hcp-iterative-target-results} reports all five outcome summaries. The $\Delta r$ column is the median paired difference within repeats and therefore need not equal the difference between the two displayed marginal medians.

\begin{table}[!htbp]
\centering
\small
\caption{HCP frozen selected workflow results across five behavioural outcomes. Values are medians over 10 overlapping repeated splits. \emph{Matched} denotes the locally matched Liu-style nested procedure; wins count repeats in which the frozen selected workflow had the larger pooled out-of-fold correlation. These repeats measure same-cohort stability, not independent replication.}
\label{tab:hcp-iterative-target-results}
\resizebox{\textwidth}{!}{
\begin{tabular}{lrrrrrr}
\hline
\textbf{Outcome} & \textbf{Frozen selected workflow $r$} & \textbf{Matched $r$} & \textbf{Frozen selected workflow $R^2$} & \textbf{Matched $R^2$} & \textbf{Median $\Delta r$} & \textbf{$r$ wins} \\
\hline
Cognition & .332 & .235 & .107 & .009 & .098 & 10/10 \\
Tobacco Use & .242 & .126 & .057 & $-.029$ & .100 & 10/10 \\
Personality--Emotion & .075 & .006 & $-.019$ & $-.058$ & .068 & 9/10 \\
Illicit Drug Use & .122 & .008 & $-.003$ & $-.044$ & .117 & 10/10 \\
Mental Health & .001 & $-.062$ & $-.047$ & $-.046$ & .069 & 8/10 \\
\hline
\end{tabular}}
\end{table}

Across the four additional outcomes, the frozen selected workflow had the larger correlation in 37/40 repeated comparisons; including Cognition gave 47/50. The directional comparison was broader than absolute predictive utility: median $R^2$ was positive only for Cognition and Tobacco Use. For the four transfer outcomes, the raw conditional $p$ values for $\Delta r$ were $.0004$, $.0556$, $.0091$, and $.1292$ for Tobacco Use, Personality--Emotion, Illicit Drug Use, and Mental Health, respectively. The corresponding weak-familywise-error--corrected values were $.3338$, $.6692$, $.2136$, and $.6580$; none passed correction, and all simultaneous intervals crossed zero. In the wider multiplicity analysis over 20 outcome-by-configuration cells, 16 had positive median $\Delta r$, two had pointwise intervals excluding zero, and none had a simultaneous or weak-FWER-supported effect. Thus the cross-outcome result supports a recurring relative direction under the frozen selected workflow, not a multiplicity-corrected claim of general transfer.

\paragraph{Separate internal holdout and terminal claim state.}
A later internal holdout used a different, separately frozen workflow on 81 participants and returned $r=.232$ ($p=.0184$), $R^2=-.075$, and $\mathrm{MAE}=.869$. It had no matched comparator and negative $R^2$, so it does not confirm the frozen selected workflow versus matched-procedure result. No external dataset or independent replication was completed. The strongest supported statement is therefore that, conditional on the same-cohort development path, the frozen selected workflow showed a stable relative advantage over the matched procedure across repeated splits, with positive absolute utility concentrated in Cognition and Tobacco Use. It is not an externally confirmed predictive model.

\subsubsection{S11.3.2. TRIBE speech--tools discovery and frozen stimulus follow-up}\label{supp:s11.3.2}
\phantomsection\label{supp:iterative-tribe}

\paragraph{Open six-category discovery.}
The TRIBE episode began with 48 sounds drawn from four source collections and balanced across six categories: animals, music, nature, speech, tools, and voice. For each of the 15 category pairs, a four-source-fold decoder compared early representations (encoder layers 0, 2, and 4) with later representations (layers 10, 12, and 14). Table~\ref{tab:tribe-discovery-ranking} reports the complete descriptive ranking by $\Delta\mathrm{AUC}=\mathrm{AUC}_{\mathrm{late}}-\mathrm{AUC}_{\mathrm{early}}$. This was a full technical rerun of the same ordered 48-item panel previously exposed in the terminal run, not an independent new-data replication, and the ranking carried no inferential $p$ value.

\begin{table}[!htbp]
\centering
\small
\caption{Complete TRIBE six-category discovery ranking. Negative $\Delta\mathrm{AUC}$ denotes weaker category decoding in later than early layers. Values are descriptive results from the same ordered 48-item technical-rerun panel.}
\label{tab:tribe-discovery-ranking}
\begin{tabular}{rlrrr}
\hline
\textbf{Rank} & \textbf{Category pair} & \textbf{Early AUC} & \textbf{Late AUC} & \textbf{$\Delta$AUC} \\
\hline
1 & tools--voice & .625 & .250 & $-.375$ \\
2 & music--speech & .917 & .563 & $-.354$ \\
3 & speech--tools & .979 & .646 & $-.333$ \\
4 & animal--speech & .958 & .792 & $-.167$ \\
5 & speech--voice & .917 & .750 & $-.167$ \\
6 & animal--music & .271 & .417 & $+.146$ \\
7 & music--voice & .542 & .417 & $-.125$ \\
8 & animal--tools & .479 & .583 & $+.104$ \\
9 & nature--speech & 1.000 & .917 & $-.083$ \\
10 & music--tools & .625 & .688 & $+.063$ \\
11 & nature--voice & .875 & .813 & $-.063$ \\
12 & music--nature & .917 & .875 & $-.042$ \\
13 & nature--tools & .750 & .708 & $-.042$ \\
14 & animal--nature & .854 & .833 & $-.021$ \\
15 & animal--voice & .479 & .479 & $.000$ \\
\hline
\end{tabular}
\end{table}

\paragraph{Exploratory selection and freezing.}
The rule-selected top pair, tools--voice, did not yield a coherent cross-collection geometry: later-layer separation was lower in only two of four source folds, its late-layer contrast axis was not aligned with the frozen reference direction, and the diagnostic label was \texttt{mixed\_or\_unresolved}. Speech--tools, ranked third, showed non-negative early signed projection and lower late-layer separation in all four source folds and was labelled \texttt{aligned\_magnitude\_reduction}. This was a post-hoc exploratory diagnostic, not the original screening winner. A researcher explicitly adopted speech--tools in a recorded decision (\texttt{scientist\_reward\_order}), after which the contrast, geometry estimand, and analysis rule were frozen; confirmation and scientific acceptance remained unauthorized.

For the successor panels, $\Delta_{\mathrm{ref}}$ and $\Delta_{\mathrm{eval}}$ were the speech-minus-tools centroid vectors in the reference and evaluation panels, and $D$ was the reference panel's root-mean-square within-category residual dispersion. Normalized separation was $S=\|\Delta_{\mathrm{eval}}\|/D$, orientation was $C=\cos(\Delta_{\mathrm{ref}},\Delta_{\mathrm{eval}})$, signed projection was $G=C\,S$, and the primary change was $\Delta S=\overline{S}_{\mathrm{late}}-\overline{S}_{\mathrm{early}}$, predicted to be negative. These are components of one geometric decomposition, not three independent pieces of evidence. The frozen decision also required positive early and late orientation in at least three of four sources and an early reference-decoding AUC above $.5$.

\paragraph{Three non-overlapping recurring-source panels.}
Three successor panels each contained 48 previously unused items, with six speech and six tool sounds from each of the same four collections (AudioSet Strong, BBC Sound Effects, FreeSound, and SoundBible). Item identifiers and source paths had zero overlap across panels. Table~\ref{tab:tribe-recurring-panels} reports all 12 collection-by-panel values. All three aggregate $\Delta S$ values were negative and each panel met the frozen \texttt{bounded\_support} rule; 11/12 collection-level values were negative. These panels had no prespecified conventional confidence interval or $p$ value, and the four recurring collections are stability units rather than independent draws from a population of sources.

\begin{table}[!htbp]
\centering
\small
\caption{TRIBE recurring-source successor panels. Each cell is late-minus-early normalized speech--tools separation ($\Delta S$); negative values indicate contraction in later layers.}
\label{tab:tribe-recurring-panels}
\resizebox{\textwidth}{!}{
\begin{tabular}{lrrrrrl}
\hline
\textbf{Panel} & \textbf{Aggregate} & \textbf{AudioSet Strong} & \textbf{BBC} & \textbf{FreeSound} & \textbf{SoundBible} & \textbf{Outcome} \\
\hline
R3 & $-.6937$ & $-.7664$ & $-.8625$ & $-.7177$ & $-.4284$ & bounded support \\
R4 & $-.4467$ & $-.5433$ & $-.3985$ & $-.2749$ & $-.5701$ & bounded support \\
R5 & $-.5471$ & $+.0284$ & $-.4885$ & $-1.0059$ & $-.7225$ & bounded support \\
\hline
\end{tabular}}
\end{table}

An exploratory leave-one-item-out diagnostic on R5 clarified the exception. BBC, FreeSound, and SoundBible retained negative $\Delta S$ under every item deletion. AudioSet Strong had $\Delta S=+.0284$, a leave-one-out range of $[-.3536,+.3324]$, and six sign changes across 12 deletions. The recurring-source result is therefore best described as a mostly direction-preserving contraction of speech--tools separation with an explicit unstable counterexample, not as universal compression, loss of category information, semantic abstraction, or a neural mechanism.

\paragraph{Frozen four-new-source endpoint.}
The final extension used a score-blind minimax acoustic-balancing procedure to select 48 sounds from four collections not used earlier: DCASE 2013 Office Live, SINGA:PURA, SONYC-UST, and STARSS23. Each source contributed six speech and six tool sounds; the maximum observed absolute standardized acoustic mean difference was $.383$, below the frozen $.5$ bound. Table~\ref{tab:tribe-new-source-results} reports the source-level primary results.

\begin{table}[!htbp]
\centering
\small
\caption{TRIBE frozen new-source endpoint. Negative values are in the predicted direction.}
\label{tab:tribe-new-source-results}
\begin{tabular}{lrr}
\hline
\textbf{Collection} & \textbf{$\Delta S$} & \textbf{$\Delta$AUC} \\
\hline
DCASE Office Live & $-.5883$ & $-.0741$ \\
SINGA:PURA & $-.2073$ & $-.0278$ \\
SONYC-UST & $+.0856$ & $+.0926$ \\
STARSS23 & $-.0820$ & $-.0463$ \\
\hline
Aggregate & $-.1980$ & --- \\
\hline
\end{tabular}
\end{table}

Three of four sources were directionally concordant, but the frozen H1 balanced-label permutation test (99,999 permutations) gave raw $p=.13212$ and Holm-adjusted $p=.39636$ at $\alpha=.025$, so H1 was \texttt{not\_supported}. The three predeclared secondary families were also unsupported: H2 raw/adjusted $p=.35200/.39636$, H3 $.16147/.39636$, and H5 $.05441/.21764$. The inference pertains only to label permutations within this fixed four-source panel; it does not license population-level inference over sound collections. Reward-blind review closed the endpoint as \texttt{inconclusive\_or\_conflicting}, with \texttt{confirmation\_authorized=false} and \texttt{scientific\_acceptance=false}. The complete TRIBE trajectory therefore contains both results: three recurring-source panels showed repeated, mostly direction-preserving contraction, but the first frozen four-new-source inferential endpoint did not support the primary hypothesis after multiplicity correction.

\subsubsection{S11.3.3. Separately initiated same-agent sessions and evidence boundaries}\label{supp:s11.3.3}
\phantomsection\label{supp:iterative-controls}

The sessions without Brain Researcher were run separately by the same coding agent and are reported as qualitative process observations, not as matched randomized ablations. The prompts, interaction histories, budgets, human interventions, stopping rules, and opportunities for follow-up were not held identical.

In HCP, the separately initiated session completed a substantial analysis and ended in a valid internally held-out null: adjusted partial $r=.141$ ($p=.261$, 95\% CI $[-.182,.409]$), predictive $r=.200$, incremental $R^2=.033$, and overall $R^2=-.013$. It did not launch a frozen successor. In TRIBE, summary/PCA representations gave positive discovery and second-subset shifts ($\Delta\rho=.112$, 95\% CI $[.026,.198]$, familywise QAP $p=.0163$; and $\Delta\rho=.115$, 95\% CI $[.019,.225]$, $p=.00024$), but the frozen primary spectrotemporal representation did not reproduce them ($\Delta\rho=-.0036$, 95\% CI $[-.046,.040]$, $p=.7783$; and $\Delta\rho=-.0162$, 95\% CI $[-.063,.032]$, $p=.2484$). The prospective firewall and frozen-representation lock were not satisfied, and the authoritative terminal state was \texttt{TECHNICAL\_FAILURE}, not a confirmed positive or a scientific null. That session also did not launch a successor.

Across both with-Brain-Researcher episodes, repeated splits or recurring collections are stability checks rather than independent replications. HCP remained retrospective, same-cohort, and search-conditional, with no external confirmation. TRIBE's recurring-source pattern did not survive its first multiplicity-corrected new-source endpoint. Accordingly, \emph{self-evolving} denotes a governed trajectory in which evidence changes the next frozen analysis; it does not denote autonomous model improvement or scientific confirmation.

\subsubsection{S11.3.4. Review-layer calibration}\label{supp:s11.3.4}

To attach a number to how often the review layer's triage errs, we scored its severity classification, blind, against the 60-case calibration library (C01--C60; \hyperref[app:g]{Appendix~G}, G9.5), whose cases carry an expected verdict for common neuroimaging analysis situations (16 \emph{block}, 39 \emph{warn}, 5 \emph{allow}). The layer's classification arm produced no false accepts (0 of 16 invalid analyses were waved through as valid; 0\%, rule-of-three 95\% upper bound 19\%; and 0 of the 55 cases warranting any flag were classed \emph{allow}) and no false blocks (0 of 5 valid controls were blocked, although five controls provide only a loose upper bound). Exact three-way agreement with the expected verdict was 88\% (53/60); the seven residual disagreements were all \emph{block}-versus-\emph{warn} severity differences. The library was assembled after the directionality check introduced in response to NM-H2 and shares provenance with the review policy. These values are therefore an internal-consistency ceiling over short canonical scenarios, not an independent field error rate or evidence that the pre-patch review layer would have caught NM-H2.

\subsection{S11.4. Resource usage and user experience}\label{supp:s11.4}

We report token usage, derived cost, and interaction profile for one collaborator episode (a cocaine-use-disorder resting-state case) as an illustrative reference. Transcripts were not captured for the other collaborator cases, so per-case token accounting is shown for this case only. This episode was run on Claude Sonnet 4.6 and comprised 2,350 assistant turns and 1,347 tool calls (902 Bash, 89 Edit, 64 Read, and 53 Write, plus Model-Context-Protocol calls that included 45 cluster (Slurm) submissions and 41 deep-research queries), spread over roughly six calendar days of interactive use rather than continuous compute.

Tokens are the primary reported quantity, and we present the components separately (Table~\ref{tab:resource-usage}). The episode consumed about 0.89M output tokens, 19.9M cache-creation (cache-write) input tokens, 180.2M cache-read input tokens, and under 0.03M non-cached input tokens. We do not headline the roughly 201M raw token sum: it is dominated by cache reads, an artifact of long-context caching across 2,350 turns, where the persisted episode context is re-read on each turn and billed at roughly one-tenth the standard input rate. The meaningful generation is therefore about 0.89M output tokens plus about 20M cache-creation tokens.

From these token counts we derive an approximate cost at standard Claude Sonnet 4.6 API list prices (input \$3, output \$15, cache write \$3.75, and cache read \$0.30 per 1M tokens): about \$13 for output, about \$75 for cache-creation, about \$54 for cache-read, and under \$1 for uncached input, for a total of approximately \$140 for this single episode. Cost is driven mainly by cache-creation and output rather than by the large cache-read volume. This is an approximate API-list-equivalent figure: actual billing depends on the user's plan and tier, and subscription plans differ from per-token API pricing.

\begin{center}
\small
\setlength{\tabcolsep}{5pt}
\renewcommand{\arraystretch}{1.15}
\begin{tabular}{|>{\RaggedRight\hspace{0pt}\arraybackslash}p{0.40\textwidth}|>{\RaggedRight\hspace{0pt}\arraybackslash}r|>{\RaggedRight\hspace{0pt}\arraybackslash}r|}
\hline
\rowcolor{brtablehead}\textbf{Token component} & \textbf{Tokens} & \textbf{Approx. cost} \\
\hline
Output (generation) & 0.89M & \$13 \\
\hline
Cache-creation (cache write) & 19.9M & \$75 \\
\hline
Cache-read & 180.2M & \$54 \\
\hline
Uncached input & 0.03M & $<$\$1 \\
\hline
\rowcolor{brtablehead}\textbf{Total (single episode)} & \textbf{$\sim$201M (raw sum)} & \textbf{$\sim$\$140} \\
\hline
\end{tabular}
\captionof{table}{Token usage and approximate API-list-equivalent cost for one collaborator episode (cocaine-use-disorder resting-state case, Claude Sonnet 4.6). Components are reported separately; the raw sum is dominated by cache reads (billed at roughly one-tenth the input rate) and is not the headline quantity. Costs are derived from the per-component token counts at standard list prices (input \$3, output \$15, cache write \$3.75, cache read \$0.30 per 1M tokens) and are approximate; actual billing depends on the user's plan and tier.}
\label{tab:resource-usage}
\end{center}

Usage scales with how much the user reviews generated code rather than only reading it, and it varies by model and tier. In practice the binding constraint is high token throughput rather than marginal cost: throughput alone can exhaust a standard plan, and one collaborator case accordingly required a higher-tier plan.

In this episode, the user supplied fMRIPrep-preprocessed SUDMEX resting-state data and asked, exploratorily, how to analyse individual functional networks given a limited cross-sectional sample. Brain Researcher resolved this request into a typed episode (dataset and target functional networks) and committed a 36-specification multiverse at the commitment gate; the agent executed it on a cluster backend through Brain Researcher's submission tools, and Brain Researcher reviewed the resulting run bundle and returned a bounded null claim. In a post-hoc self-assessment, the collaborator on this case estimated that Brain Researcher saved on the order of three months of analysis time relative to assembling and running the equivalent multiverse by hand. The division of responsibility between the researcher and the system across this flow follows the human-agent authority boundary (Fig.~\ref{fig:human-agent-boundary}).

\subsection{S11.5. Analysis-level methods provenance (COBIDAS-style)}\label{supp:s11.5}

Brain Researcher did not acquire MRI data or run raw-image preprocessing for any reported case. It operates at the analysis and statistical-inference level, on already-preprocessed derivatives (functional-connectivity matrices, fMRIPrep outputs, minimally-preprocessed connectomes) or on published coordinates and reference maps. Acquisition and image-preprocessing provenance is therefore inherited from the cited source datasets and is not re-derived here; the acquisition/preprocessing column of Table~\ref{tab:cobidas-provenance} records the \emph{origin} of each derivative rather than a pipeline Brain Researcher executed. What the system records for every case is the available analysis-level provenance: the estimator, statistical model, inference and multiple-comparison procedure, robustness or multiverse design, applicable prospective gates or post-hoc review criteria, and the resulting claim state. These fields are written to the case's run or audit bundle (\hyperref[supp:s7.4]{S7.4}), including available software identity and versions, random seeds, and the specification ledger. Table~\ref{tab:cobidas-provenance} summarizes this analysis-relevant subset of the COBIDAS reporting standard \citep{nichols_best_2017} across the five reported episodes; per-case reports (Appendix~L) and the specification ledgers (Appendix~F) carry the complete records.

{
\footnotesize
\setlength{\tabcolsep}{3pt}
\renewcommand{\arraystretch}{1.2}
\begin{longtable}{|
  >{\RaggedRight\arraybackslash}p{0.12\linewidth}|
  >{\RaggedRight\arraybackslash}p{0.15\linewidth}|
  >{\RaggedRight\arraybackslash}p{0.19\linewidth}|
  >{\RaggedRight\arraybackslash}p{0.22\linewidth}|
  >{\RaggedRight\arraybackslash}p{0.13\linewidth}|
  >{\RaggedRight\arraybackslash}p{0.12\linewidth}|}
\caption{\textbf{Analysis-level methods provenance across the five reported episodes (COBIDAS-style).} Brain Researcher performs no raw acquisition or image preprocessing; the acquisition/preprocessing column records the origin of the derivatives or coordinates each episode consumed. All remaining columns describe analysis-level choices that Brain Researcher recorded and reviewed, with prospective commitment gates applied where applicable (\hyperref[supp:s7.4]{S7.4}); the compute itself was executed by the agent, except in the engine-executed Neurosynth reference. Full per-case records are in Appendices~F and~L.}
\label{tab:cobidas-provenance} \\
\hline
\rowcolor{brtablehead}\textbf{Episode (data, $N$)} & \textbf{Acquisition / preprocessing origin} & \textbf{Features \& estimator} & \textbf{Statistical model, inference \& correction} & \textbf{Robustness / review criteria} & \textbf{Claim state(s)} \\
\hline
\endfirsthead
\hline
\rowcolor{brtablehead}\textbf{Episode (data, $N$)} & \textbf{Acquisition / preprocessing origin} & \textbf{Features \& estimator} & \textbf{Statistical model, inference \& correction} & \textbf{Robustness / review criteria} & \textbf{Claim state(s)} \\
\hline
\endhead
\hline
\endfoot
\hline
\endlastfoot
NeuroMark / schizophrenia FNC (FBIRN; $N=363$, 181 control / 182 patient) & NeuroMark~2.2 template-ICA FNC derivatives (framework/collaborator-provided) \citep{du_neuromark_2020} & 5{,}460 FNC edges, two domain granularities; four connectivity estimators (Pearson, Spearman, partial correlation, mutual information); five dimensionality-reduction methods; four classifiers & Patient-vs-control classification (AUC) and group-difference contrasts; one-sided permutation test with sign-aware acceptance ($p<0.05$ and $\Delta\,\mathrm{mean}|d|>0$); Wilcoxon test for loading mass & 480-specification multiverse ($4\times3\times5\times4\times2$); 384 latent-evaluable; 24 unique sign-aware H2 contrasts & NM-H1 qualified; NM-H2 qualified; NM-H3 qualified \\
\hline
Cocaine-use-disorder connectivity (SUDMEX CONN, OpenNeuro ds003346 v1.1.3; $N=138$) & fMRIPrep derivatives, collaborator-supplied \citep{angeles-valdez_mexican_2022} & Network segregation; atlas varied across specifications; same-data meta-analysis (SDMA-GLS) \citep{lefort-besnard_statistical_2025} & Connectivity--behavior associations; FDR correction (confirmatory all $Z<1.24$, $q>0.58$; exploratory max $Z_{\mathrm{GLS}}=2.18$, 0/70) & 36-specification multiverse (atlas $\times$ framewise-displacement threshold $\times$ confound); 70-combination exploratory screen & 5 confirmatory rejected; exploratory screen blocked \\
\hline
Cross-cultural social cognition (coordinate literature; 21 studies, 85 peaks) & Published MNI peak coordinates (no raw imaging data) & Cell-wise activation-likelihood estimation (ALE), NiMARE \citep{salo_nimare_2023} & ALE with per-cell $k=6$--$8$ (below recommended $k\ge17$ \citep{eickhoff_behavior_2016}); centroid shift reported as descriptive only & Paradigm-composition imbalance check across four culture-by-relationship cells & mPFC-topology interpretation blocked as exploratory \\
\hline
HCP-YA behavioural prediction (search cohort $N=326$; matched comparison $N=244$, 243 families; internal holdout $N=81$) & HCP S1200 minimally-preprocessed connectomes and locally reconstructed Liu-component counterparts \citep{van_essen_wu-minn_2013,liu_benchmarking_2025} & 116 allocated candidate evaluations; frozen selected workflow using a coherence-magnitude representation with cosine-kernel ridge regression; locally matched Liu-style nested comparator & Ten repeated family-grouped $5\times3$ nested-CV splits with pooled out-of-fold metrics; family-cluster pointwise bootstrap and conditional one-sided test; weak-FWER transfer analysis & 104/116 parent-run scores plus 12 recorded transport failures and separate recovery; frozen selected workflow refits across outcomes; retrospective same-cohort, search-unadjusted sensitivity & The frozen selected workflow exceeded the matched comparator in 10/10 Cognition splits; 47/50 directional wins across five outcomes, but no transfer cell passed weak-FWER; separate holdout did not confirm; no external acceptance \\
\hline
TRIBE speech--tools representation geometry (TRIBE~v2; discovery and successor panels of natural sounds) & No subject-level BOLD; model-representation analysis over six sound categories, four recurring collections, and four previously unused collections \citep{dascoli_foundation_2026} & Early-versus-late encoder decoding and frozen geometric decomposition of normalized separation, orientation, and signed projection & Descriptive 15-pair discovery; three non-overlapping recurring-source panels; 99,999 balanced-label permutations and Holm correction at the frozen four-new-source endpoint & Researcher-authorized post-hoc choice of speech--tools; frozen estimand; item non-overlap; score-blind acoustic balancing; collection-specific and leave-one-out diagnostics & Recurring-source bounded support in 3/3 panels (11/12 cells), followed by unsupported new-source H1 (Holm $p=.396$); terminal outcome inconclusive or conflicting; no scientific acceptance \\
\hline
\end{longtable}
}

\section{S12. Limitations and reporting boundaries}\label{supp:s12}

Brain Researcher verifies whether an analysis package satisfies declared validity constraints, produces required artifacts, and calibrates claims to available evidence. It does not prove biological truth, replace expert scientific judgment, or replace independent replication.

\begin{center}
\small
\setlength{\tabcolsep}{5pt}
\renewcommand{\arraystretch}{1.15}
\begin{tabular}{|>{\RaggedRight\hspace{0pt}\arraybackslash}p{0.22\textwidth}|>{\RaggedRight\hspace{0pt}\arraybackslash}p{0.44\textwidth}|>{\RaggedRight\hspace{0pt}\arraybackslash}p{0.25\textwidth}|}
\hline
\rowcolor{brtablehead}\textbf{Limitation domain} & \textbf{What can go wrong} & \textbf{Reporting boundary} \\
\hline
Knowledge graph & Incomplete negative knowledge, uneven graph density, source-specific gaps, sparse higher-tier evidence. & Evidence tier and coverage caveats are reported. \\
\hline
Evidence connectors & Slow or failed connectors reduce coverage. & Failures are logged and may trigger caveats; they are not treated as absence of evidence. \\
\hline
Planning and constraints & Incomplete automated power analysis, limited interactive replanning during execution, missing persistent rejection rationales in some workflows. & Constraint state and planning limitations are recorded. \\
\hline
Execution & Dataset-access constraints, missing derivatives, backend failures, and weaker provenance for real-time paths. & Non-executable outcomes and backend failures are retained. \\
\hline
Review & Rule registry is heuristic, not formal verification. It may miss subtle domain errors or over-warn on unusual valid methods. & Review verdicts reduce obvious errors but do not prove truth. \\
\hline
Benchmark construction & Tool-calling and evidence-citation items and their reference answers are internally curated and scored by LLM judges, so they may embed designer assumptions and LLM-judge biases, including self-preference where the agent and a judge share a model family. & Scoring contracts, reference answers, and provenance files are released for independent re-scoring; benchmark results are reported as upstream-input gains rather than scientific conclusions and use three condition-blind judges combined by majority vote. \\
\hline
Routing-ablation scope & This seven-model ablation evaluates exact-label top-1 route selection with direct KG-named calls unavailable. & Report 362/420 separately from Correct route/tool@1, which has a different scoring contract. \\
\hline
Memory & Similarity-based retrieval and critic relation classification can err; partition enforcement is required. & Memory is append-only and partitioned during evaluation. \\
\hline
Bounded autonomy & Harbor reward can reflect file/\allowbreak{}schema/\allowbreak{}statistic completion without scientific validity. & Scientific acceptance still requires Brain Researcher review and claim calibration. \\
\hline
Hypothesis triage & Sparse-coverage domains can produce plausible but generic semantic permutations. & Used for expert-in-the-loop ideation, not autonomous novelty or truth claims. \\
\hline
Governance & IRB, DUA, licensing, and institutional compliance remain investigator responsibilities. & BR can record blockers but does not replace human compliance review. \\
\hline
\end{tabular}
\end{center}

Four further boundaries are not fully captured by the table above. First, evidence-support judgments rely on LLM judges; we report inter-judge agreement (three-judge Fleiss' $\kappa = 0.50$ with-BR, $0.72$ without-BR; S11.1.2) and a reproducible 272-item hand audit (S11.1.3) but did not perform full-scale human-expert re-labeling of all judge outputs, so a residual gap between LLM and human judgment cannot be ruled out. Second, the absolute level of verified groundedness remains low (0.22 with BR under a three-judge majority vote): the benchmark demonstrates a between-condition improvement, not a solved grounding problem. Third, we did not run a randomized user study and did not directly measure runtime or researcher effort; statements about reduced workflow friction are expected consequences of absorbing the execution layer rather than measured outcomes. Fourth, the foundation models and provider APIs used are evolving, and although every reported claim is fixed to recorded model versions and snapshots (S1.2), exact behaviour may not be reproducible once providers retire or update those versions.

Review labels create a separate risk of automation complacency. A ``BR-reviewed'' status means only that the formalized conditions active for that review were checked; it is not expert endorsement or evidence that errors outside those conditions are absent. NM-H2 illustrates this boundary: automated review missed a sign-blind acceptance rule that direct inspection revealed. We did not measure whether the approving outcome reduced scrutiny, but treating such labels as conclusive could discourage the inspection needed to identify failures outside implemented checks. Human inspection therefore remains necessary, especially for claim--method alignment and other judgments not yet formalized.

Capabilities not evaluated in the reported protocol should not be used as evidence for the main claims. These include adaptive optimization modules, community contribution pathways, planned cross-modal fMRI-to-concept alignment, and autonomous scientific-discovery claims beyond the bounded validation contracts explicitly reported.

\section{Appendix / Data Card Overview}\label{supp:appendix-overview}

\subsection{Appendix A. Episode and control-plane card}\label{supp:appendix-a-overview}\label{app:a}
\noindent\emph{The full episode/control-plane ledger is released in the archival repository (run-bundle schemas and provenance fields; Data availability); the card below is the summary of its fields.}

The episode/\allowbreak{}control-plane card records the fixed identity and configuration of an episode: episode ID, mode, scientific question, model version, prompt-template version, registry snapshot, BR-KG snapshot, policy flags, MCP operation summary, memory namespace, checkpoint events, recovery events, run state, and completion status.

Example excerpt:
\begin{suppcode}
episode_id: ep_hcp_predict_001
mode: benchmark
state: completed_with_qualifications
registry_snapshot: registry_2026_05_15
brkg_snapshot: brkg_2026_05_10
memory_namespace: benchmark_isolated
\end{suppcode}

\subsection{Appendix B. Evidence bundle / BR-KG card}\label{supp:appendix-b-overview}\label{app:b}
\noindent\emph{The full BR-KG atlas and evidence-bundle ledger (graph snapshots, node/edge schemas, source-coverage and provenance tables, and the atlas figure panels) is released in the archival repository (Data availability). The card below retains the evidence-bundle schema and one worked method-condition record, the object the main text points to for a source's verbatim quote and grounding label.}

The evidence bundle card records input query, resolved entities, ONVOC mappings, evidence tiers, graph paths, literature retrieval outputs, connector failures, dataset links, tool links, prior findings, method-condition records, coverage notes, and embedding-lane metadata.

Example excerpt:
\begin{suppcode}
resolved_entities:
  dataset: HCP-YA
  feature_space: Schaefer100x7
evidence_tiers:
  - accepted_graph_record
  - real_time_retrieval
connector_failures:
  - PubMed_timeout
graph_path:
  - Dataset:HCP-YA
  - HAS_FEATURE_SPACE:Schaefer100x7
  - REQUIRES:fold_manifest
method_condition_record:            # one source-supported claim, field by field
  claim: "resting-state FC predicts a fluid-intelligence component"
  cohort_sample_size: "N=1003"
  task_paradigm: resting_state
  preprocessing: "ICA-FIX + 24-parameter motion regression"
  statistical_model: "ridge regression, family-aware cross-validation"
  grounding_label: br_kg_gabriel_cache_candidate
  verbatim_quote: "connectome-based models predicted a general
    intelligence factor (r approx 0.29) with leave-one-family-out CV"
\end{suppcode}

\subsection{Appendix C. Dataset/resource card}\label{supp:appendix-c-overview}\label{app:c}
\noindent\emph{The full per-dataset readiness ledger is released in the archival repository (dataset records and provenance fields; Data availability); the card below is the summary of its fields.}

The dataset/\allowbreak{}resource card records dataset identifier, aliases, access class, local or remote path, BIDS root, derivative root, phenotype manifest, target variables, covariates, missing derivatives, backend reachability, readiness status, and blocker status. It includes at least one pass example and one block example.

\subsection{Appendix D. Tool registry and specification ledger card}\label{supp:appendix-d-overview}\label{app:d}
\noindent\emph{The full tool-registry, candidate-ranking, and specification ledger is released in the archival repository (registry links and scoring tables; Data availability); the card below is the summary with a representative candidate-accounting excerpt.}

The tool registry card records family cards, ranked tool candidates, rejected candidates, canonical IDs, backend recipes, parameter schemas, tool-contract clauses, compatibility checks, Neurodesk mappings, registry snapshot identifiers, allowlist mode, and embedding-lane metadata.

Example candidate accounting:
\begin{center}
\small
\setlength{\tabcolsep}{5pt}
\renewcommand{\arraystretch}{1.15}
\begin{tabular}{|>{\RaggedRight\hspace{0pt}\arraybackslash}p{0.34\textwidth}|>{\RaggedRight\hspace{0pt}\arraybackslash}p{0.18\textwidth}|>{\RaggedRight\hspace{0pt}\arraybackslash}p{0.39\textwidth}|}
\hline
\rowcolor{brtablehead}\textbf{Candidate} & \textbf{Decision} & \textbf{Reason} \\
\hline
predictive\_modeling\_family & accepted & Input resources and expected outputs match. \\
\hline
whole\_brain\_glm & rejected & Wrong output type for prediction target. \\
\hline
dynamic\_fc\_family & rejected & Missing time-series asset. \\
\hline
\end{tabular}
\end{center}

\subsection{Appendix E. Constraint and commitment card}\label{supp:appendix-e-overview}\label{app:e}
\noindent\emph{The full constraint-compiler and commitment-gate record is released in the archival repository (Data availability); the card below is the summary of its fields.}

The constraint/\allowbreak{}commitment card records active constraints, hard checks, soft checks, rule provenance, pass/\allowbreak{}warn/\allowbreak{}block verdicts, required sensitivity analyses, gate authority, commitment decision, and benchmark pass-through flag.

\subsection{Appendix F. Run bundle and provenance card}\label{supp:appendix-f-overview}\label{app:f}
\noindent\emph{The full run-bundle and provenance ledger, including a worked claim-record example, is released in the archival repository (run-bundle schemas and provenance fields; Data availability); the card below is the summary of its fields.}

The run-bundle card records event trace, trajectory document, observation record, analysis bundle, run card, expected artifacts, produced artifacts, missing artifacts, checksums, backend versions, software versions, container or module identifiers, preflight results, failures, retries, recovery events, and final execution status.

\subsection{Appendix G. Review card}\label{supp:appendix-g-overview}

The review card records verification inputs, deterministic checks, BLOCK findings, WARN findings, robustness checks, sensitivity checks, claim families, verdicts, caveat language, revision routing, claim eligibility, and artifact-completeness ratio. Each revise or block verdict identifies the triggering artifact or assumption, rule ID, hard/\allowbreak{}soft status, and evidence needed for resolution.

\subsection{Appendix H. Memory card}\label{supp:appendix-h-overview}\label{app:h}
\noindent\emph{The full memory, claim-relation, and BR-KG-promotion ledger is released in the archival repository (Data availability); the card below is the summary of its fields.}

The memory card records claim text, polarity, condition vector, review verdict, caveats, provenance pointer, related prior claims, relation type, relation confidence, stable key, memory namespace, writeback eligibility, and BR-KG promotion status. Accepted, qualified, and rejected outcomes are shown separately so readers can see what does and does not enter accepted memory.

\subsection{Appendix I. Operational-mode card}\label{supp:appendix-i-overview}

The operational-mode card records interactive gate records, bounded instruction file, action vocabulary, budget, stopping criteria, validation ladder, supervisor decisions, critic decisions, Harbor verifier outputs, reward or partial-credit metrics, escalation events, final termination status, and memory-writeback decision.

\subsection{Appendix J. Evaluation card}\label{supp:appendix-j-overview}

The evaluation card records benchmark suite list, task manifests, scoring contracts, model list, with-BR/\allowbreak{}without-BR protocol, memory partition policy, collaborator case table, bounded campaign contract, metric denominators, secondary metrics, ablation protocols, and excluded or non-executable cases. The card states whether scoring used binary success, partial credit, or both.

\subsection{Case-report skeleton for later appendix expansion}\label{supp:case-report-skeleton}

Each case report uses the same skeleton:

\begin{center}
\small
\setlength{\tabcolsep}{5pt}
\renewcommand{\arraystretch}{1.15}
\begin{tabular}{|>{\RaggedRight\hspace{0pt}\arraybackslash}p{0.12\textwidth}|>{\RaggedRight\hspace{0pt}\arraybackslash}p{0.79\textwidth}|}
\hline
\rowcolor{brtablehead}\textbf{Step} & \textbf{Case-report element} \\
\hline
1 & Question. \\
\hline
2 & Dataset/\allowbreak{}resource status. \\
\hline
3 & Tool pathway. \\
\hline
4 & Constraints and commitment gate. \\
\hline
5 & Execution artifacts. \\
\hline
6 & Review verdict. \\
\hline
7 & Final claim language. \\
\hline
8 & Human/\allowbreak{}system responsibility split. \\
\hline
9 & Memory writeback status. \\
\hline
\end{tabular}
\end{center}

This skeleton documents the reporting contract for the case-report appendices. Apart from the representative generated-code excerpt in \hyperref[app:k]{Appendix K}, the automatically generated reports released through \hyperref[supp:case-reports]{Appendix~L} are frozen audit artifacts. The earlier HCP retention-gate and TRIBE language-alignment reports are historical records of separate campaigns and do not supply the current iterative-episode results. The operative current results are those in \hyperref[supp:s11.2]{Supplementary Methods S11.2}--\hyperref[supp:s11.3]{S11.3} and \hyperref[app:j]{Appendix J}.

\clearpage
\section*{Appendix/Data Cards}
\phantomsection\label{app:data-cards}
\addcontentsline{toc}{section}{Appendix/Data Cards}

\section{\texorpdfstring{\textbf{Appendix G. Scientific review, BLOCK/WARN findings, and claim-calibration card}}{Appendix G. Scientific review, BLOCK/WARN findings, and claim-calibration card}}\label{appendix-g.-scientific-review-blockwarn-findings-and-claim-calibration-card}\label{app:g}

\hyperref[app:g]{Appendix G} is the scientific review ledger: it consumes the run bundle and decides whether artifacts and candidate claims are accepted, qualified, revised, blocked, rejected, or deferred, with unresolved cases escalated for expert judgment.

\subsection{\texorpdfstring{\textbf{G1. Purpose and scope}}{G1. Purpose and scope}}\label{g1.-purpose-and-scope}

This card records the scientific review outcome for an episode. It consumes the selected plan, active constraints, execution trace, artifact manifest, logs, QC outputs, scorecards, candidate claims, and prior evidence. It reports deterministic checks, BLOCK and WARN findings, robustness and sensitivity coverage, claim-calibration decisions, final verdicts, caveat language, revision routing, and claim eligibility.

\subsection{\texorpdfstring{\textbf{G2. Review input manifest}}{G2. Review input manifest}}\label{g2.-review-input-manifest}

{
\footnotesize
\setlength{\tabcolsep}{3pt}
\renewcommand{\arraystretch}{1.15}
\begin{longtable}{|
  >{\RaggedRight\hspace{0pt}\arraybackslash}p{(\linewidth - 7\tabcolsep) * \real{0.3340}}|
  >{\RaggedRight\hspace{0pt}\arraybackslash}p{(\linewidth - 7\tabcolsep) * \real{0.3116}}|
  >{\RaggedRight\hspace{0pt}\arraybackslash}p{(\linewidth - 7\tabcolsep) * \real{0.3067}}|}
\hline
\rowcolor{brtablehead}
\begin{minipage}[t]{\linewidth}\RaggedRight\hspace{0pt}
\textbf{Input}
\end{minipage} & \begin{minipage}[t]{\linewidth}\RaggedRight\hspace{0pt}
\textbf{Source appendix}
\end{minipage} & \begin{minipage}[t]{\linewidth}\RaggedRight\hspace{0pt}
\textbf{Status}
\end{minipage} \\
\hline
\endhead
\hline
\endlastfoot

\begin{minipage}[t]{\linewidth}\RaggedRight\hspace{0pt}
Selected plan
\end{minipage} & \begin{minipage}[t]{\linewidth}\RaggedRight\hspace{0pt}
\hyperref[app:d]{Appendix D}/\allowbreak{}\hyperref[app:e]{E}
\end{minipage} & \begin{minipage}[t]{\linewidth}\RaggedRight\hspace{0pt}
present
\end{minipage} \\
\hline
\begin{minipage}[t]{\linewidth}\RaggedRight\hspace{0pt}
Active constraints
\end{minipage} & \begin{minipage}[t]{\linewidth}\RaggedRight\hspace{0pt}
\hyperref[app:e]{Appendix E}
\end{minipage} & \begin{minipage}[t]{\linewidth}\RaggedRight\hspace{0pt}
present
\end{minipage} \\
\hline
\begin{minipage}[t]{\linewidth}\RaggedRight\hspace{0pt}
Run bundle
\end{minipage} & \begin{minipage}[t]{\linewidth}\RaggedRight\hspace{0pt}
\hyperref[app:f]{Appendix F}
\end{minipage} & \begin{minipage}[t]{\linewidth}\RaggedRight\hspace{0pt}
present
\end{minipage} \\
\hline
\begin{minipage}[t]{\linewidth}\RaggedRight\hspace{0pt}
Artifact manifest
\end{minipage} & \begin{minipage}[t]{\linewidth}\RaggedRight\hspace{0pt}
\hyperref[app:f]{Appendix F}
\end{minipage} & \begin{minipage}[t]{\linewidth}\RaggedRight\hspace{0pt}
present
\end{minipage} \\
\hline
\begin{minipage}[t]{\linewidth}\RaggedRight\hspace{0pt}
Logs/\allowbreak{}errors/\allowbreak{}warnings
\end{minipage} & \begin{minipage}[t]{\linewidth}\RaggedRight\hspace{0pt}
\hyperref[app:f]{Appendix F}
\end{minipage} & \begin{minipage}[t]{\linewidth}\RaggedRight\hspace{0pt}
present
\end{minipage} \\
\hline
\begin{minipage}[t]{\linewidth}\RaggedRight\hspace{0pt}
Scorecard and QC outputs
\end{minipage} & \begin{minipage}[t]{\linewidth}\RaggedRight\hspace{0pt}
\hyperref[app:f]{Appendix F}
\end{minipage} & \begin{minipage}[t]{\linewidth}\RaggedRight\hspace{0pt}
present
\end{minipage} \\
\hline
\begin{minipage}[t]{\linewidth}\RaggedRight\hspace{0pt}
Prior evidence /\allowbreak{} graph paths
\end{minipage} & \begin{minipage}[t]{\linewidth}\RaggedRight\hspace{0pt}
\hyperref[app:b]{Appendix B}
\end{minipage} & \begin{minipage}[t]{\linewidth}\RaggedRight\hspace{0pt}
present
\end{minipage} \\
\hline
\begin{minipage}[t]{\linewidth}\RaggedRight\hspace{0pt}
Dataset/\allowbreak{}resource ledger
\end{minipage} & \begin{minipage}[t]{\linewidth}\RaggedRight\hspace{0pt}
\hyperref[app:c]{Appendix C}
\end{minipage} & \begin{minipage}[t]{\linewidth}\RaggedRight\hspace{0pt}
present
\end{minipage} \\
\end{longtable}
}

Important boundary: review consumes the run bundle, not the model transcript alone.

\subsection{\texorpdfstring{\textbf{G3. Validity-layer matrix}}{G3. Validity-layer matrix}}\label{g3.-validity-layer-matrix}

{
\footnotesize
\setlength{\tabcolsep}{3pt}
\renewcommand{\arraystretch}{1.15}
\begin{longtable}{|
  >{\RaggedRight\hspace{0pt}\arraybackslash}p{(\linewidth - 7\tabcolsep) * \real{0.3116}}|
  >{\RaggedRight\hspace{0pt}\arraybackslash}p{(\linewidth - 7\tabcolsep) * \real{0.3051}}|
  >{\RaggedRight\hspace{0pt}\arraybackslash}p{(\linewidth - 7\tabcolsep) * \real{0.3373}}|}
\hline
\rowcolor{brtablehead}
\begin{minipage}[t]{\linewidth}\RaggedRight\hspace{0pt}
\textbf{Validity layer}
\end{minipage} & \begin{minipage}[t]{\linewidth}\RaggedRight\hspace{0pt}
\textbf{Review question}
\end{minipage} & \begin{minipage}[t]{\linewidth}\RaggedRight\hspace{0pt}
\textbf{Example checks}
\end{minipage} \\
\hline
\endhead
\hline
\endlastfoot

\begin{minipage}[t]{\linewidth}\RaggedRight\hspace{0pt}
Statistical validity
\end{minipage} & \begin{minipage}[t]{\linewidth}\RaggedRight\hspace{0pt}
Is the inference valid?
\end{minipage} & \begin{minipage}[t]{\linewidth}\RaggedRight\hspace{0pt}
exchangeability; multiple comparisons; permutation/\allowbreak{}null
\end{minipage} \\
\hline
\begin{minipage}[t]{\linewidth}\RaggedRight\hspace{0pt}
Measurement validity
\end{minipage} & \begin{minipage}[t]{\linewidth}\RaggedRight\hspace{0pt}
Are data and QC adequate?
\end{minipage} & \begin{minipage}[t]{\linewidth}\RaggedRight\hspace{0pt}
motion; registration; QC; reliability
\end{minipage} \\
\hline
\begin{minipage}[t]{\linewidth}\RaggedRight\hspace{0pt}
Construct validity
\end{minipage} & \begin{minipage}[t]{\linewidth}\RaggedRight\hspace{0pt}
Does the measure support the construct?
\end{minipage} & \begin{minipage}[t]{\linewidth}\RaggedRight\hspace{0pt}
task/\allowbreak{}construct mapping; reverse inference risk
\end{minipage} \\
\hline
\begin{minipage}[t]{\linewidth}\RaggedRight\hspace{0pt}
Generalization validity
\end{minipage} & \begin{minipage}[t]{\linewidth}\RaggedRight\hspace{0pt}
Does the result generalize?
\end{minipage} & \begin{minipage}[t]{\linewidth}\RaggedRight\hspace{0pt}
leakage; grouped CV; held-out/\allowbreak{}external support
\end{minipage} \\
\hline
\begin{minipage}[t]{\linewidth}\RaggedRight\hspace{0pt}
Claim validity
\end{minipage} & \begin{minipage}[t]{\linewidth}\RaggedRight\hspace{0pt}
Is language calibrated?
\end{minipage} & \begin{minipage}[t]{\linewidth}\RaggedRight\hspace{0pt}
no causal/\allowbreak{}biomarker/\allowbreak{}external claim without evidence
\end{minipage} \\
\end{longtable}
}

\subsection{\texorpdfstring{\textbf{G4. Review rule registry excerpt and lifecycle status}}{G4. Review rule registry excerpt and lifecycle status}}\label{g4.-review-rule-registry-excerpt}

The review registry separates rule severity from implementation lifecycle. Severity determines what a triggered rule does to a candidate claim. Lifecycle status records whether the rule is already operational, can be implemented from structured metadata, requires additional review-context fields, requires text interpretation, or is used only for calibration and reviewer training.

{
\footnotesize
\setlength{\tabcolsep}{3pt}
\renewcommand{\arraystretch}{1.15}
\begin{longtable}{|
  >{\RaggedRight\hspace{0pt}\arraybackslash}p{(\linewidth - 7\tabcolsep) * \real{0.2570}}|
  >{\RaggedRight\hspace{0pt}\arraybackslash}p{(\linewidth - 7\tabcolsep) * \real{0.4150}}|
  >{\RaggedRight\hspace{0pt}\arraybackslash}p{(\linewidth - 7\tabcolsep) * \real{0.2800}}|}
\hline
\rowcolor{brtablehead}
\begin{minipage}[t]{\linewidth}\RaggedRight\hspace{0pt}
\textbf{Lifecycle status}
\end{minipage} & \begin{minipage}[t]{\linewidth}\RaggedRight\hspace{0pt}
\textbf{Meaning}
\end{minipage} & \begin{minipage}[t]{\linewidth}\RaggedRight\hspace{0pt}
\textbf{How it is reported}
\end{minipage} \\
\hline
\endhead
\hline
\endlastfoot

\begin{minipage}[t]{\linewidth}\RaggedRight\hspace{0pt}
Implemented
\end{minipage} & \begin{minipage}[t]{\linewidth}\RaggedRight\hspace{0pt}
The rule is operational in the review system for the required metadata fields.
\end{minipage} & \begin{minipage}[t]{\linewidth}\RaggedRight\hspace{0pt}
May support an enforced BLOCK or WARN verdict.
\end{minipage} \\
\hline
\begin{minipage}[t]{\linewidth}\RaggedRight\hspace{0pt}
Deterministic candidate
\end{minipage} & \begin{minipage}[t]{\linewidth}\RaggedRight\hspace{0pt}
The rule can likely be implemented from structured metadata, manifests, or pipeline logs.
\end{minipage} & \begin{minipage}[t]{\linewidth}\RaggedRight\hspace{0pt}
Reported as a candidate check unless the episode records enforcement.
\end{minipage} \\
\hline
\begin{minipage}[t]{\linewidth}\RaggedRight\hspace{0pt}
Schema-dependent candidate
\end{minipage} & \begin{minipage}[t]{\linewidth}\RaggedRight\hspace{0pt}
The rule requires additional review-context fields before deterministic enforcement.
\end{minipage} & \begin{minipage}[t]{\linewidth}\RaggedRight\hspace{0pt}
Reported as a missing-field or protocol requirement.
\end{minipage} \\
\hline
\begin{minipage}[t]{\linewidth}\RaggedRight\hspace{0pt}
NLP/LLM candidate
\end{minipage} & \begin{minipage}[t]{\linewidth}\RaggedRight\hspace{0pt}
The rule requires text interpretation, claim extraction, or semantic comparison.
\end{minipage} & \begin{minipage}[t]{\linewidth}\RaggedRight\hspace{0pt}
Reported as assisted review or calibration unless independently verified.
\end{minipage} \\
\hline
\begin{minipage}[t]{\linewidth}\RaggedRight\hspace{0pt}
Calibration-only
\end{minipage} & \begin{minipage}[t]{\linewidth}\RaggedRight\hspace{0pt}
The rule is used for examples, annotator training, or benchmark calibration.
\end{minipage} & \begin{minipage}[t]{\linewidth}\RaggedRight\hspace{0pt}
Does not by itself produce an enforced verdict.
\end{minipage} \\
\end{longtable}
}

{
\footnotesize
\setlength{\tabcolsep}{3pt}
\renewcommand{\arraystretch}{1.15}
\begin{longtable}{|
  >{\RaggedRight\hspace{0pt}\arraybackslash}p{(\linewidth - 9\tabcolsep) * \real{0.2942}}|
  >{\RaggedRight\hspace{0pt}\arraybackslash}p{(\linewidth - 9\tabcolsep) * \real{0.2122}}|
  >{\RaggedRight\hspace{0pt}\arraybackslash}p{(\linewidth - 9\tabcolsep) * \real{0.2283}}|
  >{\RaggedRight\hspace{0pt}\arraybackslash}p{(\linewidth - 9\tabcolsep) * \real{0.2186}}|}
\hline
\rowcolor{brtablehead}
\begin{minipage}[t]{\linewidth}\RaggedRight\hspace{0pt}
\textbf{Rule ID}
\end{minipage} & \begin{minipage}[t]{\linewidth}\RaggedRight\hspace{0pt}
\textbf{Severity}
\end{minipage} & \begin{minipage}[t]{\linewidth}\RaggedRight\hspace{0pt}
\textbf{Validity layer /\allowbreak{} reason tag}
\end{minipage} & \begin{minipage}[t]{\linewidth}\RaggedRight\hspace{0pt}
\textbf{Default action}
\end{minipage} \\
\hline
\endhead
\hline
\endlastfoot

\begin{minipage}[t]{\linewidth}\RaggedRight\hspace{0pt}
feature\_\allowbreak{}selection\_\allowbreak{}outside\_\allowbreak{}cv
\end{minipage} & \begin{minipage}[t]{\linewidth}\RaggedRight\hspace{0pt}
BLOCK
\end{minipage} & \begin{minipage}[t]{\linewidth}\RaggedRight\hspace{0pt}
generalization /\allowbreak{} leakage
\end{minipage} & \begin{minipage}[t]{\linewidth}\RaggedRight\hspace{0pt}
revise or block
\end{minipage} \\
\hline
\begin{minipage}[t]{\linewidth}\RaggedRight\hspace{0pt}
uncorrected\_\allowbreak{}whole\_\allowbreak{}brain\_\allowbreak{}inference
\end{minipage} & \begin{minipage}[t]{\linewidth}\RaggedRight\hspace{0pt}
BLOCK
\end{minipage} & \begin{minipage}[t]{\linewidth}\RaggedRight\hspace{0pt}
statistical /\allowbreak{} multiple comparisons
\end{minipage} & \begin{minipage}[t]{\linewidth}\RaggedRight\hspace{0pt}
revise or block
\end{minipage} \\
\hline
\begin{minipage}[t]{\linewidth}\RaggedRight\hspace{0pt}
same\_\allowbreak{}data\_\allowbreak{}roi\_\allowbreak{}definition
\end{minipage} & \begin{minipage}[t]{\linewidth}\RaggedRight\hspace{0pt}
BLOCK
\end{minipage} & \begin{minipage}[t]{\linewidth}\RaggedRight\hspace{0pt}
circularity
\end{minipage} & \begin{minipage}[t]{\linewidth}\RaggedRight\hspace{0pt}
block
\end{minipage} \\
\hline
\begin{minipage}[t]{\linewidth}\RaggedRight\hspace{0pt}
gsr\_\allowbreak{}without\_\allowbreak{}sensitivity
\end{minipage} & \begin{minipage}[t]{\linewidth}\RaggedRight\hspace{0pt}
WARN
\end{minipage} & \begin{minipage}[t]{\linewidth}\RaggedRight\hspace{0pt}
measurement /\allowbreak{} controversial choice
\end{minipage} & \begin{minipage}[t]{\linewidth}\RaggedRight\hspace{0pt}
sensitivity or caveat
\end{minipage} \\
\hline
\begin{minipage}[t]{\linewidth}\RaggedRight\hspace{0pt}
small\_\allowbreak{}sample\_\allowbreak{}biomarker\_\allowbreak{}claim
\end{minipage} & \begin{minipage}[t]{\linewidth}\RaggedRight\hspace{0pt}
WARN
\end{minipage} & \begin{minipage}[t]{\linewidth}\RaggedRight\hspace{0pt}
claim /\allowbreak{} claim inflation
\end{minipage} & \begin{minipage}[t]{\linewidth}\RaggedRight\hspace{0pt}
downgrade claim
\end{minipage} \\
\hline
\begin{minipage}[t]{\linewidth}\RaggedRight\hspace{0pt}
missing\_\allowbreak{}random\_\allowbreak{}seed
\end{minipage} & \begin{minipage}[t]{\linewidth}\RaggedRight\hspace{0pt}
WARN
\end{minipage} & \begin{minipage}[t]{\linewidth}\RaggedRight\hspace{0pt}
reproducibility /\allowbreak{} reporting gap
\end{minipage} & \begin{minipage}[t]{\linewidth}\RaggedRight\hspace{0pt}
report or fix
\end{minipage} \\
\end{longtable}
}

\begin{quote}
review\_\allowbreak{}rule:\\
rule\_\allowbreak{}id: feature\_\allowbreak{}selection\_\allowbreak{}outside\_\allowbreak{}cv\\
severity: BLOCK\\
validity\_\allowbreak{}layer: generalization\\
reason\_\allowbreak{}tags:\\
- leakage\\
detection\_\allowbreak{}field: pipeline\_\allowbreak{}dag\\
trigger: feature\_\allowbreak{}selection\_\allowbreak{}fit\_\allowbreak{}before\_\allowbreak{}cv\_\allowbreak{}split\\
default\_\allowbreak{}action: revise\_\allowbreak{}or\_\allowbreak{}block\\
required\_\allowbreak{}fix: "Move feature selection inside the cross-validation loop."
\end{quote}

\subsection{\texorpdfstring{\textbf{G5. BLOCK findings and WARN findings}}{G5. BLOCK findings and WARN findings}}\label{g5.-block-findings-and-warn-findings}

{
\footnotesize
\setlength{\tabcolsep}{3pt}
\renewcommand{\arraystretch}{1.15}
\begin{longtable}{|
  >{\RaggedRight\hspace{0pt}\arraybackslash}p{(\linewidth - 7\tabcolsep) * \real{0.3116}}|
  >{\RaggedRight\hspace{0pt}\arraybackslash}p{(\linewidth - 7\tabcolsep) * \real{0.3180}}|
  >{\RaggedRight\hspace{0pt}\arraybackslash}p{(\linewidth - 7\tabcolsep) * \real{0.3228}}|}
\hline
\rowcolor{brtablehead}
\begin{minipage}[t]{\linewidth}\RaggedRight\hspace{0pt}
\textbf{Finding type}
\end{minipage} & \begin{minipage}[t]{\linewidth}\RaggedRight\hspace{0pt}
\textbf{Examples}
\end{minipage} & \begin{minipage}[t]{\linewidth}\RaggedRight\hspace{0pt}
\textbf{Consequence}
\end{minipage} \\
\hline
\endhead
\hline
\endlastfoot

\begin{minipage}[t]{\linewidth}\RaggedRight\hspace{0pt}
BLOCK
\end{minipage} & \begin{minipage}[t]{\linewidth}\RaggedRight\hspace{0pt}
leakage outside CV; invalid null; circular ROI; uncorrected whole-brain inference
\end{minipage} & \begin{minipage}[t]{\linewidth}\RaggedRight\hspace{0pt}
Cannot accept until resolved.
\end{minipage} \\
\hline
\begin{minipage}[t]{\linewidth}\RaggedRight\hspace{0pt}
WARN
\end{minipage} & \begin{minipage}[t]{\linewidth}\RaggedRight\hspace{0pt}
external validation missing; GSR unresolved; small sample; missing seed; partial QC
\end{minipage} & \begin{minipage}[t]{\linewidth}\RaggedRight\hspace{0pt}
May accept with caveats, sensitivity, or downgraded claim.
\end{minipage} \\
\end{longtable}
}

\begin{quote}
block\_\allowbreak{}findings: {[}{]}\\
block\_\allowbreak{}status: none\_\allowbreak{}detected\\
\strut \\
warn\_\allowbreak{}findings:\\
- rule\_\allowbreak{}id: external\_\allowbreak{}validation\_\allowbreak{}missing\\
validity\_\allowbreak{}layer: generalization\\
reason\_\allowbreak{}tag: claim\_\allowbreak{}inflation\\
status: unresolved\\
required\_\allowbreak{}caveat: "Internal validation only; no external generalization claim."
\end{quote}

\subsubsection{\texorpdfstring{\textbf{G5.1 Common review failure modes and default actions}}{G5.1 Common review failure modes and default actions}}\label{g5.1.-common-review-failure-modes-and-default-actions}

The review card is easier to interpret if the rule registry is read as a set of common failure modes. These rows are not additional rules; they translate frequent neuroimaging review problems into the evidence the reviewer expects, the default BLOCK or WARN action, and the claim-language change that follows.

{
\footnotesize
\setlength{\tabcolsep}{3pt}
\renewcommand{\arraystretch}{1.15}
\begin{longtable}{|
  >{\RaggedRight\hspace{0pt}\arraybackslash}p{(\linewidth - 9\tabcolsep) * \real{0.2350}}|
  >{\RaggedRight\hspace{0pt}\arraybackslash}p{(\linewidth - 9\tabcolsep) * \real{0.3050}}|
  >{\RaggedRight\hspace{0pt}\arraybackslash}p{(\linewidth - 9\tabcolsep) * \real{0.2050}}|
  >{\RaggedRight\hspace{0pt}\arraybackslash}p{(\linewidth - 9\tabcolsep) * \real{0.2080}}|}
\hline
\rowcolor{brtablehead}
\begin{minipage}[t]{\linewidth}\RaggedRight\hspace{0pt}
\textbf{Failure mode}
\end{minipage} & \begin{minipage}[t]{\linewidth}\RaggedRight\hspace{0pt}
\textbf{Evidence inspected}
\end{minipage} & \begin{minipage}[t]{\linewidth}\RaggedRight\hspace{0pt}
\textbf{Default action}
\end{minipage} & \begin{minipage}[t]{\linewidth}\RaggedRight\hspace{0pt}
\textbf{Claim-language effect}
\end{minipage} \\
\hline
\endhead
\hline
\endlastfoot

\begin{minipage}[t]{\linewidth}\RaggedRight\hspace{0pt}
Leakage outside cross-validation
\end{minipage} & \begin{minipage}[t]{\linewidth}\RaggedRight\hspace{0pt}
Pipeline DAG, fit scope for feature selection, scaling, PCA, harmonization, or confound regression, and fold manifest.
\end{minipage} & \begin{minipage}[t]{\linewidth}\RaggedRight\hspace{0pt}
BLOCK until the fitting step is moved inside the training fold.
\end{minipage} & \begin{minipage}[t]{\linewidth}\RaggedRight\hspace{0pt}
No generalization, prediction, or biomarker claim.
\end{minipage} \\
\hline
\begin{minipage}[t]{\linewidth}\RaggedRight\hspace{0pt}
Circular ROI or feature definition
\end{minipage} & \begin{minipage}[t]{\linewidth}\RaggedRight\hspace{0pt}
ROI provenance, localizer source, contrast identity, dataset identity, and whether the same data also test the effect.
\end{minipage} & \begin{minipage}[t]{\linewidth}\RaggedRight\hspace{0pt}
BLOCK unless an independent localizer, atlas, or non-circular provenance is supplied.
\end{minipage} & \begin{minipage}[t]{\linewidth}\RaggedRight\hspace{0pt}
No confirmatory effect claim.
\end{minipage} \\
\hline
\begin{minipage}[t]{\linewidth}\RaggedRight\hspace{0pt}
Invalid whole-brain or map-level inference
\end{minipage} & \begin{minipage}[t]{\linewidth}\RaggedRight\hspace{0pt}
Correction method, correction domain, primary threshold, permutation or spatial-null framework, and file space.
\end{minipage} & \begin{minipage}[t]{\linewidth}\RaggedRight\hspace{0pt}
BLOCK for uncorrected whole-brain tests, spatial-domain mismatch, or missing spatial null.
\end{minipage} & \begin{minipage}[t]{\linewidth}\RaggedRight\hspace{0pt}
Only descriptive or exploratory language remains.
\end{minipage} \\
\hline
\begin{minipage}[t]{\linewidth}\RaggedRight\hspace{0pt}
Unresolved controversial preprocessing choice
\end{minipage} & \begin{minipage}[t]{\linewidth}\RaggedRight\hspace{0pt}
Preprocessing record and sensitivity coverage for GSR, dynamic FC windows, graph thresholds, HRF model, or harmonization.
\end{minipage} & \begin{minipage}[t]{\linewidth}\RaggedRight\hspace{0pt}
WARN with required sensitivity analysis.
\end{minipage} & \begin{minipage}[t]{\linewidth}\RaggedRight\hspace{0pt}
Claim is qualified to the tested preprocessing choices.
\end{minipage} \\
\hline
\begin{minipage}[t]{\linewidth}\RaggedRight\hspace{0pt}
Missing external or held-out validation
\end{minipage} & \begin{minipage}[t]{\linewidth}\RaggedRight\hspace{0pt}
Validation ladder, held-out split, external dataset availability, post-selection correction, and resource-gate status.
\end{minipage} & \begin{minipage}[t]{\linewidth}\RaggedRight\hspace{0pt}
WARN or DEFER, depending on whether internal support is sufficient.
\end{minipage} & \begin{minipage}[t]{\linewidth}\RaggedRight\hspace{0pt}
No external generalization or deployment claim.
\end{minipage} \\
\hline
\begin{minipage}[t]{\linewidth}\RaggedRight\hspace{0pt}
Overstated interpretation
\end{minipage} & \begin{minipage}[t]{\linewidth}\RaggedRight\hspace{0pt}
Candidate claim text, evidence type, behavioral covariates, out-of-sample evidence, ablations, and alternatives considered.
\end{minipage} & \begin{minipage}[t]{\linewidth}\RaggedRight\hspace{0pt}
WARN with claim downgrade.
\end{minipage} & \begin{minipage}[t]{\linewidth}\RaggedRight\hspace{0pt}
Prediction, causal, mechanistic, or reverse-inference language is removed unless directly supported.
\end{minipage} \\
\hline
\end{longtable}
}

\subsection{\texorpdfstring{\textbf{G6. Robustness, sensitivity, and artifact review}}{G6. Robustness, sensitivity, and artifact review}}\label{g6.-robustness-sensitivity-and-artifact-review}

{
\footnotesize
\setlength{\tabcolsep}{3pt}
\renewcommand{\arraystretch}{1.15}
\begin{longtable}{|
  >{\RaggedRight\hspace{0pt}\arraybackslash}p{(\linewidth - 7\tabcolsep) * \real{0.3212}}|
  >{\RaggedRight\hspace{0pt}\arraybackslash}p{(\linewidth - 7\tabcolsep) * \real{0.3164}}|
  >{\RaggedRight\hspace{0pt}\arraybackslash}p{(\linewidth - 7\tabcolsep) * \real{0.3148}}|}
\hline
\rowcolor{brtablehead}
\begin{minipage}[t]{\linewidth}\RaggedRight\hspace{0pt}
\textbf{Check}
\end{minipage} & \begin{minipage}[t]{\linewidth}\RaggedRight\hspace{0pt}
\textbf{Required by}
\end{minipage} & \begin{minipage}[t]{\linewidth}\RaggedRight\hspace{0pt}
\textbf{Claim effect}
\end{minipage} \\
\hline
\endhead
\hline
\endlastfoot

\begin{minipage}[t]{\linewidth}\RaggedRight\hspace{0pt}
with/\allowbreak{}without GSR
\end{minipage} & \begin{minipage}[t]{\linewidth}\RaggedRight\hspace{0pt}
controversial-choice rule
\end{minipage} & \begin{minipage}[t]{\linewidth}\RaggedRight\hspace{0pt}
qualified if not performed or unstable
\end{minipage} \\
\hline
\begin{minipage}[t]{\linewidth}\RaggedRight\hspace{0pt}
parcellation sensitivity
\end{minipage} & \begin{minipage}[t]{\linewidth}\RaggedRight\hspace{0pt}
validation ladder
\end{minipage} & \begin{minipage}[t]{\linewidth}\RaggedRight\hspace{0pt}
supports robustness level if stable
\end{minipage} \\
\hline
\begin{minipage}[t]{\linewidth}\RaggedRight\hspace{0pt}
confound model sensitivity
\end{minipage} & \begin{minipage}[t]{\linewidth}\RaggedRight\hspace{0pt}
review rule or soft constraint
\end{minipage} & \begin{minipage}[t]{\linewidth}\RaggedRight\hspace{0pt}
caveat if attenuated
\end{minipage} \\
\hline
\begin{minipage}[t]{\linewidth}\RaggedRight\hspace{0pt}
permutation/\allowbreak{}null check
\end{minipage} & \begin{minipage}[t]{\linewidth}\RaggedRight\hspace{0pt}
hard statistical validity
\end{minipage} & \begin{minipage}[t]{\linewidth}\RaggedRight\hspace{0pt}
accept/\allowbreak{}block depending on result
\end{minipage} \\
\hline
\begin{minipage}[t]{\linewidth}\RaggedRight\hspace{0pt}
post-selection correction
\end{minipage} & \begin{minipage}[t]{\linewidth}\RaggedRight\hspace{0pt}
replayed configurations
\end{minipage} & \begin{minipage}[t]{\linewidth}\RaggedRight\hspace{0pt}
accept/\allowbreak{}block depending on result
\end{minipage} \\
\hline
\begin{minipage}[t]{\linewidth}\RaggedRight\hspace{0pt}
external dataset validation
\end{minipage} & \begin{minipage}[t]{\linewidth}\RaggedRight\hspace{0pt}
L4 ladder
\end{minipage} & \begin{minipage}[t]{\linewidth}\RaggedRight\hspace{0pt}
no external claim if deferred
\end{minipage} \\
\end{longtable}
}

{
\footnotesize
\setlength{\tabcolsep}{3pt}
\renewcommand{\arraystretch}{1.15}
\begin{longtable}{|
  >{\RaggedRight\hspace{0pt}\arraybackslash}p{(\linewidth - 7\tabcolsep) * \real{0.3164}}|
  >{\RaggedRight\hspace{0pt}\arraybackslash}p{(\linewidth - 7\tabcolsep) * \real{0.3148}}|
  >{\RaggedRight\hspace{0pt}\arraybackslash}p{(\linewidth - 7\tabcolsep) * \real{0.3212}}|}
\hline
\rowcolor{brtablehead}
\begin{minipage}[t]{\linewidth}\RaggedRight\hspace{0pt}
\textbf{Artifact class}
\end{minipage} & \begin{minipage}[t]{\linewidth}\RaggedRight\hspace{0pt}
\textbf{Required for review?}
\end{minipage} & \begin{minipage}[t]{\linewidth}\RaggedRight\hspace{0pt}
\textbf{Review consequence if missing}
\end{minipage} \\
\hline
\endhead
\hline
\endlastfoot

\begin{minipage}[t]{\linewidth}\RaggedRight\hspace{0pt}
Scorecard
\end{minipage} & \begin{minipage}[t]{\linewidth}\RaggedRight\hspace{0pt}
yes
\end{minipage} & \begin{minipage}[t]{\linewidth}\RaggedRight\hspace{0pt}
review not ready or revise
\end{minipage} \\
\hline
\begin{minipage}[t]{\linewidth}\RaggedRight\hspace{0pt}
Predictions
\end{minipage} & \begin{minipage}[t]{\linewidth}\RaggedRight\hspace{0pt}
yes for prediction
\end{minipage} & \begin{minipage}[t]{\linewidth}\RaggedRight\hspace{0pt}
statistical review incomplete
\end{minipage} \\
\hline
\begin{minipage}[t]{\linewidth}\RaggedRight\hspace{0pt}
Null-test output
\end{minipage} & \begin{minipage}[t]{\linewidth}\RaggedRight\hspace{0pt}
yes if claimed
\end{minipage} & \begin{minipage}[t]{\linewidth}\RaggedRight\hspace{0pt}
revise/\allowbreak{}block
\end{minipage} \\
\hline
\begin{minipage}[t]{\linewidth}\RaggedRight\hspace{0pt}
Config manifest
\end{minipage} & \begin{minipage}[t]{\linewidth}\RaggedRight\hspace{0pt}
yes
\end{minipage} & \begin{minipage}[t]{\linewidth}\RaggedRight\hspace{0pt}
reproducibility warning or revise
\end{minipage} \\
\hline
\begin{minipage}[t]{\linewidth}\RaggedRight\hspace{0pt}
QC report
\end{minipage} & \begin{minipage}[t]{\linewidth}\RaggedRight\hspace{0pt}
yes or conditional
\end{minipage} & \begin{minipage}[t]{\linewidth}\RaggedRight\hspace{0pt}
measurement WARN
\end{minipage} \\
\end{longtable}
}

\begin{quote}
artifact\_\allowbreak{}review:\\
expected\_\allowbreak{}artifacts: 6\\
produced\_\allowbreak{}artifacts: 6\\
artifact\_\allowbreak{}completeness\_\allowbreak{}ratio: 1.00\\
review\_\allowbreak{}ready: true
\end{quote}

\subsection{\texorpdfstring{\textbf{G7. Claim-family and claim-calibration table}}{G7. Claim-family and claim-calibration table}}\label{g7.-claim-family-and-claim-calibration-table}

{
\footnotesize
\setlength{\tabcolsep}{3pt}
\renewcommand{\arraystretch}{1.15}
\begin{longtable}{|
  >{\RaggedRight\hspace{0pt}\arraybackslash}p{(\linewidth - 7\tabcolsep) * \real{0.3164}}|
  >{\RaggedRight\hspace{0pt}\arraybackslash}p{(\linewidth - 7\tabcolsep) * \real{0.3164}}|
  >{\RaggedRight\hspace{0pt}\arraybackslash}p{(\linewidth - 7\tabcolsep) * \real{0.3196}}|}
\hline
\rowcolor{brtablehead}
\begin{minipage}[t]{\linewidth}\RaggedRight\hspace{0pt}
\textbf{Candidate claim}
\end{minipage} & \begin{minipage}[t]{\linewidth}\RaggedRight\hspace{0pt}
\textbf{Overclaim blocked}
\end{minipage} & \begin{minipage}[t]{\linewidth}\RaggedRight\hspace{0pt}
\textbf{Allowed language}
\end{minipage} \\
\hline
\endhead
\hline
\endlastfoot

\begin{minipage}[t]{\linewidth}\RaggedRight\hspace{0pt}
Model predicts behavior
\end{minipage} & \begin{minipage}[t]{\linewidth}\RaggedRight\hspace{0pt}
clinical biomarker
\end{minipage} & \begin{minipage}[t]{\linewidth}\RaggedRight\hspace{0pt}
internally validated predictive association
\end{minipage} \\
\hline
\begin{minipage}[t]{\linewidth}\RaggedRight\hspace{0pt}
FC is associated with score
\end{minipage} & \begin{minipage}[t]{\linewidth}\RaggedRight\hspace{0pt}
causal mechanism
\end{minipage} & \begin{minipage}[t]{\linewidth}\RaggedRight\hspace{0pt}
exploratory or condition-qualified association
\end{minipage} \\
\hline
\begin{minipage}[t]{\linewidth}\RaggedRight\hspace{0pt}
Result robust to parcellation
\end{minipage} & \begin{minipage}[t]{\linewidth}\RaggedRight\hspace{0pt}
external validity
\end{minipage} & \begin{minipage}[t]{\linewidth}\RaggedRight\hspace{0pt}
robustness under tested specifications
\end{minipage} \\
\hline
\begin{minipage}[t]{\linewidth}\RaggedRight\hspace{0pt}
Null-control passed
\end{minipage} & \begin{minipage}[t]{\linewidth}\RaggedRight\hspace{0pt}
biological truth
\end{minipage} & \begin{minipage}[t]{\linewidth}\RaggedRight\hspace{0pt}
leakage/\allowbreak{}null-controlled internal result
\end{minipage} \\
\end{longtable}
}

\subsection{\texorpdfstring{\textbf{G8. Final review verdict and revision routing}}{G8. Final review verdict and revision routing}}\label{g8.-final-review-verdict-and-revision-routing}

{
\footnotesize
\setlength{\tabcolsep}{3pt}
\renewcommand{\arraystretch}{1.15}
\begin{longtable}{|
  >{\RaggedRight\hspace{0pt}\arraybackslash}p{(\linewidth - 7\tabcolsep) * \real{0.3196}}|
  >{\RaggedRight\hspace{0pt}\arraybackslash}p{(\linewidth - 7\tabcolsep) * \real{0.3196}}|
  >{\RaggedRight\hspace{0pt}\arraybackslash}p{(\linewidth - 7\tabcolsep) * \real{0.3148}}|}
\hline
\rowcolor{brtablehead}
\begin{minipage}[t]{\linewidth}\RaggedRight\hspace{0pt}
\textbf{Verdict}
\end{minipage} & \begin{minipage}[t]{\linewidth}\RaggedRight\hspace{0pt}
\textbf{Meaning}
\end{minipage} & \begin{minipage}[t]{\linewidth}\RaggedRight\hspace{0pt}
\textbf{Next action}
\end{minipage} \\
\hline
\endhead
\hline
\endlastfoot

\begin{minipage}[t]{\linewidth}\RaggedRight\hspace{0pt}
accept
\end{minipage} & \begin{minipage}[t]{\linewidth}\RaggedRight\hspace{0pt}
Artifacts and claims satisfy active constraints.
\end{minipage} & \begin{minipage}[t]{\linewidth}\RaggedRight\hspace{0pt}
Eligible for memory.
\end{minipage} \\
\hline
\begin{minipage}[t]{\linewidth}\RaggedRight\hspace{0pt}
accept with qualifications
\end{minipage} & \begin{minipage}[t]{\linewidth}\RaggedRight\hspace{0pt}
Result usable only with caveats.
\end{minipage} & \begin{minipage}[t]{\linewidth}\RaggedRight\hspace{0pt}
Qualified memory claim.
\end{minipage} \\
\hline
\begin{minipage}[t]{\linewidth}\RaggedRight\hspace{0pt}
revise
\end{minipage} & \begin{minipage}[t]{\linewidth}\RaggedRight\hspace{0pt}
Fixable issue remains.
\end{minipage} & \begin{minipage}[t]{\linewidth}\RaggedRight\hspace{0pt}
Return to planning or execution.
\end{minipage} \\
\hline
\begin{minipage}[t]{\linewidth}\RaggedRight\hspace{0pt}
block
\end{minipage} & \begin{minipage}[t]{\linewidth}\RaggedRight\hspace{0pt}
Hard validity failure.
\end{minipage} & \begin{minipage}[t]{\linewidth}\RaggedRight\hspace{0pt}
Stop or redesign.
\end{minipage} \\
\hline
\begin{minipage}[t]{\linewidth}\RaggedRight\hspace{0pt}
reject
\end{minipage} & \begin{minipage}[t]{\linewidth}\RaggedRight\hspace{0pt}
Claim unsupported or contradicted.
\end{minipage} & \begin{minipage}[t]{\linewidth}\RaggedRight\hspace{0pt}
No writeback.
\end{minipage} \\
\hline
\begin{minipage}[t]{\linewidth}\RaggedRight\hspace{0pt}
escalate
\end{minipage} & \begin{minipage}[t]{\linewidth}\RaggedRight\hspace{0pt}
Human expert needed.
\end{minipage} & \begin{minipage}[t]{\linewidth}\RaggedRight\hspace{0pt}
Pause or route.
\end{minipage} \\
\end{longtable}
}

\begin{quote}
review\_\allowbreak{}card:\\
episode\_\allowbreak{}id: ep\_\allowbreak{}hcp\_\allowbreak{}predict\_\allowbreak{}001\\
run\_\allowbreak{}id: run\_\allowbreak{}hcp\_\allowbreak{}pathb\_\allowbreak{}001\\
verdict: accept\_\allowbreak{}with\_\allowbreak{}qualifications\\
\strut \\
block\_\allowbreak{}findings: {[}{]}\\
warn\_\allowbreak{}findings:\\
- external\_\allowbreak{}validation\_\allowbreak{}missing\\
\strut \\
robustness\_\allowbreak{}checks:\\
family\_\allowbreak{}block\_\allowbreak{}permutation: pass\\
post\_\allowbreak{}selection\_\allowbreak{}correction: pass\\
external\_\allowbreak{}validation: deferred\\
\strut \\
claim\_\allowbreak{}calibration:\\
allowed\_\allowbreak{}claim\_\allowbreak{}strength: "validated internal result"\\
prohibited\_\allowbreak{}claims:\\
- "externally validated biomarker"\\
- "causal mechanism"\\
- "clinical prediction claim"\\
\strut \\
memory\_\allowbreak{}eligibility: qualified\_\allowbreak{}claim\_\allowbreak{}only
\end{quote}

\clearpage
\subsection{\texorpdfstring{\textbf{G9. Scientific review rule registry (release) and calibration library}}{G9. Scientific review rule registry (release) and calibration library}}\label{g9.-full-scientific-review-rule-registry}

This appendix gives the reader-facing view of the scientific review rule registry used to define, calibrate, and prioritize Brain Researcher's verification layer. The full per-rule registry, its organization fields, rule families, encoded policy decisions, implementation priority queue, and review-context fields and sensitivity templates, is released as a machine-readable file in the archival repository (Data availability), so it is not reprinted here. A rule's presence in the released registry does not mean that it is enforced in every episode; the lifecycle table in G4 distinguishes implemented rules, deterministic candidates, schema-dependent candidates, text-interpretation candidates, and calibration-only cases. Retained below is the calibration case library (G9.5), because the main text points to it for the first false-accept and false-block accounting.

\subsubsection{\texorpdfstring{\textbf{G9.5 Calibration case library}}{G9.5 Calibration case library}}

The registry includes 60 calibration cases (C01--C60) for annotator training and regression testing. They are summarized by rule family here; the machine-readable release contains the case-level labels, severity, novelty flag, and rule identifiers.

\begin{longtable}{|>{\RaggedRight\hspace{0pt}\arraybackslash}p{0.18\textwidth}|>{\RaggedRight\hspace{0pt}\arraybackslash}p{0.48\textwidth}|>{\RaggedRight\hspace{0pt}\arraybackslash}p{0.24\textwidth}|}
\hline
\rowcolor{brtablehead}\textbf{Cases} & \textbf{Scenarios covered} & \textbf{Typical label} \\
\hline
C01--C06 & Repeated-measures, mixed-design, longitudinal, and spatial-domain errors, plus valid paired-test and motor-task controls. & BLOCK for invalid design or spatial correction; allow for valid controls. \\
\hline
C07--C12 & Extreme or unexpected effects, ordinary morphometry effects, motion-confounded small-sample FC, and expected task activation. & WARN for prior conflict or confounding; allow for expected, well-scoped effects. \\
\hline
C13--C24 & Whole-brain correction, cluster-threshold, localization, double-dipping, ROI multiplicity, analytic flexibility, pseudoreplication, fixed-effects population claims, exchangeability, and spatial-null examples. & BLOCK for hard inference errors; WARN for soft multiplicity or reporting issues. \\
\hline
C25--C36 & Motion reporting, motion imbalance, global signal regression, dynamic FC, graph thresholding, task FC/PPI, EPI distortion, MRIQC, small sample, missing external validation, and multiband QC. & WARN except for high-confidence motion confounds that block inference. \\
\hline
C37--C41 & Reverse inference, stimulus generalization, missing behavioral covariates, ICV/TIV omission, and multi-site site confounding. & WARN with claim narrowing or added covariates/sensitivity. \\
\hline
C42--C55 & Harmonization, feature selection, standardization, test-set selection, split grouping, permutation, fold-wise error, prediction language, temporal splits, post-hoc layer selection, RSA multiplicity and hierarchy, and missing ICC. & BLOCK for leakage and test-set use; WARN for missing reliability, uncertainty, or permutation support. \\
\hline
C56--C60 & Extreme effects, single-pipeline significance, COBIDAS fields, BIDS validation, and BIDS Stats Models reporting. & WARN for incomplete robustness or reporting; allow/positive modifier for structured reporting. \\
\hline
\end{longtable}

\clearpage

\section{\texorpdfstring{\textbf{Appendix I. Operational-mode, bounded-episode, and Harbor-based validation card}}{Appendix I. Operational-mode, bounded-episode, and Harbor-based validation card}}\label{appendix-i.-operational-mode-bounded-episode-and-harbor-style-validation-card}\label{app:i}

\hyperref[app:i]{Appendix I} is the operational control trace: it records how an episode or campaign was bounded, budgeted, advanced, verified, stopped, escalated, and permitted or denied memory writeback. The bounded self-evolving and loop-closure schematics are summarized in Supplementary Figs.~\ref{fig:self-evolving-workflow}--\ref{fig:result-to-hypothesis}.

\subsection{\texorpdfstring{\textbf{I1. Purpose and scope}}{I1. Purpose and scope}}\label{i1.-purpose-and-scope}

This card records the operational mode and control contract for interactive, benchmark, collaborator-facing, and bounded autonomous episodes. It is especially important for Harbor-based validation and supervisor/\allowbreak{}critic cycles. Harbor verification is recorded as task-level execution evidence, not scientific acceptance.

\subsection{\texorpdfstring{\textbf{I2. Operational-mode identity}}{I2. Operational-mode identity}}\label{i2.-operational-mode-identity}

{
\footnotesize
\setlength{\tabcolsep}{3pt}
\renewcommand{\arraystretch}{1.15}
\begin{longtable}{|
  >{\RaggedRight\hspace{0pt}\arraybackslash}p{(\linewidth - 5\tabcolsep) * \real{0.4556}}|
  >{\RaggedRight\hspace{0pt}\arraybackslash}p{(\linewidth - 5\tabcolsep) * \real{0.4957}}|}
\hline
\rowcolor{brtablehead}
\begin{minipage}[t]{\linewidth}\RaggedRight\hspace{0pt}
\textbf{Field}
\end{minipage} & \begin{minipage}[t]{\linewidth}\RaggedRight\hspace{0pt}
\textbf{Recommended entry}
\end{minipage} \\
\hline
\endhead
\hline
\endlastfoot

\begin{minipage}[t]{\linewidth}\RaggedRight\hspace{0pt}
Episode or campaign ID
\end{minipage} & \begin{minipage}[t]{\linewidth}\RaggedRight\hspace{0pt}
ep\_\allowbreak{}hcp\_\allowbreak{}predict\_\allowbreak{}001 /\allowbreak{} campaign\_\allowbreak{}autonomous\_\allowbreak{}001
\end{minipage} \\
\hline
\begin{minipage}[t]{\linewidth}\RaggedRight\hspace{0pt}
Mode
\end{minipage} & \begin{minipage}[t]{\linewidth}\RaggedRight\hspace{0pt}
interactive /\allowbreak{} benchmark /\allowbreak{} collaborator /\allowbreak{} bounded autonomous
\end{minipage} \\
\hline
\begin{minipage}[t]{\linewidth}\RaggedRight\hspace{0pt}
Gate authority
\end{minipage} & \begin{minipage}[t]{\linewidth}\RaggedRight\hspace{0pt}
human researcher /\allowbreak{} evaluation harness /\allowbreak{} critic /\allowbreak{} supervisor
\end{minipage} \\
\hline
\begin{minipage}[t]{\linewidth}\RaggedRight\hspace{0pt}
Instruction or task contract
\end{minipage} & \begin{minipage}[t]{\linewidth}\RaggedRight\hspace{0pt}
instruction.md /\allowbreak{} bounded\_\allowbreak{}episode\_\allowbreak{}contract.yaml
\end{minipage} \\
\hline
\begin{minipage}[t]{\linewidth}\RaggedRight\hspace{0pt}
Allowed resources
\end{minipage} & \begin{minipage}[t]{\linewidth}\RaggedRight\hspace{0pt}
dataset/\allowbreak{}resource IDs and approved roots
\end{minipage} \\
\hline
\begin{minipage}[t]{\linewidth}\RaggedRight\hspace{0pt}
Action vocabulary
\end{minipage} & \begin{minipage}[t]{\linewidth}\RaggedRight\hspace{0pt}
search, plan, run, inspect, review, branch, stop
\end{minipage} \\
\hline
\begin{minipage}[t]{\linewidth}\RaggedRight\hspace{0pt}
Budget
\end{minipage} & \begin{minipage}[t]{\linewidth}\RaggedRight\hspace{0pt}
cycle, compute, wall-clock, retry, or token budget
\end{minipage} \\
\hline
\begin{minipage}[t]{\linewidth}\RaggedRight\hspace{0pt}
Stopping criteria
\end{minipage} & \begin{minipage}[t]{\linewidth}\RaggedRight\hspace{0pt}
success, unresolved block, no progress, budget exhausted, escalation
\end{minipage} \\
\end{longtable}
}

\subsection{\texorpdfstring{\textbf{I3. Bounded episode contract}}{I3. Bounded episode contract}}\label{i3.-bounded-episode-contract}

{
\footnotesize
\setlength{\tabcolsep}{3pt}
\renewcommand{\arraystretch}{1.15}
\begin{longtable}{|
  >{\RaggedRight\hspace{0pt}\arraybackslash}p{(\linewidth - 5\tabcolsep) * \real{0.4701}}|
  >{\RaggedRight\hspace{0pt}\arraybackslash}p{(\linewidth - 5\tabcolsep) * \real{0.4813}}|}
\hline
\rowcolor{brtablehead}
\begin{minipage}[t]{\linewidth}\RaggedRight\hspace{0pt}
\textbf{Contract field}
\end{minipage} & \begin{minipage}[t]{\linewidth}\RaggedRight\hspace{0pt}
\textbf{What it records}
\end{minipage} \\
\hline
\endhead
\hline
\endlastfoot

\begin{minipage}[t]{\linewidth}\RaggedRight\hspace{0pt}
Target question
\end{minipage} & \begin{minipage}[t]{\linewidth}\RaggedRight\hspace{0pt}
The scientific objective for the bounded episode.
\end{minipage} \\
\hline
\begin{minipage}[t]{\linewidth}\RaggedRight\hspace{0pt}
Admissible design space
\end{minipage} & \begin{minipage}[t]{\linewidth}\RaggedRight\hspace{0pt}
Which analyses and branches are allowed.
\end{minipage} \\
\hline
\begin{minipage}[t]{\linewidth}\RaggedRight\hspace{0pt}
Cycle budget
\end{minipage} & \begin{minipage}[t]{\linewidth}\RaggedRight\hspace{0pt}
Maximum number of supervisor/\allowbreak{}critic cycles.
\end{minipage} \\
\hline
\begin{minipage}[t]{\linewidth}\RaggedRight\hspace{0pt}
Expected artifacts
\end{minipage} & \begin{minipage}[t]{\linewidth}\RaggedRight\hspace{0pt}
Files and records required for verification/\allowbreak{}review.
\end{minipage} \\
\hline
\begin{minipage}[t]{\linewidth}\RaggedRight\hspace{0pt}
Validation ladder
\end{minipage} & \begin{minipage}[t]{\linewidth}\RaggedRight\hspace{0pt}
Target and achieved L0-L5 level.
\end{minipage} \\
\hline
\begin{minipage}[t]{\linewidth}\RaggedRight\hspace{0pt}
Review rules
\end{minipage} & \begin{minipage}[t]{\linewidth}\RaggedRight\hspace{0pt}
BLOCK/\allowbreak{}WARN rules required for acceptance.
\end{minipage} \\
\hline
\begin{minipage}[t]{\linewidth}\RaggedRight\hspace{0pt}
Memory policy
\end{minipage} & \begin{minipage}[t]{\linewidth}\RaggedRight\hspace{0pt}
accepted\_\allowbreak{}only /\allowbreak{} qualified\_\allowbreak{}only /\allowbreak{} no\_\allowbreak{}write.
\end{minipage} \\
\hline
\begin{minipage}[t]{\linewidth}\RaggedRight\hspace{0pt}
Escalation conditions
\end{minipage} & \begin{minipage}[t]{\linewidth}\RaggedRight\hspace{0pt}
When to route to a human researcher.
\end{minipage} \\
\end{longtable}
}

operational\_\allowbreak{}mode\_\allowbreak{}card:\\
episode\_\allowbreak{}id: ep\_\allowbreak{}hcp\_\allowbreak{}predict\_\allowbreak{}001\\
mode: bounded\_\allowbreak{}autonomous\\
gate\_\allowbreak{}authority: critic\\
action\_\allowbreak{}vocabulary:\\
- resolve\_\allowbreak{}dataset\\
- select\_\allowbreak{}workflow\\
- run\_\allowbreak{}analysis\\
- inspect\_\allowbreak{}artifacts\\
- request\_\allowbreak{}review\\
- branch\_\allowbreak{}sensitivity\\
- terminate\\
\strut \\
budgets:\\
cycle\_\allowbreak{}budget: 3\\
wall\_\allowbreak{}clock\_\allowbreak{}budget: fixed\\
compute\_\allowbreak{}budget: fixed\\
\strut \\
stopping\_\allowbreak{}criteria:\\
- validation\_\allowbreak{}ladder\_\allowbreak{}reached\\
- hard\_\allowbreak{}review\_\allowbreak{}block\_\allowbreak{}unresolved\\
- no\_\allowbreak{}progress\_\allowbreak{}after\_\allowbreak{}cycle\_\allowbreak{}limit\\
- compute\_\allowbreak{}budget\_\allowbreak{}exhausted\\
- human\_\allowbreak{}escalation\_\allowbreak{}required

\subsection{\texorpdfstring{\textbf{I4. Supervisor/critic decisions and Harbor verifier output}}{I4. Supervisor/critic decisions and Harbor verifier output}}\label{i4.-supervisorcritic-decisions-and-harbor-verifier-output}

{
\footnotesize
\setlength{\tabcolsep}{3pt}
\renewcommand{\arraystretch}{1.15}
\begin{longtable}{|
  >{\RaggedRight\hspace{0pt}\arraybackslash}p{(\linewidth - 5\tabcolsep) * \real{0.4717}}|
  >{\RaggedRight\hspace{0pt}\arraybackslash}p{(\linewidth - 5\tabcolsep) * \real{0.4797}}|}
\hline
\rowcolor{brtablehead}
\begin{minipage}[t]{\linewidth}\RaggedRight\hspace{0pt}
\textbf{Record}
\end{minipage} & \begin{minipage}[t]{\linewidth}\RaggedRight\hspace{0pt}
\textbf{Recommended fields}
\end{minipage} \\
\hline
\endhead
\hline
\endlastfoot

\begin{minipage}[t]{\linewidth}\RaggedRight\hspace{0pt}
Branch-decision record
\end{minipage} & \begin{minipage}[t]{\linewidth}\RaggedRight\hspace{0pt}
branch type; reason; proposed action; expected resolution; budget cost
\end{minipage} \\
\hline
\begin{minipage}[t]{\linewidth}\RaggedRight\hspace{0pt}
Critic decision
\end{minipage} & \begin{minipage}[t]{\linewidth}\RaggedRight\hspace{0pt}
continue /\allowbreak{} investigate anomaly /\allowbreak{} broaden /\allowbreak{} revise /\allowbreak{} escalate /\allowbreak{} terminate
\end{minipage} \\
\hline
\begin{minipage}[t]{\linewidth}\RaggedRight\hspace{0pt}
Harbor verifier output
\end{minipage} & \begin{minipage}[t]{\linewidth}\RaggedRight\hspace{0pt}
artifact manifest present; required statistic present; schema valid; command completed; reward or partial credit
\end{minipage} \\
\hline
\begin{minipage}[t]{\linewidth}\RaggedRight\hspace{0pt}
Validation-ladder status
\end{minipage} & \begin{minipage}[t]{\linewidth}\RaggedRight\hspace{0pt}
highest achieved level and unmet next gate
\end{minipage} \\
\hline
\begin{minipage}[t]{\linewidth}\RaggedRight\hspace{0pt}
Termination status
\end{minipage} & \begin{minipage}[t]{\linewidth}\RaggedRight\hspace{0pt}
completed /\allowbreak{} blocked /\allowbreak{} budget exhausted /\allowbreak{} escalated /\allowbreak{} terminated
\end{minipage} \\
\end{longtable}
}

harbor\_\allowbreak{}verifier:\\
artifact\_\allowbreak{}manifest\_\allowbreak{}present: true\\
required\_\allowbreak{}statistic\_\allowbreak{}present: true\\
schema\_\allowbreak{}valid: true\\
command\_\allowbreak{}completed: true\\
partial\_\allowbreak{}credit: 0.8\\
\strut \\
critic\_\allowbreak{}decision:\\
decision: terminate\_\allowbreak{}with\_\allowbreak{}qualified\_\allowbreak{}claim\\
reason: "L3 internal validation reached; external validation deferred."\\
\strut \\
final\_\allowbreak{}status: completed\_\allowbreak{}with\_\allowbreak{}qualifications\\
memory\_\allowbreak{}writeback\_\allowbreak{}allowed: qualified\_\allowbreak{}claim\_\allowbreak{}only

\section{\texorpdfstring{\textbf{Appendix J. Evaluation protocol, scoring contracts, and metric ledger}}{Appendix J. Evaluation protocol, scoring contracts, and metric ledger}}\label{appendix-j.-evaluation-protocol-scoring-contracts-and-metric-ledger}\label{app:j}

\hyperref[app:j]{Appendix J} records the evaluation protocol used for the quantitative benchmarks, collaborator cases, and bounded self-evolving campaigns reported in the main text. It defines the evaluated surfaces, condition comparisons, metric denominators, and aggregate values used in main-text Fig.~3. The benchmark surfaces are reported separately because tool calling and evidence citation have different item types, denominators, scoring rules, and ceilings.

\subsection{\texorpdfstring{\textbf{J1. Evaluation surfaces}}{J1. Evaluation surfaces}}\label{j1.-evaluation-surfaces}

{
\footnotesize
\setlength{\tabcolsep}{3pt}
\renewcommand{\arraystretch}{1.15}
\begin{longtable}{|
  >{\RaggedRight\hspace{0pt}\arraybackslash}p{0.24\linewidth}|
  >{\RaggedRight\hspace{0pt}\arraybackslash}p{0.31\linewidth}|
  >{\RaggedRight\hspace{0pt}\arraybackslash}p{0.34\linewidth}|}
\hline
\rowcolor{brtablehead}\textbf{Surface} & \textbf{Unit} & \textbf{What is evaluated} \\
\hline
\endhead
\hline
\endlastfoot
Tool-calling benchmark & 60 neuroimaging task manifests; seven models per condition; 420 model-item trajectories per condition. & Whether a model maps a natural-language neuroimaging request to the correct analysis tool or executable route before execution. \\
\hline
Routing ablation without direct KG calls & Same 60-task manifest; seven model routes; 420 route--task episodes. & Exact-label top-1 route selection when direct KG-named calls are unavailable. \\
\hline
Evidence-citation benchmark & 76-item open-ended NeuroimageKnowledge manifest; descriptive Results aggregate reported on a 50-question analysis set over all generated evidence-basis rows. & Whether model-generated neuroimaging claims cite evidence that can be located and judged supportive by three condition-blind LLM judges (Claude Opus 4.8, Codex GPT-5.5, and Gemini 3.1 Pro). All reference-bearing rows were sent to all three judges; a row entered the numerator when at least two judges voted verified, while every generated evidence-basis row remained in the denominator. \\
\hline
Collaborator cases & Three active scientific questions from collaborating researchers. & Whether Brain Researcher returns bounded claim records rather than unqualified findings. \\
\hline
Bounded self-evolving campaigns & Two predeclared episode contracts. & Whether fixed validation gates reject, defer, or revise claims during exploratory search. \\
\hline
\end{longtable}
}

\subsection{\texorpdfstring{\textbf{J2. Benchmark conditions and model matrix}}{J2. Benchmark conditions and model matrix}}\label{j2.-paired-benchmark-conditions-and-model-matrix}

Both quantitative benchmarks compare with-BR and without-BR conditions on the same task or question set. The tool-calling benchmark is paired at the model--item level. The evidence-citation headline is instead a descriptive question-level aggregate over all generated evidence-basis rows; only rows receiving at least two verified votes contribute to the numerator, and it is not treated as a seven-model paired estimate. Benchmark memory was isolated from collaborator and bounded-campaign memory, and no benchmark item was promoted as a scientific claim.

{
\footnotesize
\setlength{\tabcolsep}{3pt}
\renewcommand{\arraystretch}{1.15}
\begin{longtable}{|
  >{\RaggedRight\hspace{0pt}\arraybackslash}p{0.25\linewidth}|
  >{\RaggedRight\hspace{0pt}\arraybackslash}p{0.65\linewidth}|}
\hline
\rowcolor{brtablehead}\textbf{Field} & \textbf{Entry} \\
\hline
\endhead
\hline
\endlastfoot
Conditions & Without-BR baseline; with-BR condition. \\
\hline
Model variants & Claude Code / Claude Opus 4.8; Codex GPT-5.5; Gemini 3.1 Pro; GLM-5.1; DeepSeek-V4-Pro; Kimi K2.5; Qwen3.6-Plus. \\
\hline
Tool-calling items & Natural-language neuroimaging requests requiring an analysis tool call or executable route with the required inputs available before execution. \\
\hline
Evidence-citation item families & Open-ended neuroimaging review questions covering preprocessing, statistical inference, functional connectivity, multivariate decoding, meta-analysis, and reproducibility. \\
\hline
Isolation policy & Benchmark runs use fixed task manifests, fixed scoring contracts, and isolated memory partitions. \\
\hline
\end{longtable}
}

\subsection{\texorpdfstring{\textbf{J2a. Tool-calling benchmark scoring summary}}{J2a. Tool-calling benchmark scoring summary}}\label{j2a.-tool-calling-benchmark-scoring-summary}

The tool-calling benchmark is retained as an aggregate routing surface, but the prompt manifest is not reproduced in the appendix. Each item supplied the agent with a short routing request; the reference route and capabilities were held out for scoring. The scoring contract credits any response that covers the required capabilities, whether through a Brain Researcher call or an equivalent executable route. The release bundle records the exact prompt surfaces and provenance files for independent re-scoring.

\subsection{\texorpdfstring{\textbf{J2b. NeuroimageKnowledge benchmark manifests (released)}}{J2b. NeuroimageKnowledge benchmark manifests (released)}}\label{j2b.-neuroimageknowledge-benchmark-item-manifest}

The full NeuroimageKnowledge groundable item manifest (the open-ended scenarios passed to the benchmark harness), representative paired with-BR/without-BR answers, the unified benchmark-bundle artifact inventory, and representative benchmark cases are released as machine-readable manifests in the benchmark package (Data availability), so they are not reprinted here. These manifests document the paired prompts used to elicit answers and evidence-basis rows under both conditions; they are not used for automatic correctness scoring, and the reported 50-question aggregate is defined by the scoring contracts below. Representative paired examples illustrate the interpretation used in the Results: strong models often answer canonical methods questions well without BR, while BR changes the evidence boundary by adding resolvable, source-supported rows, including cases where the without-BR answer wins the paired judgment because a memory-based citation resolves and supports the claim.

\subsection{\texorpdfstring{\textbf{J3. Scoring contracts and denominators}}{J3. Scoring contracts and denominators}}\label{j3.-scoring-contracts-and-denominators}

{
\footnotesize
\setlength{\tabcolsep}{3pt}
\renewcommand{\arraystretch}{1.15}
\begin{longtable}{|
  >{\RaggedRight\hspace{0pt}\arraybackslash}p{0.315\linewidth}|
  >{\RaggedRight\hspace{0pt}\arraybackslash}p{0.34\linewidth}|
  >{\RaggedRight\hspace{0pt}\arraybackslash}p{0.245\linewidth}|}
\hline
\rowcolor{brtablehead}\textbf{Metric} & \textbf{Numerator} & \textbf{Denominator} \\
\hline
\endhead
\hline
\endlastfoot
Capability@k & Fraction of required task capabilities covered by the selected call or route after the first k non-neutral actions. & Model-item trajectories in the tool-calling benchmark; values are averaged as coverage scores. \\
\hline
Correct route/tool@k & Tasks that reach exact task success under the capability scorer after the first k non-neutral actions. & All tool-calling model-item trajectories. \\
\hline
Restricted-interface exact-label top-1 & Requested routes with an exact top-1 label match under \texttt{route\_search}$\rightarrow$\texttt{tool\_search}; protocol violations count as errors. & All 60 task-route episodes; gold candidate availability is reported separately as an oracle-coverage diagnostic. \\
\hline
Handoff score@k & Selected call or route contains enough information for a receiving agent to continue the analysis after the first k non-neutral actions. & Tool-calling trajectories with parsed actions. \\
\hline
\texttt{verified\_groundedness\_rate} & Evidence items whose cited evidence is both locatable and judged supportive. & All evidence-basis rows generated for each question, followed by an equal-weight mean across the 50 questions. \\
\hline
\texttt{verified\_among\_claimed\_grounded} & Model-claimed grounded items whose cited evidence is both locatable and judged supportive. & Model-claimed grounded items. \\
\hline
\texttt{citation\_spam\_rate} & Model-claimed grounded items whose cited evidence is judged unrelated to the claim. & Model-claimed grounded items. \\
\hline
\texttt{answer\_correctness\_rate} & Answers passing the rubric-based source sanity check. & Scored NeuroimageKnowledge outputs with parseable answer fields. \\
\hline
\end{longtable}
}

\subsection{\texorpdfstring{\textbf{J4. Aggregate benchmark values reported in Fig.~3}}{J4. Aggregate benchmark values reported in the benchmark summary figure}}\label{j4.-aggregate-benchmark-values-reported-in-the-benchmark-summary-figure}

{
\footnotesize
\setlength{\tabcolsep}{3pt}
\renewcommand{\arraystretch}{1.15}
\begin{longtable}{|
  >{\RaggedRight\hspace{0pt}\arraybackslash}p{0.36\linewidth}|
  >{\RaggedRight\hspace{0pt}\arraybackslash}p{0.20\linewidth}|
  >{\RaggedRight\hspace{0pt}\arraybackslash}p{0.20\linewidth}|
  >{\RaggedRight\hspace{0pt}\arraybackslash}p{0.12\linewidth}|}
\hline
\rowcolor{brtablehead}\textbf{Metric} & \textbf{Without BR} & \textbf{With BR} & \textbf{Direction} \\
\hline
\endhead
\hline
\endlastfoot
Capability@1 (all tool-calling trajectories) & 0.498 & 0.945 & Higher \\
\hline
Correct route/tool@1 & 0.233 & 0.936 & Higher \\
\hline
Handoff score@1 & 0.474 & 0.761 & Higher \\
\hline
\texttt{verified\_groundedness\_rate} (three-judge majority; 50-question mean) & 0.046 & 0.220 & Higher \\
\hline
\texttt{verified\_groundedness\_rate} (Gemini 3.1 Pro recovered judge; 50-question mean) & 0.0435 & 0.2097 & Higher \\
\hline
\texttt{verified\_groundedness\_rate} (Codex GPT-5.5 judge; 50-question mean) & 0.0389 & 0.3088 & Higher \\
\hline
\texttt{verified\_groundedness\_rate} (Claude Opus 4.8 judge; 50-question mean) & 0.0796 & 0.2187 & Higher \\
\hline
\texttt{verified\_among\_claimed\_grounded} (Gemini 2.5 Flash single-judge safeguard; by-model mean) & 0.273 & 0.583 & Higher \\
\hline
\texttt{citation\_spam\_rate} (Gemini 2.5 Flash single-judge safeguard; by-model mean) & 0.317 & 0.249 & Lower \\
\hline
\texttt{answer\_correctness\_rate} (rubric-based safeguard; by-model mean) & 0.749 & 0.789 & Higher \\
\hline
\end{longtable}
}

Tool-calling action-budget metrics are trajectory-level quantities over 420 model-item trajectories per condition. The grounding headline and judge-specific rates are equal-weight means of the per-question fractions over all generated evidence-basis rows; only rows receiving at least two verified votes contribute to the headline numerator, while all other rows add no verified count. A separate exact-label diagnostic among the 444 non-verified with-BR rows present in all three judge outputs assigned a \emph{no\_unrelated} majority to 289 (65\%), a \emph{partial} majority to 124 (28\%), and no exact-label majority or \texttt{cannot\_judge} majority to the remaining 31 (7\%); no row had a fabricated or malformed exact-label majority. The precision and citation-spam safeguards use the earlier Gemini 2.5 Flash single-judge scoring and are not three-judge-majority estimates. Evidence-citation metrics use evidence-item, claimed-grounded-item, or answer-output denominators as specified above and are not pooled with tool-calling metrics.

\subsection{\texorpdfstring{\textbf{J5. Collaborator-case ledger}}{J5. Collaborator-case ledger}}\label{j5.-collaborator-case-ledger}

{
\footnotesize
\setlength{\tabcolsep}{3pt}
\renewcommand{\arraystretch}{1.15}
\begin{longtable}{|
  >{\RaggedRight\hspace{0pt}\arraybackslash}p{0.18\linewidth}|
  >{\RaggedRight\hspace{0pt}\arraybackslash}p{0.29\linewidth}|
  >{\RaggedRight\hspace{0pt}\arraybackslash}p{0.42\linewidth}|}
\hline
\rowcolor{brtablehead}\textbf{Case} & \textbf{Evaluation object} & \textbf{Reported outcome boundary} \\
\hline
\endhead
\hline
\endlastfoot
NeuroMark schizophrenia FNC & FBIRN cohort, N = 363; 480-specification multiverse over connectivity, confound, dimensionality-reduction, classifier, and domain granularity. & NM-H1, NM-H2, and NM-H3 all recorded as qualified (NM-H2 favorable in 12 of 24 contrasts after sign-aware rescoring); support depended on analytic regime. \\
\hline
Cocaine-use-disorder connectivity & SUDMEX CONN / OpenNeuro ds003346, N = 138; 36-specification multiverse over atlases, framewise-displacement thresholds, and confound strategies. & Five prespecified network-systemic-segregation associations rejected under SDMA-GLS (all Z \ensuremath{<} 1.24, FDR q \ensuremath{>} 0.58); exploratory 70-combination screen yielded no FDR-surviving effects (max \ensuremath{Z_{\mathrm{GLS}} = 2.18}; 0/70 surviving). \\
\hline
Cross-cultural social cognition & 21 published studies, 85 MNI coordinates, four culture-by-relationship ALE cells with k = 6--8 cell-level entries. & Mechanistic mPFC-topology interpretation blocked as exploratory because cell counts were below recommended ALE stability thresholds and paradigm composition was imbalanced. \\
\hline
\end{longtable}
}

\subsection{\texorpdfstring{\textbf{J6. Bounded self-evolving campaign ledger}}{J6. Bounded self-evolving campaign ledger}}\label{j6.-bounded-self-evolving-campaign-ledger}

{
\footnotesize
\setlength{\tabcolsep}{3pt}
\renewcommand{\arraystretch}{1.15}
\begin{longtable}{|
  >{\RaggedRight\hspace{0pt}\arraybackslash}p{0.20\linewidth}|
  >{\RaggedRight\hspace{0pt}\arraybackslash}p{0.31\linewidth}|
  >{\RaggedRight\hspace{0pt}\arraybackslash}p{0.38\linewidth}|}
\hline
\rowcolor{brtablehead}\textbf{Campaign} & \textbf{Predeclared controls} & \textbf{Claim-status outcome} \\
\hline
\endhead
\hline
\endlastfoot
HCP-YA workflow search and transfer & Two staged search bundles with all 116 slots retained in the denominator (104 scored in parent runs; 12 transport failures); frozen selected workflow designated before a locally matched comparison; repeated family-grouped nested CV; frozen selected workflow refits across four additional outcomes; separate internal holdout. & The frozen selected workflow exceeded the matched comparator in 10/10 Cognition splits and in 37/40 additional-outcome splits, but the result remained retrospective and same-cohort, no transfer cell survived weak-FWER correction, and the separate holdout did not confirm the comparator claim. No external scientific acceptance. \\
\hline
TRIBE speech--tools geometry & Complete 15-pair six-category discovery; researcher-authorized post-hoc contrast choice; frozen geometry estimand; three non-overlapping recurring-source panels; score-blind acoustic balancing and Holm-corrected permutation inference on four previously unused collections. & Three recurring-source panels met bounded support (11/12 collection cells in the predicted direction), but the frozen four-new-source H1 was not supported after correction (Holm $p=.396$). Terminal outcome inconclusive or conflicting; no scientific acceptance. \\
\hline
\end{longtable}
}

\subsection{\texorpdfstring{\textbf{J7. Exclusions and denominator handling}}{J7. Exclusions and denominator handling}}\label{j7.-exclusions-and-denominator-handling}

Non-executable or blocked cases are retained in the relevant denominator unless the scoring contract explicitly defines a successful-run subset, as in the first-correct budget. Resource failures, failed preflight checks, missing required artifacts, governance blocks, and invalid tool routes are recorded as outcomes rather than silently removed. For evidence citation, claims without locatable evidence remain in the all-claims denominator for verified groundedness.

\section{Appendix K. Cocaine-use-disorder case: representative generated code}\label{app:k}

The listing below is representative analysis code that Brain Researcher generated and executed for the cocaine-use-disorder case, a resting-state functional-connectivity multiverse on the SUDMEX cohort (CUD versus HC). The excerpt captures the multiverse specification: a header describing the design (cortical and subcortical atlas configurations crossed with framewise-displacement thresholds, aggregated by SDMA-GLS; this representative excerpt shows the atlas-by-FD grid, while the full 36-specification multiverse additionally crosses confound strategies), the 10-network definitions, the atlas-configuration grid, and a representative core function signature. Templated project tokens (a dollar-brace project-directory variable) are shown verbatim, and concrete filesystem locations are written as the generic placeholder \texttt{<DATA\_ROOT>}. The complete code is released per the Code availability statement.

\begin{suppcode}
"""
Multiverse Functional Connectivity Analysis -- SUDMEX CUD vs HC
===============================================================
Specifications = atlas configs x FD motion thresholds (SDMA-GLS aggregated)

Atlas configs:
  Sch200_Tian3  : Schaefer-200 + Tian Scale III (250 parcels)
  Sch400_Tian3  : Schaefer-400 + Tian Scale III (450 parcels)  <- primary
  Sch600_Tian3  : Schaefer-600 + Tian Scale III (650 parcels)
  Sch400_HCP    : Schaefer-400 + HCP subcortical (419 parcels)
  Gordon_Tian3  : Gordon-333  + Tian Scale III  (383 parcels)
  Glasser_Tian3 : Glasser-360 + Tian Scale III  (410 parcels)

FD motion thresholds: post-hoc re-censoring of XCP-D timeseries.
XCP-D originally censored at FD=0.5; stricter thresholds drop more frames.
Aggregation: SDMA-GLS (Lefort-Besnard et al., 2025, Imaging Neuroscience).
"""
# Project-relative paths use a templated project-directory token.
XCPD  = Path('${PROJ_DIR}/derivatives/xcpd')  # e.g. <DATA_ROOT>/derivatives/xcpd
PHENO = Path('${PROJ_DIR}/work/analysis_ready_surface.tsv')

# Network definitions: Yeo-7 cortical + 3 subcortical systems (10 total).
YEO7 = ['Vis', 'SomMot', 'DorsAttn', 'SalVentAttn', 'Limbic', 'Cont', 'Default']
SUBCORT_GROUPS = {'Striatum': ['PUT-', 'CAU-', 'NAc-', 'GP'],
                  'Thalamus': ['THA-'], 'HippAmyg': ['HIP-', 'AMY']}
ALL_NETWORKS = YEO7 + list(SUBCORT_GROUPS.keys())
FD_THRESHOLDS = [0.3, 0.5]

ATLAS_CONFIGS = [
    {'id': 'Sch400_Tian3', 'ctx_seg': '4S456Parcels', 'n_ctx': 400, 'assign': 'schaefer'},
    {'id': 'Glasser_Tian3', 'ctx_seg': 'Glasser', 'n_ctx': None, 'assign': 'glasser'},
    # ... remaining atlas configurations crossed with FD_THRESHOLDS ...
]

def compute_network_fc(mat, parcel_network):
    """Mean within- and between-network FC (Fisher-z averaged) per spec."""
    # core loop: for net in ALL_NETWORKS -> within_{net};
    #            for n1,n2 in combinations(ALL_NETWORKS,2) -> between_{n1}_{n2}
\end{suppcode}

\clearpage
\section{Appendix L. Per-case reports (released in the repository)}\label{supp:case-reports}

The automatically generated reports released with the repository are frozen episode snapshots. They preserve the system output, figures, tables, statistics, and provenance available when each report was closed, and they are not silently rewritten when a later episode supersedes the scientific trajectory. The NeuroMark report reflects the corrected, direction-aware re-audit described in \hyperref[supp:s11.2.1]{Supplementary Methods S11.2.1}. The earlier HCP retention-gate and TRIBE language-alignment reports are historical records of separate campaigns; they are not the evidentiary source for the current HCP workflow-search and TRIBE speech--tools episodes. The complete current numerical results and claim boundaries are given in \hyperref[supp:s11.3.1]{S11.3.1}--\hyperref[supp:s11.3.3]{S11.3.3}, with the underlying run bundles and current research-line reports identified through the Data availability statement.

\clearpage
\section{Appendix M. Benchmark human-audit sheet (released in the repository)}\label{supp:human-audit-sheet}

The graded sheet from the human audit of the automated benchmark judges (\hyperref[supp:s11.1]{Supplementary Methods S11.1}) is released with the benchmark package (Data availability; \hyperref[app:j]{Appendix J}). It records a reproducible random 20\% sample (seed~20260630) of the scored results across the reported benchmarks (272 items across tool routing, NIK answer keys, and grounding judgments), each adjudicated by hand against the recorded automated verdict, with a per-item grade and, for the 11 flagged grounding items, the reason the verdict is debatable. The summary and the flagged-item accounting are given in Tables~\ref{tab:human-audit} and~\ref{tab:human-audit-items}.

\bibliography{sn-bibliography}